\documentclass{article}

\usepackage[preprint]{iclr2027_conference}
\usepackage{times}

\usepackage{amsmath,amsfonts,bm}

\def\eqref#1{equation~\ref{#1}}
\def\Eqref#1{Equation~\ref{#1}}

\def\1{\bm{1}}

\DeclareMathAlphabet{\mathsfit}{\encodingdefault}{\sfdefault}{m}{sl}
\SetMathAlphabet{\mathsfit}{bold}{\encodingdefault}{\sfdefault}{bx}{n}

\usepackage{algorithm}
\usepackage{algpseudocode}

\usepackage{graphicx}
\usepackage{amsmath}
\usepackage{amssymb}
\usepackage{dsfont}

\usepackage[utf8]{inputenc}
\usepackage[T1]{fontenc}
\usepackage{hyperref}
\usepackage{url}
\usepackage{booktabs}
\usepackage{amsfonts}
\usepackage{nicefrac}
\usepackage{microtype}
\usepackage{xcolor}

\title{Sven: Singular Value Descent as a Computationally Efficient Natural Gradient Method}

\usepackage{color}

\newcommand{\numCostMemoryNArchs}{5}
\newcommand{\numCostMemoryPMax}{11{,}181{,}642}
\newcommand{\numCostMemoryPMaxArch}{ResNet18}
\newcommand{\numCostMemoryPMin}{593}

\newcommand{\numMnistCESvenKSaturate}{8}

\newcommand{\numMnistLRSvenKSaturate}{32}

\newcommand{\numNConfirmSeeds}{5}

\newcommand{\numPolyEpochs}{20}

\newcommand{\numPolySvenKSaturate}{16}

\newcommand{\numToySvenKSaturate}{16}

\newcommand{\numBatchsizeGrid}{8, 16, 32, 64, 128, 256}

\newcommand{\numBudgetSvenDistinctFracMax}{66.4\%}
\newcommand{\numBudgetSvenDistinctFracMin}{47.5\%}

\newcommand{\numKappaB}{64}

\newcommand{\numKappaMatchedSteps}{0.25, 0.5, 1}

\newcommand{\numKappaNMatched}{3}

\newcommand{\numKappaNSeeds}{5}

\newcommand{\numKappaValues}{1, 2, 3}

\newcommand{\numParamfracToyValues}{0.1, 0.25, 0.5, 0.75, 1}

\newcommand{\numCifarLRParams}{11,181,642}

\newcommand{\numFigFiveCollapseFrac}{0.1}

\newcommand{\numFigFiveHoldFrac}{0.25}

\newcommand{\numProfAmortPct}{80}

\newcommand{\numProfCifarClassicSolveMs}{134}

\newcommand{\numProfCifarFullSolveMs}{22.8}

\newcommand{\numProfMeasuredSteps}{50}

\newcommand{\numProfNArchs}{5}

\newcommand{\numProfStagedBatches}{8}

\newcommand{\numProfWarmupSteps}{10}

\newcommand{\numCifarCESvenEnergyFinal}{76.7}
\newcommand{\numCifarCESvenEnergyFirst}{90.1}

\newcommand{\numCifarLRSvenB}{128}

\newcommand{\numMnistCESvenEnergyFinal}{95.2}
\newcommand{\numMnistCESvenEnergyFirst}{31}

\newcommand{\numPolySvenEnergyFinal}{99.6}
\newcommand{\numPolySvenEnergyFirst}{92.3}

\newcommand{\numSpectraProbeCeiling}{$10^{12}$}
\newcommand{\numSpectraProbeFloor}{$10^{-12}$}

\newcommand{\numToySvenEnergyFinal}{91.5}
\newcommand{\numToySvenEnergyFirst}{92.1}

\newcommand{\numToySvenProbeRankFirst}{14}

\newcommand{\numToySvenProbeWidth}{593}

\ifdefined\svntblbox\else\newsavebox{\svntblbox}\fi
\providecommand{\svnfittable}[1]{%
  \sbox\svntblbox{#1}%
  \ifdim\wd\svntblbox>\linewidth
    \resizebox{\linewidth}{!}{\usebox\svntblbox}%
  \else
    \usebox\svntblbox
  \fi}

\author{%
  Samuel Bright-Thonney$^{1,2}$ \quad Thomas R. Harvey$^{1,2}$ \quad Andre Lukas$^{3}$ \quad Jesse Thaler$^{1,2,4,5}$ \\
  $^1$Department of Physics, Massachusetts Institute of Technology \\
  $^2$ The NSF Institute for Artificial Intelligence and Fundamental Interactions \\
  $^3$ Rudolf Peierls Centre for Theoretical Physics, University of Oxford \\
  $^4$ Institut des Hautes \'Etudes Scientifiques\\
  $^5$ Institut de Physique Th\'eorique, CEA Paris-Saclay\\
}

\iclrfinalcopy

\begin{document}

\rhead{\ifnum\value{page}=1 MIT-CTP/6022\fi}

\maketitle

\begin{abstract}
We introduce Sven (Singular Value dEsceNt), a new optimization algorithm for neural networks that exploits the natural decomposition of loss functions into a sum over individual data points, rather than reducing the full loss to a single scalar before computing a parameter update. Sven treats each data point's residual as a separate condition to be satisfied simultaneously, using the Moore-Penrose pseudoinverse of the loss Jacobian to find the minimum-norm parameter update that best satisfies all conditions at once.  In practice, this pseudoinverse is approximated via a truncated singular value decomposition, retaining only the $k$ most significant directions. We show that Sven can be understood as a natural gradient method generalized to the overparametrized regime, recovering natural gradient descent in the underparametrized limit. We test Sven on a variety of regression and classification tasks, including small-scale language modeling with transformers, and find that it is competitive with leading baselines such as Adam, Muon, and K-FAC.  We also discuss Sven's memory overhead, which presents a barrier to scaling under a naive implementation, and introduce an optimized implementation that keeps memory usage on par with standard baselines under mild restrictions on model architecture.  Beyond standard machine learning benchmarks, we anticipate that Sven will find natural application in scientific computing settings where custom loss functions decompose into several conditions.
\end{abstract}

\section{Introduction}

Every standard loss function is a sum. Whether fitting a regression model or training a classifier, the objective decomposes into a sum over data points, with each term encoding an individual condition the model should satisfy. Despite this structure, the dominant paradigm in machine learning (ML) discards it immediately. Gradient descent reduces the entire collection of conditions to a single scalar and updates parameters accordingly, treating the decomposition as an implementation detail rather than a source of information.

In this paper, we develop a new optimizer for neural networks which takes a global view of this loss decomposition. Our methodology, which we call \textit{Sven} (\textbf{S}ingular \textbf{V}alue d\textbf{E}sce\textbf{N}t), can be viewed as an efficient approximation to natural gradient methods, extrapolated to the overparametrized regime. At each training step, rather than computing a single gradient direction for the total loss, Sven asks a more ambitious question: given the individual residuals of every data point in the batch, what single parameter update would bring each of them closest to zero simultaneously? This is a linear algebra problem, the solution to which makes use of the Moore-Penrose pseudoinverse~\citep{moore1920reciprocal, penrose1955generalized}. In practice, computing the full pseudoinverse can be expensive, so Sven approximates it via a truncated singular value decomposition (SVD), retaining only the $k$ most significant directions. The result is an optimizer that is aware of the geometry of the loss landscape in a way that standard gradient descent is not. 

We focus our proof-of-concept studies on relatively small-scale regression, classification, and language modeling tasks. 
Scaling to larger models is a natural direction for future work, and one we believe is tractable due to a memory-efficient implementation of Sven that we introduce in this work.\footnote{Efficient relative to a naive implementation; ours uses more memory than Adam, but only modestly so.} 
We find that Sven performs on par with standard baselines such as Adam(W)~\citep{kingma2015adam,loshchilov2019adamw}, SGD, RMSprop~\citep{hinton2012rmsprop}, Muon~\citep{jordan2024muon}, and K-FAC~\citep{martens2015kfac}. 
It is also competitive with methodologically similar methods such as Half-Inverse Gradients (HIG)~\citep{schnell2022hig} and Jacobian Descent (JD)~\citep{quinton2024jacobian}, and with the popular quasi-Newton method L-BFGS~\citep{liu1989limited}.
Beyond standard ML benchmarks, Sven finds natural application in scientific computing, with a recent application in theoretical physics~\citep{Benjamin:2026lbj}.

The paper is organized as follows. Section~\ref{sec:meth} derives the update rule, relates it to natural gradients, and describes our optimized implementation; Section~\ref{sec:related} surveys related work and optimizer baselines; Section~\ref{sec:experiments} gives experimental results and analyzes Sven's learning dynamics; Section~\ref{sec:conc} concludes and states limitations. The appendices contain extensive additional information on Sven's mathematical basis (\ref{app:NatGrad}, \ref{app:metric_grad}), experimental details (\ref{app:exp_details}, \ref{app:gram}), further experimental results (\ref{app:additional_results}--\ref{app:lims}) including an illuminating singular value spectrum analysis (\ref{app:spectra}), stability studies (\ref{app:robustness}), and memory/compute profiling (\ref{app:profiling}).

\section{Methodology}\label{sec:meth}

In this section, we present the Sven update rule and build intuition for it through the lens of least-squares regression.
The theoretical backbone is in Appendices~\ref{app:NatGrad} and~\ref{app:mgd}, which develop the connection to natural gradient methods from a functional perspective. Neither is needed to understand the core methodology.

In ML, a common problem is the minimization of a loss function composed of several components:
\begin{equation}\label{eqn:loss_split}
    L(\theta) = \sum_{\alpha \in \aleph} \ell^{\alpha}(\theta),
\end{equation}
where each sub-loss is non-negative, $\ell^\alpha(\theta) \geq 0$ for all $\theta\in \mathbb R^P$ and $\alpha \in \aleph$. Each term $\ell^\alpha$ encodes a condition the model should satisfy, and the overall loss aggregates these conditions into a single objective. This structure naturally arises, for instance, in the $L_2$-loss for regression and the cross-entropy loss for classification, both of which decompose as a sum over individual data points. From now on, we take the index set $\aleph$ to enumerate individual data points, keeping the more general framework in mind and returning to it when relevant.

Despite this structure, minimizing $L(\theta)$ typically proceeds by applying gradient descent to the aggregate loss, disregarding its decomposition (though in practice this is usually done over batches). A more principled approach, however, can explicitly account for the fact that $L(\theta)$ aggregates several distinct conditions. The algorithm we propose, Sven, is given in Algorithm~\ref{alg:update}. 
We now describe its derivation and its relation to existing methods.

\begin{algorithm}[t]
  \caption{Sven update step. The truncated SVD considers only the $k$ largest singular values, and sets singular values that are a factor of \texttt{rtol} smaller than the largest singular value to zero. Appendix~\ref{app:gram} details computational backends for the SVD.}
  \label{alg:update}
  \begin{algorithmic}
    \State {\bfseries Input:} Jacobian $M$, residual vector $R$, rank $k$, tolerance $\text{rtol}$, parameters $\theta$, learning rate $\eta$
    \State {\bfseries Output:} Updated parameters $\theta'$
    \State $U, s, V^T \leftarrow \text{TruncatedSVD}(M, k, \text{rtol})$ \Comment{Compute truncated SVD}
    \For{$i = 1$ to $\text{length}(s)$}
      \If{$s_i > 0$}
        \State $s_i^{-1} \leftarrow 1/s_i$
      \Else
        \State $s_i^{-1} \leftarrow 0$
      \EndIf
    \EndFor
    \State $M^+ \leftarrow V \text{diag}(s^{-1}) U^T$ \Comment{Compute Moore-Penrose inverse}
    \State $\delta\theta \leftarrow - \eta M^+ R$ \Comment{Compute parameter update}
    \State \Return $\theta + \delta\theta$
  \end{algorithmic}
\end{algorithm}

\subsection{Inspiration from Least-Squares Regression}
\label{sec:lstsqr_kappa}

Although we will generalize to an arbitrary loss shortly, we begin by considering a regression problem, where we use the $L_2$-loss between our model $f_\theta(x)$ and some target function $g(x)$. Such a loss is of the form
\begin{equation} \label{eqn:reg_loss}
    L(\theta) = \sum_{\alpha \in \aleph}{\big(\mathcal R^\alpha(\theta)\big)^2} =\sum_{\alpha\in\mathcal \aleph}  \big(f_\theta(\alpha) - g(\alpha) \big)^2.
\end{equation}
Consider the linear expansion of the residual $\mathcal{R}^\alpha(\theta)$ around $\theta_0$:
\begin{equation}
    L(\theta_0 +\delta\theta) = \sum_{\alpha\in\aleph}\left(\mathcal R^\alpha(\theta_0) + \sum_i M^\alpha_{i} \, \delta\theta^i\right) ^2+\mathcal{O}\left(|\delta\theta|^2\right),
\end{equation}
where the Jacobian is
\begin{equation}
    \label{eqn:jacobian}
    M^\alpha_{\:\:i} \equiv \left.\frac{\partial \mathcal{R}^\alpha}{\partial \theta^i}\right|_{\theta = \theta_0} = \frac{\partial f_\theta(\alpha)}{\partial \theta^i},
\end{equation}
and we will omit the higher order terms from now on, taking them as understood. In such a situation, this can be viewed as a linear least squares regression problem for $\delta\theta^i$.

Within the region that the linear expansion is reliable, we seek to solve for all contributions simultaneously. That is, we seek solutions to
\begin{equation}\label{eqn:linear}
    \mathcal R^\alpha(\theta_0) + \sum_i M^\alpha_{\:\:i} \, \delta\theta^i = 0,
\end{equation}
such that the total loss is zero.
Although a solution may not exist, the \textit{closest} approximation to one (in a sense to be made precise shortly) is given by
\begin{equation}\label{eqn:lin_sol_reg}
    \delta \theta^i = -\sum_{\alpha \in \aleph}(M^+)^i_{\:\alpha} \, \mathcal R^{\alpha}(\theta_0) \equiv -(M^+R)_i,
\end{equation}
where $M^+$ is the Moore--Penrose pseudoinverse of $M$~\citep{moore1920reciprocal, penrose1955generalized} and $R \in \mathbb{R}^{|\aleph |}$ is the vector of residuals. We defer the discussion of its efficient computation and approximation to later, and for now focus on its interpretation. For a linear system such as \Eqref{eqn:linear}, the update $\delta\theta^i$ in \Eqref{eqn:lin_sol_reg} admits the following interpretation depending on the regime: with more conditions than parameters, \Eqref{eqn:linear} is overdetermined, generically has no exact solution, and $\delta\theta^i$ is the unique least-squares minimizer; with fewer (overparametrized) it is the \textit{minimum-norm solution} among those that minimize the residual.
Our interest lies primarily in the second case, especially with large neural networks and batched data, though the first case is relevant for drawing connections with natural gradient methods. In either case, we propose a learning scheme where each update step consists of
\begin{equation}\label{eqn:update}
\boxed{
\delta \theta = - \eta M^+ R(\theta_0), \qquad M^\alpha_{\:\:i} \equiv \left.\frac{\partial \mathcal{R}^\alpha}{\partial \theta^i}\right|_{\theta = \theta_0},}
\end{equation}
and the proportionality constant $\eta$ is the learning rate.

In a typical machine learning context, $\alpha$ indexes a batch of size $B$ sampled from training set of size $N$, and the model has parameters $\theta \in \mathbb{R}^P$. The Jacobian $M$ is a $B \times P$ matrix with rows containing per-sample \text{loss} gradients, and is generally distinct from the \textit{model} Jacobian $\partial f_i/\partial\theta_j$.\footnote{The equality in \eqref{eqn:jacobian} is a special case, and only holds for an MSE loss on a 1D model output.} We distinguish two notions of overparametrization: \textit{batchwise} ($P > B$), which applies in any typical ML context and motivates the pseudoinverse Sven update rule; and \textit{dataset-wise} ($P > N$), studied in Appendix~\ref{app:overparam}.

\subsection{General Case}
\label{sec:sven_general_case}
Let us now return to the generic loss of \Eqref{eqn:loss_split}, which we can rewrite suggestively as
\begin{equation}
    \label{eqn:kappa_generalization}
    L(\theta) = \sum_{\alpha\in\aleph} \Big( \mathcal{R}_{\rm eff}^\alpha(\theta) \Big)^{\frac{2}{\kappa}},
    \qquad
    \mathcal{R}_{\rm eff}^\alpha(\theta) \equiv \big( \ell^\alpha(\theta)\big)^{\frac{\kappa}{2}},
\end{equation}
where $\kappa>0$ is a hyperparameter and we term $\mathcal{R}_{\rm eff}$ \textit{effective residuals}. 
\Eqref{eqn:kappa_generalization} is trivially true for every $\kappa$: it does not change the loss being minimized but rather the quantity being treated as the residual, and hence the rows of $M$.
If $\kappa = 1$, then we are simply treating the $\sqrt{\ell^\alpha(\theta)}$ terms as if they were $L_2$-loss residuals $\mathcal{R}^\alpha(\theta)$, and the above derivation goes through essentially unchanged.
For generic $\kappa$, this takes the form of an $L_{2/\kappa}$-loss, which does not have a closed form minimum, even in the linear approximation.
We can, however, simply treat the $\ell^\alpha(\theta)^{\kappa/2}$ terms as if they were residuals of an $L_2$-loss, and use the effective residuals to define the update step and Jacobian in \Eqref{eqn:update}.
We emphasize that the $L_{2/\kappa}$ and $L_2$ norms differ, so this update step is $\kappa$-dependent.%
\footnote{In the underparametrized limit, the choice of $\kappa$ only rescales the update and can be absorbed into $\eta$; it becomes meaningful only in the overparametrized regime, due to the singular value truncation.}
Throughout this work we keep $\kappa = 2$ ($\mathcal{R}_{\rm eff}^\alpha = \ell^{\alpha}$) as the default to avoid pathologies associated with taking fractional powers of generic loss function. We investigate varying $\kappa$ in Appendix~\ref{app:kappaScans}, and find that it has little impact on Sven's performance.

\subsection{Underparametrized Limit and the Relation to Natural Gradients}\label{sec:under-param}
In gradient descent, parameters are updated according to
\begin{equation}
    \delta\theta^i = -\eta\frac{\partial L(\theta)}{\partial \theta^i},
\end{equation}
where $\eta$ is the learning rate. However, this can be generalized by introducing an inverse metric $g^{ij}$ (with corresponding metric $g_{ij}$) on the space of parameters:
\begin{equation}
    \delta\theta^i = -\eta \, \sum_j g^{ij}\frac{\partial L(\theta)}{\partial \theta^j}.
\end{equation}
Such an inverse metric is called a preconditioner in the ML literature, with Adam being the most famous example~\citep{kingma2015adam, harvey2025optimiser,absil2009optimization,fei2025survey,hinton2012rmsprop,jordan2024muon}.
Natural gradients offer a principled, though computationally expensive, choice of preconditioner~\citep{natgrad,bernacchia2018exact, shrestha2023natural}. If one had access to this preconditioner at no cost, it would be provably the most efficient choice for training. In particular, in ideal conditions the loss decreases exponentially with training time, rather than with power-like scaling laws~\citep{hestness2017deep,kaplan2020scaling}.

For our purposes, when dealing with functions, the natural gradient metric is given by
\begin{equation}\label{eq:nat_met}
    g_{ij} = \frac{1}{|\mathcal D|}\sum_{x\in \mathcal D} \frac{\partial f_\theta(x)}{\partial\theta^i} \frac{\partial f_\theta(x)}{\partial\theta^j},
\end{equation}
where $\mathcal D$ is our dataset. An analogous expression exists for probability distributions, where the metric is given the name Fisher Information Metric (FIM).

Returning to the regression example of \Eqref{eqn:reg_loss}, consider the underparametrized limit where the parameter index $i$ spans fewer values than the data index $\alpha$. In this special case, the Moore-Penrose inverse can be written as
\begin{equation}\label{eqn:MPUnder}
    M^+ = (M^T M)^{-1} M^T.
\end{equation}
Substituting this into \Eqref{eqn:update}, we find
\begin{equation}
    \delta \theta^i \propto -\sum_{j} \left[\left(M^T M\right)^{-1}\right]^{ij} \frac{\partial L(\theta)}{\partial \theta^j} ,
\end{equation}
which, up to a normalization, is the same as what we expect from natural gradient methods, where the sum is replaced with an integral over the data.

\subsection{Overparametrized Limit and Our Algorithm}
Natural gradient methods are not straightforwardly defined in the overparametrized regime: the metric of \Eqref{eq:nat_met} becomes singular once $P$ exceeds the number of data points and cannot be inverted to yield an update. This is precisely the regime modern ML operates in, and the natural resolution --- the pseudoinverse of $g_{ij}$ --- is an intractably large $P \times P$ matrix. Sven instead takes the pseudoinverse of the $B \times P$ Jacobian $M$, which is much smaller in the typical case where $B \ll P$.
The derivation in the overparametrized limit follows the same initial steps as the previous section until the substitution of the specific form of the pseudoinverse (\Eqref{eqn:MPUnder}). In the $P > B$ regime, we calculate the pseudoinverse $M^{+}$ using the SVD of the matrix $M$.

We can further approximate this inverse by retaining only the $k$ largest singular values, and the cost then depends on how the truncated pseudoinverse is obtained. Using randomized algorithms~\citep{halko2011finding} to compute the truncated SVD of the $B \times P$ Jacobian costs $\mathcal{O}(k P B)$, a factor of $k$ over a gradient descent step, and the Jacobian must be stored in memory. 
In Appendix~\ref{app:gram} we show this cost can often be avoided, as the Sven update depends on $M$ only through the eigendecomposition of the $B \times B$ Gram matrix $MM^T$ and a vector--Jacobian product, so one step reduces to a constant number of forward and backward passes and a small eigendecomposition. In many cases, the Gram matrix can be accumulated from autograd artifacts without ever instantiating $M$ fully in memory. The full cost is then \textit{independent} of $k$, as $k$ and \texttt{rtol} decide how many directions are inverted rather than how many are computed.

Sven can be understood as a natural gradient method generalized to the overparametrized regime. The pseudoinverse provides a principled update rule when the natural gradient metric becomes singular, and reduces cleanly to natural gradient descent in the underparametrized limit.
Combining the above, we are left with the update step shown in Algorithm~\ref{alg:update}. Code for a PyTorch~\citep{anselpytorch} implementation of Sven is linked in App.~\ref{sec:code}, with implementation details in Appendix~\ref{app:gram}.
All experiments in the main text use the efficient Sven optimization referenced above, which works for architectures \textit{not} including BatchNorm~\citep{ioffe2015batch} or tied weights. Appendix~\ref{app:cifar_10} applies Sven to a ResNet18 trained on CIFAR-10, where BatchNorm prohibits the efficient implementation and introduces larger memory overheads for Sven. Appendix~\ref{app:profiling} provides detailed memory and wall-time profiling data for all Sven backend implementations.

\section{Related Work}
\label{sec:related}

Sven sits at the intersection of several active areas of research: natural gradient methods, second-order and quasi-Newton optimizers, and Jacobian-based pseudoinverse methods. We survey the most closely related work, highlighting how Sven differs from and extends prior approaches.

\paragraph{Natural gradient methods.}
Natural gradient descent~\citep{natgrad} replaces the Euclidean update with one computed using the FIM, yielding parameter-invariant updates that are theoretically optimal in the information-geometric sense. \citet{martens2020new} shows that the FIM is equivalent to the generalized Gauss-Newton (GGN) matrix for exponential family losses. As we showed in Section~\ref{sec:under-param}, Sven recovers natural gradient descent exactly in the underparametrized limit, and generalizes it to the overparametrized regime by using the pseudoinverse of the full loss Jacobian in place of the singular natural gradient metric.

K-FAC~\citep{martens2015kfac} approximates the FIM as a block-diagonal matrix with Kronecker-product blocks, yielding efficient natural gradient updates at roughly $2$--$3\times$ the cost of SGD. Unlike K-FAC, which imposes a Kronecker structure within each layer and accumulates statistics across iterations, Sven computes a per-step truncated SVD of the loss Jacobian across all parameters simultaneously. K-FAC does not resolve the overparametrized singularity that motivates Sven: its Fisher is still a $P \times P$ block-Kronecker object, rank-deficient when $P$ exceeds the number of conditions, and exists only through damping, which gives a biased rather than a minimum-norm update.

\paragraph{Second-order and quasi-Newton optimizers.}
Adam~\citep{kingma2015adam,loshchilov2019adamw} remains the dominant optimizer for deep learning, maintaining diagonal estimates of first and second gradient moments as a cheap preconditioner; Sven includes off-diagonal structure that Adam ignores entirely. Shampoo~\citep{gupta2018shampoo} and its descendant SOAP~\citep{vyas2025soap} maintain Kronecker-factored preconditioning matrices, accumulating second moment statistics over time rather than computing a per-step spectral decomposition as Sven does. Muon~\citep{jordan2024muon} orthogonalizes the gradient matrix via its polar factorization $UV^T$, equivalently replacing all singular values with 1; in contrast, Sven \emph{inverts} the singular values, applying more weight to directions with smaller singular values as a curvature-correcting strategy.
The Levenberg-Marquardt algorithm~\citep{marquardt1963algorithm,hagan1994training} interpolates between gradient descent and Gauss-Newton via a damping parameter, shrinking all singular values uniformly; Sven's hard truncation of the top-$k$ directions provides an alternative regularization strategy, though we do not compare the two empirically. 

\paragraph{Jacobian pseudoinverse and Gauss-Newton methods.}
The most closely related work to Sven is Exact Gauss-Newton (EGN)~\citep{korbit2024egn}, which also factorizes a $Bc \times Bc$ matrix\footnote{$c$ is the model output dimension} rather than a $P \times P$ one when the batch size $B \ll P$, doing so via the Woodbury identity. Sven differs in operating on the per-sample \emph{loss} Jacobian rather than the output Jacobian $J_f$, and in using a truncated SVD of tunable rank $k$ rather than solving the system exactly.
Jacobian Descent (JD)~\citep{quinton2024jacobian} shares the same starting point as Sven, but aggregates Jacobian rows with a projection-based operator whereas Sven takes the pseudoinverse. The natural-gradient framing is also absent from JD.
Half-Inverse Gradients (HIG)~\citep{schnell2022hig} is mechanistically the closest precursor to Sven: it computes the SVD of a Jacobian and raises singular values to a fractional power $\kappa$, with $\kappa = -1$ recovering the full pseudoinverse. Sven differs in using hard truncation rather than a continuous fractional power, targeting general neural network training rather than physics-in-the-loop optimization, and operating on the per-sample loss Jacobian with an explicit natural gradient interpretation. HIG is also much more computationally expensive than Sven, as it uses the \textit{model} Jacobian which scales with the number of outputs.\footnote{For example, HIG takes $\sim$10x longer than Sven in our MNIST experiments, and is totally intractable for a transformer-scale model.}

To summarize the key distinction: while prior work has taken the SVD over other summation indices of the loss, we are unique in decomposing over the data index and drawing the resulting connection to natural gradient methods.

\section{Experiments}
\label{sec:experiments}

\begin{figure}[!ht]
    \centering
    \includegraphics[width=0.97\linewidth]{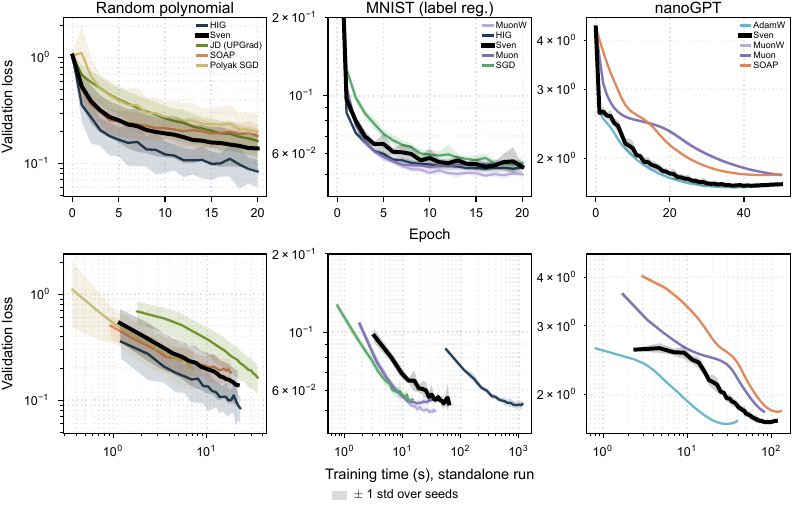}
    \caption{Validation loss against epoch (top) and against wall time (bottom) for Sven and four best-performing baselines on the three headline tasks: random polynomial regression (left), MNIST classification with label regression (middle), and tiny-Shakespeare language modeling with nanoGPT (right). Curves are means over 5 seeds, with optimizer parameter settings chosen according to lowest mean validation loss at the end of training.}
    \label{fig:main_results}
\end{figure}

We demonstrate Sven on relatively small-scale problems where we can run detailed hyperparameter scans to understand its convergence properties. Code for reproducing all experiments is linked in App.~\ref{sec:code}. Our main results, summarized in Fig.~\ref{fig:main_results}, involve three datasets:
\begin{enumerate}
	\item \textbf{Random polynomial}: a random polynomial over $\mathbb{R}^6$ built from all monomials of total degree $\leq 4$, each coefficient variance-normalized so no degree dominates. Inputs are IID Gaussian and we evaluate on three independently drawn polynomial targets to account for variation in learnability.
	\item \textbf{MNIST}~\citep{lecun1998mnist} under a \emph{label regression} loss $\|f_\theta(\mathbf{x}_i) - \mathbf{y}_i\|^2$ against the one-hot label, which puts classification on the same footing as regression; cross-entropy on the same model and splits is in Appendix~\ref{app:mnist_ce}. 
	\item \textbf{nanoGPT}~\citep{karpathy2022nanogpt}: a four-layer decoder-only transformer on a character-level Shakespeare dataset. 
\end{enumerate}
Appendix~\ref{app:datasets} provides full details for each. We also analyze a toy 1D regression problem (Appendix~\ref{app:additional_results}), ResNet18 on CIFAR-10 (Appendix~\ref{app:cifar_10}), and a limited pilot LLM training run using GPT-2-small~\citep{radford2019gpt2} on FineWebText-edu~\citep{penedo2024fineweb} (Appendix~\ref{app:transformers}). We omit CIFAR-10 from the main results because its BatchNorm prevents using the efficient Sven implementation, and omit GPT-2 because resource constraints prevented us from performing hyperparameter optimization or running multiple seeds.

We compare Sven against 14 common baselines: SGD, SGD with momentum, RMSprop~\citep{hinton2012rmsprop}, Adam~\citep{kingma2015adam}, AdamW~\citep{loshchilov2019adamw}, Polyak-step SGD~\citep{polyak1969minimization, hazan2019revisiting}, stochastic L-BFGS~\citep{liu1989limited}, SOAP~\citep{vyas2025soap}, Shampoo~\cite{gupta2018shampoo}, K-FAC~\citep{martens2015kfac}, Muon~\citep{jordan2024muon} with and without weight decay, HIG~\citep{schnell2022hig}, and JD~\citep{quinton2024jacobian}. On MNIST, K-FAC was too unstable to optimize and SOAP failed to converge; both are omitted. Muon optimizes embedding layers and output heads with AdamW, using the orthogonalized Muon update for all other linear and convolutional weights (with kernels flattened), and the step RMS-matched to an AdamW step~\citep{liu2025muon}. Due to resource constraints, we optimize nanoGPT with only Sven, AdamW, Muon/MuonW\footnote{We denote Muon + weight decay as MuonW; Sven is un-regularized, so we include un-regularized Adam and Muon in the baseline pool}, and SOAP. Polynomial regression and MNIST are trained using MLPs with three hidden layers and a larger hidden width for MNIST ($P = 673$ and $27562$, respectively), while nanoGPT uses a four-layer decoder-only transformer ($P \approx 826$k). Each baseline is tuned over a learning rate grid, while Sven is tuned over a $(\eta,k,\mathrm{\texttt{rtol}})$ grid. For comparison, we select the best-performing hyperparameter configuration from the grid for each optimizer, where performance is measured using final validation loss. Results are averaged over five independent model seeds. See Appendix~\ref{app:exp_details} for full experimental details.

\subsection{Results}

Figure~\ref{fig:main_results} summarizes the main results, showing validation loss over the course of training (top) and as a function of training wall-time (bottom) for Sven and the four best-performing baselines (Fig.~\ref{fig:main_results_with_errorBars} in Appendix~\ref{app:additional_results} shows all optimizers). Sven matches or beats most baselines within seed uncertainty. HIG outperforms Sven on the polynomial task but suffers when scaled even modestly to the MNIST task, where it costs a factor of ten more wall-time. Sven matches AdamW on nanoGPT, showing that it can perform well on cross-entropy targets as well, and costs similar wall-time to Muon and SOAP. 

\begin{figure}[t]
    \centering
    \includegraphics[width=\linewidth]{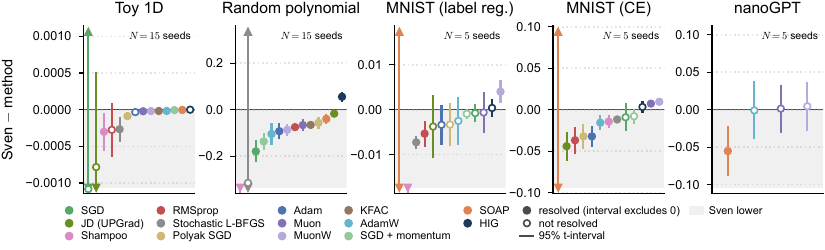}
    \caption{Mean paired differences in test set loss relative to Sven for all baseline optimizers on the synthetic regression, MNIST, and nanoGPT tasks. Error bars are Student-$t$ 95\% intervals on the mean difference over the $N$ seed pairs (indicated in each plot); a filled marker is an interval that excludes zero. An interval wider than the panel is clipped to an arrow.}
    \label{fig:paired_main}
\end{figure}

Figure~\ref{fig:paired_main} provides a more detailed account of Sven's performance against all baselines, and includes toy 1D regression and MNIST cross-entropy results. In each panel we show \textit{paired differences} relative to Sven: each optimizer $O$ is trained on $N$ independent model seeds using its best hyperparameter configuration, and for each seed $i = 1,\ldots, N$ we compute test set loss $L^{\rm test}_{i,O}$ and the difference $\Delta_{i,O} = L^{\rm test}_{i,\mathrm{Sven}} - L^{\rm test}_{i,O}$. Fig.~\ref{fig:paired_main} plots the mean paired difference and 95\% confidence intervals over $N$ seeds. Sven beats a large fraction of the baselines in most cases, though it is never unambiguously the best. Interestingly, it more consistently outperforms baselines on MNIST cross-entropy than label regression. These results also use losses on an independent test set, which explains the larger error bars and changed order on nanoGPT relative to Fig.~\ref{fig:main_results}.\footnote{Given the small dataset size, nanoGPT likely overfits.}

\subsection{Analyzing Singular Value Spectra}
\label{sec:spectra}

Sven's optimization behavior is closely related to the singular value spectrum of a batch's loss Jacobian $M$, and in this section we analyze that structure to better understand what Sven is doing. First, we note that $k$ and \texttt{rtol} are not truly independent parameters depending on the shape of the SV spectrum: if $\sigma_i/\sigma_1 > \mathrm{\texttt{rtol}}$ for all $B$ SVs of $M$, then $k$ fully controls truncation; likewise, if $k$ is set high and the spectrum falls rapidly, \texttt{rtol} controls the truncation threshold which will vary batch-to-batch and generally be $\leq k$. In practice we find optimal $(k,\mathrm{\texttt{rtol}})$ settings via a hyperparameter scan, but portions of this grid may be effectively identical depending on the shape of the SV spectrum for a particular task.

\begin{figure}[t]
    \centering
    \includegraphics[width=\linewidth]{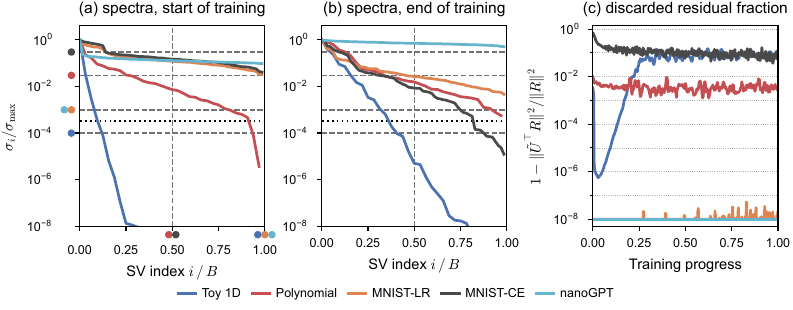}
    \caption{Normalized singular value spectra at the beginning (a) and end (b) of training. Horizontal lines mark the selected \texttt{rtol} and vertical lines the selected $k/B$, with colored dots indicating the corresponding datasets. The dotted black line is the float32 round-off floor. (c) The fraction of the batch residual the truncation discards over the course of training.}
    \label{fig:spectrum_truncation}
\end{figure}

In Fig.~\ref{fig:spectrum_truncation} (a,b) we plot full SV spectra at the beginning and end of training for the toy 1D, polynomial, MNIST, and nanoGPT tasks. Spectra are recorded from runs with the best-performing Sven hyperparameters as measured in the previous section. On each panel we indicate each task's selected \texttt{rtol} and $k$ with horizontal and vertical dashed lines, respectively. The different shapes of each problem are immediately clear. The 1D regression task is very hierarchical, with much of its spectrum lying below \texttt{rtol} and below the float32 round-off floor, below which SVs are not meaningful.\footnote{Sven takes $\sigma_i = \sqrt{\lambda_i}$ from the Gram matrix $G = MM^\top$, whose entries inherit the float32 rounding of $M$ as an absolute error ${\sim}\epsilon_{32}\sigma_{\max}^2$ on every eigenvalue. SVs are only resolvable above $\sqrt{\epsilon_{32}}\,\sigma_{\max} \approx 3\times10^{-4}\sigma_{\max}$ (forming $MM^\top$ squares the condition number).} The MNIST and nanoGPT spectra are quite the opposite, lying above their \texttt{rtol} cutoff (with the exception of MNIST-CE). The spectra also evolve during training, some becoming steeper and others shallower without a clear trend. These curves also make it clear that \texttt{rtol} seems more influential than $k$. MNIST-LR, toy 1D, and nanoGPT all select $k = B$, leaving \texttt{rtol} to control truncation, which it only does in practice for toy 1D. For MNIST-CE and polynomial, where $k = B/2$ is chosen, the spectrum shapes suggest \texttt{rtol} still controls truncation. 

Beyond inspecting the spectra, we can directly probe how much of the per-batch residual the Sven update accounts for. Denoting $R = (\mathcal{R}_1,\ldots,\mathcal{R}_B)^T$  as the vector of batch residuals (i.e.\ per-sample losses for $\kappa = 2$) and the Jacobian as $M = U\Sigma V^T$, the untruncated Sven update amounts to $\delta\theta \propto -M^+R = -V\Sigma^{-1}U^TR$. Letting $\tilde{U} = (\mathbf{u}_1,\ldots,\mathbf{u}_k)$ denote the \textit{truncated} set of singular vectors, we can measure how much of the residual vector is accounted for by the truncated update -- i.e.\ how much of $R$ lives in $\mathrm{span}\,\tilde{U}$. Fig.~\ref{fig:spectrum_truncation} (c) plots $1 - ||\tilde{U}^TR||^2/||R||^2$ -- the fraction of $R$ \textit{lost} by truncating -- over the course of training. MNIST-LR and nanoGPT use an un-truncated spectrum throughout training and sit near zero, while for MNIST-CE it appears beneficial to discard $\sim$10\% of the residual-containing directions during optimization. The toy 1D fraction evolves over training, starting near zero and ending around 10\%. Looking at the spectra, it is clear that the vast majority of the residual lives in a small handful of directions early in training (steep spectrum) but expands to significantly more by the end of training, where \texttt{rtol} cuts off half the spectrum and discards 10\% of the residual. 

These results suggest that Sven shapes and interacts with a rich and varying singular value landscape over the course of training. Studying this interplay in closer detail is a promising avenue for future work. The results of Fig.~\ref{fig:spectrum_truncation} also suggest a potential hyperparameter prescription for Sven that avoids $k$ and \texttt{rtol}: selecting a truncation threshold dynamically at each step based on a desired target for the ``fraction of explained residual'' $||\tilde{U}^TR||^2/||R||^2$. We leave this for future work.

\section{Conclusion}\label{sec:conc}

In this paper we introduced a new optimizer, Sven, which can be viewed as a natural-gradient-inspired method that is tractable in the overparametrized regime. While natural gradient methods are prohibitively expensive in practice, a Sven step with the efficient backend costs only a constant number of forward and backward passes and a $B\times B$ eigendecomposition. Our experiments show that Sven is competitive with mainstream baselines, is effective on both regression and classification problems, and is both viable and competitive on a small-scale transformer language modeling task.
We view Sven not as a replacement for existing optimization machinery, but as a complementary technique to be added to the practitioner's toolkit. Modern neural network training already combines a rich collection of methods (weight decay, gradient clipping, learning rate scheduling, momentum, and adaptive scaling among them) each addressing a distinct aspect of optimization. Sven adds to this toolkit a principled mechanism for exploiting the singular value structure of the loss Jacobian, and can be composed with these existing techniques.

\paragraph{Limitations and future work.} Sven is not easily transferable to architectures with BatchNorm or tied weights, which must use the less memory/compute-efficient backend. Appendix~\ref{app:cifar_10} reports poor performance on CIFAR-10/ResNet18, though the reasons why are not understood. Appendix~\ref{app:transformers} also shows poor performance in a preliminary scaling study on GPT-2-small, though Sven was not tuned \textit{at all} due to resource limitations and this result should be considered preliminary at best. Immediate future directions are scaling to LLMs and finding solutions that enable an efficient backend for more architectures. 
Sven is also promising for scientific computing, where losses might decompose over collocation points or boundary conditions whose contributions carry meaning, as in the theoretical physics application~\citep{Benjamin:2026lbj}.

\section*{Acknowledgments}

The authors would like to thank James Halverson, Philip Harris, and Fabian Ruehle for useful discussions.

SBT, TRH and JT are supported by the National Science Foundation under Cooperative Agreement PHY-2019786 (The NSF Institute for Artificial Intelligence and Fundamental Interactions, http://iaifi.org/), by the Simons Foundation through Investigator grant 929241, and by the MIT Generative AI Impact Consortium (MGAIC).

AL is supported by the STFC consolidated grant ST/X000761/1.

JT is also supported by the U.S.\ Department of
Energy (DOE) Office of High Energy Physics under grant number DE-SC0012567, and he thanks the Institut des Hautes \'Etudes Scientifiques (IHES) and the Institut de Physique Th\'eorique (IPhT) for providing an inspiring sabbatical environment to carry out this research.

\section*{AI Use Statement}
\label{sec:ai_statement}
In preparing this work we used LLM assistants for a wide variety of tasks (Anthropic's Claude Code and OpenAI's Codex). The core idea behind the paper came from the authors, as did the design of numerical experiments and the interpretation/commentary on their results. An early version of the Sven optimizer codebase containing its core functionality was written fully by the authors. Coding agents assisted with the improved memory-efficient backend, adding logging functionality, and optimizing the authors' implementations. They also helped clean up the codebase and convert it into an installable package form. An early version of the framework for running numerical experiments and creating figures was similarly written fully by the authors, and subsequently refined and expanded using coding agents. 

We did \textit{not} use generative AI to write any of the main text. LLM assistants helped in writing skeleton versions of the technical appendices, including tables and text summarizing the details of our experimental setup (hyperparameter grids, model configurations, datasets and splits, etc.). We also consulted LLMs for feedback on the design and scope of our numerical experiments, and incorporated their feedback when it was judged to be sound (e.g.\ expanding hyperparameter grids). We also used LLMs to aid with literature review.

Every piece of LLM-generated appendix text was thoroughly checked for correctness and heavily edited by the authors before presentation. We take full responsibility for all figures, text, code, and experimental artifacts associated with this work, regardless of the level of LLM involvement. 

\section*{Reproducibility Statement}
\label{sec:reproducibility_statement}

The results presented in this paper are reproducible from the code linked in Appendix~\ref{sec:code}, up to unavoidable variations due to floating-point error and/or randomized algorithms (where applicable). The appendices include full experimental details, and all results are derived from runs over a fixed set of model, data, split, and data loader seeds written in configuration files in the experimental codebase. Results related to memory and/or time profiling will vary based on the hardware used to run those experiments.

\bibliography{references}
\bibliographystyle{iclr2027_conference}

\appendix

\section{Code Availability}
\label{sec:code}
Source code for the Sven optimizer package is available at \url{https://github.com/sambt/sven}. Code for the experiments is available at \url{https://github.com/sambt/sven-experiments}.

\section{Natural Gradients and Functional Gradient Descent}\label{app:NatGrad}
As mentioned in Section~\ref{sec:under-param}, natural gradient methods offer a principled and powerful approach to preconditioning gradient-based optimization~\citep{natgrad,shrestha2023natural}. Furthermore, natural gradient methods can be viewed as an approximation to functional gradient descent. In this appendix, we present key elements of natural gradient methods from this functional perspective for the interested reader. Throughout this section, we assume the object being optimized is a function, as opposed to a probability distribution as would be the case in classification problems. Analogous arguments can be carried out for probability distributions, where one would begin with the Kullback–Leibler divergence between distributions, rather than the $L_2$ distance between functions.\footnote{By Chentsov's theorem, any $f$-divergence yields the same metric up to a scalar, making the FIM the unique reparametrization-invariant metric on the space of probability distributions~\citep{amari2016information, chentsov1982statistical, cover2006elements}.} This appendix is intended as a pedagogical introduction; the arguments presented are not formal proofs but instead focus on intuition rather than rigor.\footnote{\label{footnote:expressive}In particular, we will frequently transform between different parametrizations, without worrying about the expressivity of those parametrizations. Such considerations become less important as the number of parameters increases, and would only complicate this discussion.} The approach taken here also offers a somewhat different perspective than the majority of the literature on this subject.

Our approach is as follows: we define a metric on the abstract space of functions, then pull back that metric to the submanifold defined by our function parameterization. This submanifold can be thought of as the loss landscape. We can then use this metric as the preconditioner for gradient descent. Consider two functions $f(x)$ and $g(x)$, and we define the distance between them as being the $L_2$ distance:
\begin{equation}
    d_{L_2}^2[f,g] = \int d\mu(x) \big(f(x) - g(x)\big)^2,
\end{equation}
where $d \mu(x)$ is the measure induced by the data. In other words, this should be understood as an integral over the data manifold. Now if we consider the distance between $f(x)$ and $f(x) + \delta f(x)$, where $\delta f(x)$ is some small function variation, we can define a line element on the space of functions:
\begin{equation}
    |df|^2 = d_{L_2}^2[f,f+\delta f] = \int d\mu(x) \big(\delta f(x)\big)^2.
\end{equation}
We now impose that we are using some parametrized family of functions $f(x) \approx f_\theta(x)$, where $\{\theta^i:i=1\ldots N\}$ are our parameters and the space they parametrize is the loss landscape $\mathcal L$. Upon restricting to the loss landscape, our functional variations pull back to
\begin{equation}
\delta f(x)|_{\mathcal L} = \sum_i\frac{\partial f_\theta(x)}{\partial\theta^i} d\theta^i,
\end{equation}
and the metric on the abstract space of functions pulls back to
\begin{equation}
    g_{ij} = \int d\mu(x) \frac{\partial f_\theta(x)}{\partial\theta^i} \frac{\partial f_\theta(x)}{\partial\theta^j},
\end{equation}
which, once the integral is evaluated via Monte Carlo sampling, yields the natural gradient metric presented in Section~\ref{sec:under-param}.

While the above demonstrates that the natural gradient method provides a principled metric on the loss landscape, we have not yet shown why it is useful. To do so, we must invert the natural gradient metric and consider the resulting gradient flow. In other words, we seek solutions to
\begin{equation}
    \frac{d\theta^i(t)}{dt} = - \sum_j g^{ij} \frac{\partial L(\theta)}{\partial\theta^j},
\end{equation}
where $g^{ij}$ is the inverse of $g_{ij}$. To make this problem tractable, we first consider whether it is possible to find a parametrization where the natural gradient metric reduces to the Euclidean metric. Using the shorthand
\begin{equation}
    f_i(x) = \frac{\partial f_\theta(x)}{\partial \theta^i},
\end{equation}
setting $g_{ij}=\delta_{ij}$ implies that the functions $f_i(x)$ are orthonormal. In other words, if we express the function as a linear combination of orthonormal functions on the data manifold, then gradient descent on these coefficients is equivalent to natural gradient descent with an arbitrary parametrization (up to the expressivity concerns mentioned in the earlier footnote).

One may worry whether a set of orthonormal functions exists on an arbitrary data manifold. While the necessary conditions for such a set to exist are somewhat complicated, one sufficient condition is that the manifold be compact. Given a finite dataset, there are clearly infinitely many possible data manifolds from which it could have been sampled, infinitely many of which are compact. Therefore, while constructing such functions may be complicated, it is reasonable to assume their existence for our analysis. As such, the natural gradient method is actually a flat metric on the data manifold.

Let us consider a simple regression problem with the loss function
\begin{equation}
    L = \int d\mu(x) \big(f_\theta(x) - f(x)\big)^2,
\end{equation}
where $f(x)$ is some target function determined by our data. We can, at least formally, expand both $f_\theta(x)$ and $f(x)$ in our orthonormal basis
\begin{equation}
    f_{\theta}(x) = \sum_i \alpha^i(\theta) f_i(x) \quad \& \quad f(x) = \sum_i \beta^i f_i(x).
\end{equation}

We want to study gradient descent with respect to $\theta^i$ using the natural gradient metric. As argued earlier, this is equivalent to gradient descent with respect to $\alpha^i$:
\begin{equation}
    \frac{d\theta^i(t)}{dt} = -\sum_j g^{ij}\frac{\partial L}{\partial\theta^j} \quad\equiv \quad \frac{d\alpha^i(t)}{dt} = - \frac{\partial L}{\partial\alpha^i},
\end{equation}
which dramatically simplifies our analysis. In fact, we can analytically solve this differential equation. The solution is given by
\begin{equation}
    \alpha^i(t) = \left( \alpha^i(0) - \beta^i \right)e^{-2t} + \beta^i,
\end{equation}
which implies that the loss decays as
\begin{equation}
    L\propto e^{-4t}.
\end{equation}
Therefore, if we were given the natural gradient metric without computational cost (or equivalently, a set of orthonormal functions on the data manifold), the loss would decay exponentially rather than exhibiting the power-law behavior predicted by scaling laws~\citep{hestness2017deep,kaplan2020scaling}.

\section{Gradient Descent and Functional Analysis}\label{app:mgd}
\label{app:metric_grad}

In this appendix, we would like to discuss metric-driven gradient descent from a different viewpoint, using the language of functional analysis and differential geometry. This will allow us to put some of the statements in the previous section on a more rigorous footing. We start with a neural network $f_\theta:\mathbb{R}^d\rightarrow\mathbb{R}$ (real-valued for simplicity but easily generalized) with parameters $\theta = (\theta^i)$, where $i=1\ldots N$, and a loss function $L:\mathbb{R}^N\rightarrow\mathbb{R}$. The standard gradient descent equation,
\begin{equation}\label{gd0}
 \delta\theta^i=-\eta\,\frac{\partial L(\theta)}{\partial \theta^i}\; ,
\end{equation}
as written (where $\eta$ is the learning rate) is clearly somewhat unsavory since the LHS corresponds to a vector field and the RHS to a differential form (as indicated by the different positions of the indices). The obvious way to rectify this is to introduce a structure on parameter space which induces an identification of tangent and co-tangent bundle, such as a Riemannian or symplectic structure. Opting for the former (although it might be interesting to explore the latter option) we introduce a metric $g=(g_{ij})$ with inverse $g^{-1}=(g^{ij})$ on parameter space and modify \eqref{gd0} to
\begin{equation}\label{gdg}
 \delta\theta^i=-\eta\,\sum_jg^{ij}\,\frac{\partial L(\theta)}{\partial \theta^j}\; .
\end{equation}
Here is a way to choose the metric $g$. We introduce a Hilbert space $\mathcal{H}=\mathcal{L}^2_w(\mathbb{R}^d)$, an $\mathcal{L}^2$ space with weight function $w:\mathbb{R}^d\rightarrow\mathbb{R}^{>0}$, standard scalar product
\begin{equation}
\langle f,h\rangle:=\int_{\mathbb{R}^d}d^dx\, w(x)f(x)h(x)\;, 
\end{equation}
and associated norm $||\cdot ||$. We choose the weight function $w$ such that $f_\theta\in\mathcal{H}$ for all $\theta$. (This is related to footnote~\ref{footnote:expressive} on expressivity of different function classes in the previous appendix.) In this case, the map $\theta\mapsto f_\theta$ defines an $N$-dimensional manifold $\mathcal{L}$ in $\mathcal{H}$, which is the same manifold we identified with the loss-landscape in the previous appendix. Relative to an orthonormal basis $(h_I)$ on $\mathcal{H}$ we can expand the neural network as well as a function $f\in\mathcal{H}$, representing the ``data'', as
\begin{align}
f_\theta&\simeq\sum_{I\in\mathcal{I}}a_I(\theta) h_I\;,& a_I(\theta)&=\langle h_I,f_\theta\rangle,\label{ftexp}\\
 f&\simeq\sum_{I\in\mathcal{I}}\alpha_Ih_I\;,& \alpha_I&=\langle h_I,f\rangle\; , \label{fexp}
\end{align}
where we have truncated the sums to a finite index set $\mathcal{I}$, as would be required for a computational realization. The associated linear subspace $\mathcal{W}={\rm span}(h_I)_{I\in\mathcal{I}}\subset\mathcal{H}$ should be ``large enough'' (and it would be interesting to quantify this requirement) to approximate the manifold $\mathcal{L}$ sufficiently well. We note that the expansion in \eqref{fexp} provides the optimal approximation to $f$ within $\mathcal{W}$ in the sense that $||f-\sum_{I\in\mathcal{I}}\alpha_I h_I||$ is minimal for the values of $\alpha_I$ as in \eqref{fexp}. In other words, in the Hilbert space picture there is a unique and well-defined optimal approximation for the `data function' $f$, provided by \eqref{fexp}.\\[2mm]
Let us now discuss gradient descent and, for simplicity, work with a mean square loss, which can be written as $L(\theta)=\frac{1}{2}||f_\theta -f||^2\simeq\frac{1}{2}\sum_{I\in\mathcal{I}}(a_I(\theta)-\alpha_I)^2$. Applying the metric gradient descent in \eqref{gdg} to this loss function gives
\begin{equation}\label{gdtheta}
 \delta\theta^i=-\eta\,\sum_j g^{ij}\,\left\langle\frac{\partial f}{\partial\theta^j},f_\theta-f\right\rangle\simeq -\eta\,\sum_j g^{ij}\,\sum_{I\in\mathcal{I}}\frac{\partial a_I}{\partial\theta^j}(\theta)(a_I(\theta)-\alpha_I)\; .
\end{equation}
How does the training trajectory described by \eqref{gdtheta} translate to the Hilbert space expansion coefficients $a_I$? A short calculation shows that
\begin{equation}
 \delta a^I=-\eta\sum_J G^{IJ}\, (a_J-\alpha_J)\;,\qquad G^{-1}=Jg^{-1}J^T\;, \label{gda}
\end{equation}
where $J=\frac{\partial a}{\partial\theta}$ is the Jacobi matrix of the coordinate change.  There is a canonical choice of metric on $\mathcal{H}$, namely $G_{IJ}=\delta_{IJ}$, the metric induced by the scalar product relative to the orthonormal basis. For this choice of $G$, we obtain the equation
\begin{equation}
 Jg^{-1}J^T=\mathds{1}\;, \label{geq}
\end{equation}
for $g$ and, defining $\epsilon_I=a_I-\alpha_I$, the continuous version of \eqref{gda} turns into 
\begin{equation}
\frac{d \epsilon_I}{dt}(t)=-\eta\, \epsilon_I(t)\quad\Rightarrow\quad \epsilon_I(t)\sim e^{-\eta t}\; .
\end{equation}
 Evidently, this means the unique best approximation for $f$ in $\mathcal{W}\subset\mathcal{H}$, represented by the coefficients $\alpha_I$, is approached at an exponential rate. By choosing the natural metric on the Hilbert space we have obtained efficient exponentially fast training. To translate this training trajectory to the original neural network parameters we have to consider \eqref{geq}. A metric $g$ which solves this equation does not necessarily exist. However, we can introduce the Moore-Penrose pseudoinverse $J^+ \equiv (J^TJ)^{-1}J^T$ of $J$ which satisfies $J^+J=\mathds{1}$. The inverse in this equation is well-defined as long as the Jacobi matrix $J$ has maximal rank. This is (generically) the case when the number $|\mathcal{I}|={\rm dim}(\mathcal{W})$ of Hilbert space modes is taken to be greater than or equal to the number $N$ of neural network parameters. Multiplying \eqref{geq} with $J^+$ and $(J^+)^T$ from the left and the right, respectively, leads to
\begin{equation}\label{g}
 g_{ij}(\theta)=(J^TJ)_{ij}=\sum_{I\in\mathcal{I}}\frac{\partial a_I}{\partial\theta^i}(\theta)\frac{\partial a_I}{\partial\theta^j}(\theta)=\left\langle\frac{\partial f_\theta}{\partial\theta^i},\frac{\partial f_\theta}{\partial\theta^j}\right\rangle\; .
 \end{equation}
As mentioned, this metric does not necessarily solve \eqref{geq} but it provides the best approximation to a solution in the sense that the (mean square of the) difference between the LHS and RHS of \eqref{geq} is minimal. The metric $g$ in \eqref{g} has a beautiful geometrical interpretation: it is the pull-back of the natural Hilbert space metric induced by the scalar product $\langle\cdot,\cdot\rangle$ to the loss landscape $\mathcal{L}\subset\mathcal{H}$. When used in the metric gradient descent~\eqref{gdg} it is expected to lead to exponentially efficient training. For simple neural networks $f_\theta$ (such as perceptrons) and a choice of Hilbert space $g$ can be computed analytically by inserting $f_\theta$ into the RHS in \eqref{g} and carrying out the scalar product. For more complicated neural networks this is difficult to do and a data-driven approach may be more appropriate.\\[2mm]
We introduce data $\mathcal D =(x_\beta,y_\beta)_{\beta=1,\ldots ,|\mathcal D|}$, where $x_\beta\in\mathbb{R}^d$ and $y_\beta\in\mathbb{R}$, batches $B\subset\{1,\ldots ,|\mathcal D|\}$ and assume that the $x_\beta$ are distributed according to a data measure $\rho$. For this data and $\alpha,\beta\in B$ we define
\begin{equation}\label{MD}
M_{\beta i}=\frac{\partial f_\theta}{\partial \theta^i}(x_\beta)\;,\qquad \mathcal{R}_\beta=f_\theta(x_\beta)-y_\beta\;,\qquad
\mu_{\alpha\beta}=\delta_{\alpha\beta}\frac{w(x_\beta)}{\rho(x_\beta)}\; ,
\end{equation}
that is, the $|B|\times n$ matrix $M$ which represents the Jacobi matrix evaluated on the data, the neural network residual $\mathcal{R}$ and the relative measure $\mu$. With these quantities we can approximate
\begin{equation}
 \left(\left\langle\frac{\partial f_\theta}{\partial \theta^i},f_\theta-f\right\rangle\right)\simeq M^T\mu\mathcal{R}\;,\qquad
 g\simeq M^T\mu M \end{equation}
and, assuming that the data measure $\rho$ equals the Hilbert space measure $w$, so that $\mu=\mathds{1}$, the metric gradient descent~\eqref{gdtheta} can be written in the form
\begin{equation}\label{gdnum}
 \gamma\,\delta\theta=-\eta M^T\mathcal{R}\quad\mbox{where}\quad \gamma=M^T M \; .
\end{equation} 
If the numerical metric $\gamma$ is invertible this is equivalent to
\begin{equation}\label{gdnum2}
 \delta\theta=-\eta\, M^+\mathcal{R}\quad\mbox{where}\quad M^+=\gamma^{-1}M^T=(M^TM)^{-1}M^T
\end{equation}
is the Moore-Penrose pseudoinverse of $M$. If the batch size $|B|$ is greater than or equal to the number of parameters $N$ then $\gamma$ is generically invertible and \eqref{gdnum2} is well-defined. Even in this case the metric might have small eigenvalues which lead to unwanted large  excursions in parameter space. In the opposite case, when $|B|<N$, the metric $\gamma$ is necessarily singular. In either case, we can still make sense of \eqref{gdnum2} and regularize the singularities (or small eigenvalues) by writing down an approximation $M\simeq U\sigma V^T$ for $M$, based on a singular value decomposition which only keeps the singular values $\sigma={\rm diag}(\sigma_1,\ldots ,\sigma_k)$ larger than a certain given fraction of the maximal singular value $\sigma_1$. In other words, we keep all singular values $\sigma_i$, where $i=1,\ldots ,k$, for which $\sigma_i/\sigma_1\geq r$ for some $r\in(0,1)$. Then the numerical metric can be approximated by $\gamma\simeq V\sigma^2 V^T$ and the Moore-Penrose inverse of $M$ by $M^+\simeq V\sigma^{-1}U^T$. With these approximations the gradient descent \eqref{gdnum2} can be re-written as
\begin{equation}\label{gdsvd}
\delta\theta\simeq - V\sigma^{-1}U^T\mathcal{R}\; .
\end{equation}
This equation can be used in practice to realize a metric-based gradient descent, and is precisely the update rule implemented in Algorithm~\ref{alg:update}.\\[2mm]
The above argument has been carried out for a mean square loss function but it can be generalized to loss functions of the form $L(\theta) = \frac{1}{2}||F(f_\theta, f)||^2$ for some function $F$. (For the above case of a mean square loss we have $F(f_\theta ,f)=f_\theta -f$. For a cross-entropy loss we should take $F(f_\theta,f)=\sqrt{-f\ln\circ f_\theta}$.  In general, this is the $\kappa = 1$ version of \eqref{eqn:kappa_generalization}.) The gradient descent equation can then be written in the form
\begin{equation}\label{gdgen}
    \delta\theta^i=-\eta\,g^{ij}\left\langle\frac{\partial F(f_\theta ,f)}{\partial\theta^j},F(f_\theta,f)\right\rangle
    \simeq -\eta\, g^{ij}\sum_{I\in\mathcal{I}}\frac{\partial c_I}{\partial\theta^j}(\theta) c_I(\theta)
\end{equation}
where
\begin{equation}
c_I(\theta) =\langle h_I,F(f_\theta ,f)\rangle\; .
\end{equation}
Translating the training trajectory described by \eqref{gdgen} to the coefficients $c_I$ leads to
\begin{equation}
 \delta c_I=-\eta\, G^{IJ}c_J\quad\mbox{where}\quad G^{-1}=Jg^{-1}J^T\quad\mbox{and}\quad J=\frac{\partial c}{\partial\theta}\; .
\end{equation}
If we choose the natural metric $G_{IJ}=\delta_{IJ}$ the coefficients $c_I$ approach zero at an exponential rate, indicating efficient training. Finding the metric $g$ closest to solving $Jg^{-1}J^T=\mathds{1}$ proceeds as above, by using the Moore-Penrose pseudoinverse of $J$, and results in
\begin{equation}\label{ggen}
 g_{ij}=(J^TJ)_{ij}=\sum_I \frac{\partial c_I}{\partial\theta^i}\frac{\partial c_I}{\partial\theta^j}=\left\langle\frac{\partial F(f_\theta,f)}{\partial \theta^i},\frac{\partial F(f_\theta ,f)}{\partial\theta^j}\right\rangle\; .
\end{equation}

As before, this metric can be calculated analytically for simple neural networks by evaluating the RHS of \eqref{ggen}. For a  data-driven approach we define the quantities analogous to the ones in \eqref{MD} by
\begin{equation}\label{MDgen}
 M_{\beta i}=\partial_1 F(f_\theta(x_\beta),y_\beta)\,\frac{\partial f_\theta}{\partial\theta^i}(x_\beta)\;,\quad
 \mathcal{R}_\beta =F(f_\theta (x_\beta),y_\beta)\;,\quad \mu_{\alpha\beta}=\delta_{\alpha\beta}\frac{w(x_\beta)}{\rho(x_\beta)}\;,
\end{equation}
where $(x_\beta,y_\beta)$ is the data, with a distribution $\rho$ for the $x_\beta$ values. Assuming $w=\rho$, as before, we find the numerical version $\gamma=M^TM$ of the metric $g$ and the resulting gradient descent equation has the same form as \eqref{gdnum2}, however with $M$ and $\mathcal{R}$ now defined as in \eqref{MDgen}. Possible singularities in this equation can be regularized as before, by using an approximation for $M$ based on its singular value decomposition but with only the largest singular values kept. The gradient descent then takes the same form as in \eqref{gdsvd} but the singular value decomposition should now be applied to $M$ as defined in~\eqref{MDgen}.

\ifdefined\svntblbox\else\newsavebox{\svntblbox}\fi
\providecommand{\svnfittable}[1]{%
  \sbox\svntblbox{#1}%
  \ifdim\wd\svntblbox>\linewidth
    \resizebox{\linewidth}{!}{\usebox\svntblbox}%
  \else
    \usebox\svntblbox
  \fi}

\section{Experiment Details}
\label{app:exp_details}

In the subsections below, we provide detailed information about experiments performed for the results presented in this paper.
All results come from a campaign of training runs performed over six datasets and hyperparameter grids for each optimizer. We also perform three result-dependent follow-up runs to measure unbiased wall-times,\footnote{Multiple models were trained simultaneously per GPU to optimize resource usage during the hyperparameter grid scans. This artificially increases measured wall-time, so we re-run standalone at the best-performing parameter settings to measure timing.} optimizer diagnostics, and confirmation runs. We run five model seeds per grid point for most experiments, limiting to three in the CIFAR-10 parameter-fraction study, and one in the GPT-2-small comparison, whose per-run cost (Appendix~\ref{app:transformers}) rules out seed-averaging. On the seven headline grids the configuration selected for each optimizer is re-run on five fresh seeds, and those confirmation runs are the numbers reported throughout the paper.
Code for the Sven optimizer and all experiments is available at the links in Appendix~\ref{sec:code}.

\subsection{Datasets and Splits}
\label{app:datasets}

Each dataset exposes three fixed, disjoint splits---train, validation and test---of the sizes in Table~\ref{tab:splits}. The splits are identical for every optimizer, are fixed by a split seed that is independent of the model and data-loading seeds, and do not move when the training set is subsampled: the overparameterization study of Appendix~\ref{app:overparam} varies $N$ by subsampling the \emph{training} part only, so the held-out data and the target scale are the same in every arm. Hyperparameters are selected based on validation loss.
For the synthetic datasets (1D regression and random polynomials), the data are generated in a reproducible fashion using configurable seeds.

\paragraph{One-dimensional regression} We sample train/val/test data from the function
\begin{equation}
    f(x) = e^{-10x^2}\sin(2x),
\end{equation}
with inputs drawn from $U[-1,1]$. The training pool, the validation set and the test set are drawn from three independent generators, and the \emph{targets} of all three are standardized by the mean and standard deviation of the training pool.

\paragraph{Random polynomials}
We fit a random polynomial over $\mathbb{R}^6$ built from all monomials of total degree at most four,
\begin{equation}
    f(\mathbf{x}) = \sum_{\mathbf{d}} c_{\mathbf{d}}\, x_1^{d_1}\cdots x_6^{d_6},
    \qquad \mathbf{d}\in\mathbb{N}^6,\quad \textstyle\sum_{i} d_i \le 4,
\end{equation}
which is $\binom{6+4}{4} = 210$ terms including the constant and the linear terms. Inputs are drawn from $\mathcal{N}(0, I_6)$, and each coefficient is drawn as
\begin{equation}
    c_{\mathbf{d}} \sim \mathcal{N}(0,1) \Big/ \sqrt{\mathbb{E}\big[m_{\mathbf{d}}(\mathbf{x})^2\big]},
    \qquad
    \mathbb{E}\big[m_{\mathbf{d}}^2\big] = \prod_{j=1}^{6} (2 d_j - 1)!!,
\end{equation}
so that every monomial contributes the same variance and no single degree dominates the target. As for the 1D task, pool, validation and test inputs come from separate generators and the targets are standardized by the pool statistics. The coefficient draw is fixed by the data seed; the two synthetic tasks are replicated over three data seeds in the confirmation pass, so that between-instance variation is reported separately from seed variation.

\paragraph{MNIST} We use MNIST~\citep{lecun1998mnist} with the official training set split $50{,}000/10{,}000$ into train and validation by a fixed split seed, and the official $10{,}000$-image test set held out. Images are standardized with the published statistics and flattened.

\paragraph{CIFAR-10} In Appendix \ref{app:cifar_10} we present results from training a ResNet~\citep{he2016deep} on the CIFAR-10 image dataset~\citep{krizhevsky2009learning}. The official training set is split $45{,}000/5{,}000$ into train and validation by the same fixed split seed and the official $10{,}000$-image test set is held out. Images are standardized with the published channel statistics.

\paragraph{Character-level Shakespeare}
The tiny-shakespeare corpus of nanoGPT~\citep{karpathy2022nanogpt} is encoded at character level over a $65$-symbol vocabulary and split $80/10/10$ \emph{contiguously by position}, train first, so that no sequence spans a split boundary. Each part is then chunked independently into non-overlapping blocks of $128$ characters, with the target the input shifted by one. Results are in Appendix~\ref{app:transformers}.

\paragraph{FineWeb-edu (GPT-2 tokens)}
For the GPT-2-small~\citep{radford2019gpt2} comparison we use FineWeb-edu~\citep{penedo2024fineweb} tokenized with the GPT-2 BPE vocabulary (padded to $50{,}304$) and served as non-overlapping blocks of $1{,}024$ tokens. Validation and test blocks are drawn from documents \emph{disjoint} from the training documents. One pass over the training blocks is the token budget of a run. Results are in Appendix~\ref{app:transformers}.

\begin{table}[t]
  \centering
  \small
  \setlength{\tabcolsep}{4pt}
  \caption{Dataset splits. Sizes are examples, or blocks for the two language-modeling corpora.}
  \label{tab:splits}
  \begin{tabular}{llrrrl}
    \toprule
    Dataset & Loss & Train & Val. & Test & Construction \\
    \midrule
    1D regression      & $L_2$          & 10k  &  10k & 10k & independent draws; pool-standardized \\
    Random polynomial  & $L_2$          & 10k  & 10k & 10k & independent draws; pool-standardized \\
    MNIST              & label reg./CE  & 50k  & 10k & 10k & official train split; official test set \\
    CIFAR-10           & label reg./CE  & 45k  &  5k & 10k & official train split; official test set \\
    Shakespeare (char) & token CE       &  6,971  &    871 &    871 & contiguous 80/10/10 by position \\
    FineWeb-edu (BPE)  & token CE       & 210k &    200 &    200 & val/test from disjoint documents \\
    \bottomrule
  \end{tabular}
\end{table}

\subsection{Model Architectures}
\label{app:models}

We use MLPs with three hidden layers and GeLU activations~\citep{hendrycks2016gaussian} for the regression and MNIST experiments. The hidden layer width is set to 16 for the 1D and polynomial datasets, and 32 for MNIST. For the CIFAR-10 experiments in Appendix~\ref{app:cifar_10} we use the \texttt{ResNet18}~\citep{he2016deep} architecture provided in \texttt{torchvision}~\citep{anselpytorch,torchvision_software}, with every \texttt{BatchNorm2d} layer replaced by a drop-in variant that does not write its running statistics in place; the numerics are unchanged, but the forward pass then composes with the \texttt{torch.func} transforms that Sven's Jacobian capture uses (Appendix~\ref{app:gram}). The two language-modeling experiments use a decoder-only transformer of the nanoGPT form with untied embeddings and no dropout. Parameter counts are listed in Table~\ref{tab:models}.

\begin{table}[t]
  \centering
  \small
  \caption{Models and parameter counts $P$. The MLP notation lists input, hidden and output widths.}
  \label{tab:models}
  \begin{tabular}{lll r}
    \toprule
    Task & Model & Configuration & $P$ \\
    \midrule
    1D regression      & MLP  & $1\!-\!16\!-\!16\!-\!16\!-\!1$, GeLU & 593 \\
    Random polynomial  & MLP  & $6\!-\!16\!-\!16\!-\!16\!-\!1$, GeLU & 673 \\
    MNIST              & MLP  & $784\!-\!32\!-\!32\!-\!32\!-\!10$, GeLU & 27,562 \\
    CIFAR-10           & ResNet18 & 10 classes, functional BatchNorm & 11,181,642 \\
    Shakespeare (char) & nanoGPT  & 4 layers, 4 heads, $d=128$, block 128 & 826,368 \\
    FineWeb-edu        & GPT-2 small & 12 layers, 12 heads, $d=768$, block 1024 & 163,109,376 \\
    \bottomrule
  \end{tabular}
\end{table}

\subsection{Evaluation Protocol}
\label{app:protocol}

\paragraph{Model Evaluation} Validation, test and train losses are evaluated before training and after each epoch, with a fixed subset of the training data used to evaluate train loss. The reported loss is a mean over examples (over tokens for the language models), not a mean of per-batch means. Accuracy is the fraction of correct \texttt{argmax} predictions over the raw model outputs. 
The train loss subset consists of at most $10{,}000$ examples ($1{,}000$ on nanoGPT, $200$ blocks on GPT-2-small) shared by all optimizers in a scan. The evaluation batch size is a separate setting and is never the training batch size. The GPT-2-small comparison is a single $13{,}125$-step epoch, so it is evaluated every $500$ steps instead of every epoch.

\paragraph{Diverged Runs} A run fails if it ends with a non-finite training or validation loss, if it stops early on a non-finite training batch, or if its final validation loss exceeds ten times the loss of the untrained model. Failed runs are tracked and excluded from seed means, and tables print the number of runs that finished out of the number attempted. Appendix~\ref{app:robustness} gives the full accounting per scan and per method.

\paragraph{Selection.} For each optimizer and each scan we select one configuration by a fixed rule applied to validation loss only: a configuration is eligible if more than half of its seeds finished; among eligible configurations we take the one with the fewest failed seeds, breaking ties by the lowest seed-mean final validation loss.

\paragraph{Confirmation seeds.} For each of the headline scans, the selected configuration of every optimizer is re-run on five model seeds disjoint from the five it was tuned on, and the two synthetic tasks (1D and polynomial regression) are additionally replicated over three independent data draws. These confirmation runs are the numbers reported in the paper; the tuning-seed values are shown beside them, so that the optimism introduced by selection is visible rather than absorbed.

\paragraph{Paired comparisons.} Within a scan, a model seed fixes both the initialization and the data order, so every optimizer at a given seed sees the same initial weights and the same sequence of batches. The uncertainty band drawn in figures is the mean $\pm$ one standard deviation over seeds.

\paragraph{Normalization layers.} ResNet18 uses running statistics in its BatchNorm layers~\citep{ioffe2015batch}. These are updated once per optimizer step over the course of training during all CIFAR-10 scans, then frozen for evaluation. Sven's Jacobian capture makes several passes over the batch per step, so we run them in a context that suppresses updates without switching the layer to eval mode (which would normalize with the running statistics instead and change the Jacobian). We also lightly modify the PyTorch \texttt{BatchNorm} implementation to avoid in-place tensor updates that are invalid inside the functional API \texttt{torch.func}, which is used for full-Jacobian capture (see Appendix \ref{app:gram} for full details of Sven's implementation).

\paragraph{Epochs.} The four MLP scans and both CIFAR-10 scans train for 20 epochs, which was found to be sufficient for convergence in validation loss; nanoGPT trains for 50, the two full-batch overparameterization scans for 200, and GPT-2-small for a single epoch of $13{,}125$ steps. Table~\ref{tab:protocol} lists the protocol of the seven headline scans together with the number of confirmation runs that finished. The studies of Table~\ref{tab:sven-ablation-grids} run five model seeds and their parent task's 20 epochs -- three seeds for the CIFAR-10 parameter-fraction study, 200 epochs for the two full-batch $P/N$ sweeps -- and use no dedicated confirmation pass.

\begin{table}[t]
  \centering
  \caption{Summary of the hyperparameter scans. ``Seeds'' indicates the five tuning seeds plus the five fresh confirmation seeds, with the number of independent data draws for the two synthetic tasks; ``methods'' indicates the number of optimizers used; ``grid runs'' is the total number of runs across optimizers, seeds, and hyperparameters; and the last column is the fraction of non-diverged confirmation runs.}
  \label{tab:protocol}
  \svnfittable{\scriptsize\setlength{\tabcolsep}{2pt}\resizebox{\linewidth}{!}{%
\begin{tabular}{llccccccc}
\toprule
\textbf{Task} & \textbf{Loss} & \textbf{$P$} & \textbf{$B$} & \textbf{Epochs} & \textbf{Seeds} & \textbf{Methods} & \textbf{Grid runs} & \textbf{Confirm fin./att.} \\
\midrule
Toy 1D & squared error & 593 & 32 & 20 & 5+5$\times3$ & 15 & 1770 & 224/225 \\
Random Polynomial & squared error & 673 & 32 & 20 & 5+5$\times3$ & 15 & 1630 & 212/225 \\
MNIST (label reg.) & label regression & 27562 & 64 & 20 & 5+5 & 14 & 1360 & 70/70 \\
MNIST (CE) & cross-entropy & 27562 & 64 & 20 & 5+5 & 14 & 1610 & 68/70 \\
CIFAR-10 (label reg.) & label regression & 11181642 & 128 & 20 & 5+5 & 11 & 600 & 55/55 \\
CIFAR-10 (CE) & cross-entropy & 11181642 & 128 & 20 & 5+5 & 11 & 785 & 55/55 \\
nanoGPT (tiny-shakespeare) & token cross-entropy & 826368 & 64 & 50 & 5+5 & 5 & 140 & 25/25 \\
\bottomrule
\end{tabular}
}
}
\end{table}

\subsection{Hyperparameter Grids}
\label{app:grids}

We perform hyperparameter scans to find good working points for Sven and the other baseline optimizers.
Tables~\ref{tab:sven-grids} and~\ref{tab:sven-ablation-grids} give Sven's grids, Table~\ref{tab:baseline-grids} the baselines' learning-rate grids, and Table~\ref{tab:baseline-settings} the settings that are held fixed. Weight decay is not swept: AdamW uses its default of $0.01$, and MuonW\footnote{We distinguish ``Muon'' and ``MuonW'' even though Muon has nonzero weight-decay by default. Sven uses no regularization, so we focus on non-regularized baselines.} its default of $0.1$. The selected best-performing configuration of each optimizer is reported in Tables~\ref{tab:grid-toy}--\ref{tab:grid-nanogpt} of Section~\ref{app:grids_as_run}.
In all cases we use Sven with $\kappa = 2$, treating the raw losses as the ``residuals'' (see Sec.~\ref{sec:lstsqr_kappa}). Appendix~\ref{app:kappaScans} explores other settings of $\kappa$.
 Furthermore, Sven's $k$ and \texttt{rtol} are not independent: the rank actually used on a step is the smaller of $k$ and the number of singular values above $\texttt{rtol}\cdot\sigma_{\max}$. We sweep both, but depending on \texttt{rtol} and the shape of the SV spectrum for a given problem, fewer than $k$ may actually be used for Sven's parameter updates.

\begin{table}[t]
  \centering
  \small
  \caption{Sven's hyperparameter grids. $B$ is the training batch size, and ``runs'' counts the full grid including seeds. All grids use $\kappa=2$. The CIFAR-10 cross-entropy grid was extended at $k=128$, $\eta\in\{0.05,0.1,0.5\}$ with $\texttt{rtol}\in\{0.03,0.1,0.3\}$, because the optimum sat on the \texttt{rtol} edge.}
  \label{tab:sven-grids}
  \begin{tabular}{lllll r}
    \toprule
    Scan & $B$ & $k$ & $\eta$ & \texttt{rtol} & Runs \\
    \midrule
    1D regression      & 32  & 1,2,4,8,16,32    & .01,.02,.05,.1,.5,1 & $10^{-6}\!-\!10^{-2}$ (5) & 900 \\
    Random polynomial  & 32  & 1,2,4,8,16,32    & .05,.1,.5,1 & $10^{-5}\!-\!10^{-1}$ (6) & 720 \\
    MNIST label reg.   & 64  & 1,2,4,8,16,32,48,64 & .05,.1,.5,1 & $10^{-4}\!-\!10^{-1}$ (4) & 640 \\
    MNIST CE           & 64  & 1,2,4,8,16,32,48,64 & .05,.1,.5,1 & $10^{-4}\!-\!0.3$ (5)     & 800 \\
    CIFAR-10 label reg.& 128 & 64,128           & .1,.5,1     & $10^{-4},10^{-3},10^{-2}$ & 90 \\
    CIFAR-10 CE        & 128 & 64,128           & .02,.05,.1,.5,1 & $10^{-4},10^{-3},10^{-2}$ & 150 \\
    nanoGPT            & 64  & 64               & .05,.1,.5,1 & $10^{-3}$ & 20 \\
    GPT-2 small        & 16  & 16               & .02,.05,.1,.5,1 & $10^{-3}$ & 5 \\
    \bottomrule
  \end{tabular}
\end{table}

\begin{table}[t]
  \centering
  \footnotesize
  \setlength{\tabcolsep}{3pt}
  \caption{Sven's grids in follow-up studies varying $\kappa$, (micro-)batch size ($\mu$), updated parameter fraction ($f$), and overparametrization ($P/N$). For MNIST, the two \texttt{rtol} values separated by a slash are the label-regression and cross-entropy settings. The $P/N$ sweeps on the two synthetic tasks are full-batch ($B = N$).}
  \label{tab:sven-ablation-grids}
  \begin{tabular}{llllll r}
    \toprule
    Study & $B$ & $k$ & $\eta$ & \texttt{rtol} & Swept axis & Runs \\
    \midrule
    $\kappa$ (MNIST lab.\ reg.) & 64 & 32,64 & .125--1.5 (7) & $10^{-4}$ & $\kappa$: 1,2,3 & 210 \\
    Micro-batch (1D, poly.)   & 32 & 32 & .05,.1,.5,1 & $10^{-3}$ & $\mu$: 1--32 (6) & 120 \\
    Micro-batch (MNIST)       & 64 & 64 & .05,.1,.5,1 & $10^{-4}/10^{-1}$ & $\mu$: 1--64 (7) & 140 \\
    Param.\ frac.\ (1D, poly.)& 32 & 32 & .05,.1,.5,1 & $10^{-3}$ & $f$: .1--1 (5) & 100 \\
    Param.\ frac.\ (MNIST)    & 64 & 64 & .05,.1,.5,1 & $10^{-4}/10^{-1}$ & $f$: .1--1 (5) & 100 \\
    Param.\ frac.\ (CIFAR-10) & 128 & 128 & .5 & $10^{-3}$ & $f$: .05--1 (5) & 15 \\
    $P/N$ (1D)                & $N$ & $.25B,.5B,B$ & .05,.1,.5,1 & $10^{-4}\!-\!10^{-2}$ (3) & $N$: 150--1200 (4) & 720 \\
    $P/N$ (polynomial)        & $N$ & $.25B,.5B,B$ & .01--1 (6) & $10^{-4}\!-\!10^{-1}$ (5) & $N$: 170--1350 (4) & 1800 \\
    $P/N$ (MNIST)             & 64 & 16,32,48,64 & .1,.5,1 & $10^{-4}$ & $N$: 2500--50000 (6) & 360 \\
    Batch size (polynomial)   & $B$ & $k=B$ & .05,.1,.5,1 & $10^{-5}\!-\!10^{-2}$ (4) & $B$: 8--256 (6) & 480 \\
    \bottomrule
  \end{tabular}
\end{table}

\begin{table}[t]
  \centering
  \small
  \caption{Baseline learning-rate grids. The second column is the shared grid used by all standard optimizers; L-BFGS, JD and HIG have their own. L-BFGS additionally sweeps \texttt{max\_iter} $\in\{1,2,3\}$ and \texttt{history\_size} $\in\{2,5,10\}$; HIG additionally sweeps its cutoff $\tau \in \{10^{-8},10^{-6},10^{-4},10^{-2},3\!\cdot\!10^{-2},10^{-1}\}$. Polyak SGD has no learning rate to sweep. A dash means the method was not run on that scan.}
  \label{tab:baseline-grids}
  \begin{tabular}{lllll}
    \toprule
    Scan & Standard optimizers & L-BFGS & JD & HIG \\
    \midrule
    1D regression       & $10^{-5}\!-\!1$ (8) & .01,.03,.1,.5,1 & $10^{-5}\!-\!10^{-1}$ (6) & $5\!\cdot\!10^{-4}\!-\!0.25$ (7) \\
    Random polynomial   & $10^{-6}\!-\!1$ (10)& .1,.5,1,2,4     & $10^{-5}\!-\!10^{-1}$ (6) & $5\!\cdot\!10^{-3}\!-\!0.25$ (5) \\
    MNIST label reg.    & $10^{-5}\!-\!1$ (8) & .1,.5,1         & $10^{-5}\!-\!10^{-1}$ (6) & $5\!\cdot\!10^{-3}\!-\!0.25$ (5) \\
    MNIST CE            & $10^{-5}\!-\!1$ (8) & .1,.5,1,2,4     & $10^{-5}\!-\!10^{-1}$ (6) & $5\!\cdot\!10^{-3}\!-\!0.25$ (5) \\
    CIFAR-10 label reg. & $10^{-5}\!-\!0.3$ (7)& .1,.5,1,2,4    & ---                       & --- \\
    CIFAR-10 CE         & $10^{-5}\!-\!3$ (9) & .1,.5,1,2,4     & ---                       & --- \\
    nanoGPT             & $10^{-5}\!-\!3\!\cdot\!10^{-3}$ (6) & --- & ---                   & --- \\
    GPT-2 small         & $10^{-5}\!-\!3\!\cdot\!10^{-3}$ (6) & --- & ---                   & --- \\
    \bottomrule
  \end{tabular}
\end{table}

\begin{table}[t]
  \centering
  \small
  \caption{Baseline optimizers and the settings held fixed.}
  \label{tab:baseline-settings}
  \svnfittable{
  \begin{tabular}{ll}
    \toprule
    Optimizer & Fixed settings \\
    \midrule
    Adam~\citep{kingma2015adam}         & PyTorch defaults, weight decay 0 \\
    AdamW~\citep{loshchilov2019adamw}                              & PyTorch defaults, weight decay 0.01 \\
    SGD                                & no momentum \\
    SGD + momentum                     & momentum 0.9 \\
    RMSprop~\citep{hinton2012rmsprop}   & PyTorch defaults \\
    Muon~\citep{jordan2024muon}         & hidden projections only, weight decay 0 (see below) \\
    MuonW                              & as Muon, weight decay 0.1 (0.01 on the AdamW group) \\
    SOAP~\citep{vyas2025soap}           & $\beta=(0.95,0.95)$, preconditioning every 10 steps \\
    Shampoo~\citep{gupta2018shampoo}    & update frequency 1, $\epsilon = 10^{-4}$ \\
    K-FAC~\citep{martens2015kfac}       & SGD + momentum 0.9 base, damping $3\cdot10^{-3}$, factors updated every step \\
    \midrule
    Stochastic L-BFGS~\citep{liu1989limited}                  & strong-Wolfe line search; \texttt{max\_iter} and \texttt{history\_size} swept \\
    Polyak SGD~\citep{polyak1969minimization}  & $\eta = (L - f^\star)/(\|\nabla L\|^2 + \epsilon)$, $f^\star = 0$, capped at 1 \\
    JD (UPGrad)~\citep{quinton2024jacobian}    & UPGrad aggregator, Adam inner optimizer \\
    HIG~\citep{schnell2022hig}          & $\Sigma^{-1/2}$ of the stacked output Jacobian; cutoff $\tau$ swept \\
    \bottomrule
  \end{tabular}
  }
\end{table}

\paragraph{Muon parameter grouping} Muon uses a subset of parameters in its orthogonalized updates, with the remaining ones optimized by AdamW. In our configuration, we send embedding layers, final linear/convolutional layers, biases, and normalization weights to AdamW. All other linear/convolutional weights use Muon, with convolution kernels flattened to two dimensions. The single shared learning rate is made comparable across the two groups by the RMS-matching rule~\citep{liu2025muon}, which scales the orthogonalized step to the size of an AdamW step.

\paragraph{HIG computational cost} HIG computes the SVD of the stacked \emph{model} Jacobian, whose row count is the batch size times the output dimension. Its Jacobian is therefore ten times taller than Sven's on MNIST and as tall as the vocabulary on a language model. This becomes intractable for larger models such as transformers or ResNets, and HIG is thus omitted from those scans.

\paragraph{Grids in follow-up studies} The overparametrization and batch size sweeps use the same learning-rate grid for all ten standard optimizers, together with Polyak SGD and stochastic L-BFGS. L-BFGS sweeps learning rates against the same \texttt{max\_iter} and \texttt{history\_size} grids as before, except in the batch-size sweep where its shape is pinned at \texttt{max\_iter}$\,=3$, \texttt{history\_size}$\,=2$. JD and HIG are run on the four MLP headline scans only; the $\kappa$, micro-batch and parameter-fraction studies only use Sven.

\subsection{Compute Resources and Wall-Time Measurement}
\label{app:compute}

All runs were performed on either 80GB NVIDIA A100 GPUs or 20GB MIG slices of 80GB A100s. The MLP scans used small enough models that several could be run on a single device, while the CIFAR-10, nanoGPT and GPT-2 runs used a full 80GB A100. 
Per-step and per-epoch memory and wall-time costs for every method and architecture are in Appendix~\ref{app:profiling}.

Wall-clock numbers reported in the paper are not taken from the grid scans, because several processes shared a device to optimize utilization and thus increased step time relative to a standalone run. Instead, each scan has a dedicated timing run using the selected best configuration of each optimizer. These are run one process at a time on a single device, with checkpointing and diagnostic logging switched off. 
Timing runs reproduce the grid runs' hyperparameters exactly and are used for step and epoch times only; loss values come from the grid or confirmation passes.

\subsection{Full Grid Summary}
\label{app:grids_as_run}

Table~\ref{tab:gridsizes} counts the distinct configurations of each method on each scan. Tables~\ref{tab:grid-toy}, \ref{tab:grid-poly}, \ref{tab:grid-mnist}, \ref{tab:grid-nanogpt}, and~\ref{tab:grid-cifar} fully enumerate the grids for each scan, and indicate the selected best-performing configuration measured by final validation loss. A selected value carrying $\uparrow$ ($\downarrow$) is the largest (smallest) value that axis was swept over, i.e.\ a grid edge, and indicates where results may be bounded by the grid.

\begin{table}[t]
  \centering
  \caption{Number of distinct configurations of each method on each scan (excluding seed duplicates); a blank means the method was not run. Sven sweeps $(k, \eta, \texttt{rtol})$ where most baselines sweep a learning rate alone.}
  \label{tab:gridsizes}
  \svnfittable{\resizebox{\linewidth}{!}{%
\begin{tabular}{lrrrrrrr}
\toprule
\textbf{Method} & \textbf{1D reg.} & \textbf{Poly.} & \textbf{MNIST-LR} & \textbf{MNIST-CE} & \textbf{CIFAR-LR} & \textbf{CIFAR-CE} & \textbf{nanoGPT} \\
\midrule
Sven & 180 & 144 & 128 & 160 & 18 & 39 & 4 \\
Adam & 8 & 10 & 8 & 8 & 7 & 9 & -- \\
AdamW & 8 & 10 & 8 & 8 & 7 & 9 & 6 \\
HIG & 42 & 30 & 30 & 30 & -- & -- & -- \\
JD (UPGrad) & 6 & 6 & 6 & 6 & -- & -- & -- \\
KFAC & 8 & 10 & -- & -- & -- & -- & -- \\
Stochastic L-BFGS & 45 & 45 & 27 & 45 & 45 & 45 & -- \\
Muon & 8 & 10 & 8 & 8 & 7 & 9 & 6 \\
MuonW & 8 & 10 & 8 & 8 & 7 & 9 & 6 \\
Polyak SGD & 1 & 1 & 1 & 1 & 1 & 1 & -- \\
RMSprop & 8 & 10 & 8 & 8 & 7 & 9 & -- \\
SGD & 8 & 10 & 8 & 8 & 7 & 9 & -- \\
SGD + momentum & 8 & 10 & 8 & 8 & 7 & 9 & -- \\
SOAP & 8 & 10 & 8 & 8 & 7 & 9 & 6 \\
Shampoo & 8 & 10 & 8 & 8 & -- & -- & -- \\
\bottomrule
\end{tabular}
}
}
\end{table}

\begin{table}[t]
  \centering
  \caption{Hyperparameter grid for the 1D regression scan, with the best-performing configuration indicated. Weight decay is at each optimizer's default.}
  \label{tab:grid-toy}
  \svnfittable{\scriptsize\resizebox{\linewidth}{!}{%
\begin{tabular}{lp{2.2in}rrp{1.1in}}
\toprule
\textbf{Method} & \parbox[t]{\linewidth}{\raggedright \textbf{Grid swept}\strut} & \textbf{Configs} & \textbf{Runs} & \parbox[t]{\linewidth}{\raggedright \textbf{Selected}\strut} \\
\midrule
Sven & \parbox[t]{\linewidth}{\raggedright $\eta$ $\in$ {0.01, 0.02, 0.05, 0.1, 0.5, 1}; $k$ $\in$ {1, 2, 4, 8, 16, 32}; rtol $\in$ {1e-06, 1e-05, 0.0001, 0.001, 0.01}\strut} & 180 & 900 & \parbox[t]{\linewidth}{\raggedright k=32$^{\uparrow}$, lr=0.01$^{\downarrow}$, rtol=0.0001\strut} \\
Adam & \parbox[t]{\linewidth}{\raggedright $\eta$ $\in$ {1e-05, 3e-05, 0.0001, 0.001, 0.01, 0.1, 0.3, 1}\strut} & 8 & 40 & \parbox[t]{\linewidth}{\raggedright lr=0.001\strut} \\
AdamW & \parbox[t]{\linewidth}{\raggedright $\eta$ $\in$ {1e-05, 3e-05, 0.0001, 0.001, 0.01, 0.1, 0.3, 1}\strut} & 8 & 40 & \parbox[t]{\linewidth}{\raggedright lr=0.001, weight\_decay=0.01\strut} \\
HIG & \parbox[t]{\linewidth}{\raggedright $\eta$ $\in$ {0.0005, 0.0015, 0.005, 0.015, 0.05, 0.1, 0.25}; $\tau$ $\in$ {1e-08, 1e-06, 0.0001, 0.01, 0.03, 0.1}\strut} & 42 & 210 & \parbox[t]{\linewidth}{\raggedright lr=0.005, tau=1e-06\strut} \\
JD (UPGrad) & \parbox[t]{\linewidth}{\raggedright $\eta$ $\in$ {1e-05, 3e-05, 0.0001, 0.001, 0.01, 0.1}\strut} & 6 & 30 & \parbox[t]{\linewidth}{\raggedright lr=0.01, aggregator=UPGrad, inner\_optimizer=Adam\strut} \\
KFAC & \parbox[t]{\linewidth}{\raggedright $\eta$ $\in$ {1e-05, 3e-05, 0.0001, 0.001, 0.01, 0.1, 0.3, 1}\strut} & 8 & 40 & \parbox[t]{\linewidth}{\raggedright lr=0.001\strut} \\
Stochastic L-BFGS & \parbox[t]{\linewidth}{\raggedright $\eta$ $\in$ {0.01, 0.03, 0.1, 0.5, 1}; max iter $\in$ {1, 2, 3}; history $\in$ {2, 5, 10}\strut} & 45 & 225 & \parbox[t]{\linewidth}{\raggedright lr=0.03, max\_iter=3, history\_size=10\strut} \\
Muon & \parbox[t]{\linewidth}{\raggedright $\eta$ $\in$ {1e-05, 3e-05, 0.0001, 0.001, 0.01, 0.1, 0.3, 1}\strut} & 8 & 40 & \parbox[t]{\linewidth}{\raggedright lr=0.001\strut} \\
MuonW & \parbox[t]{\linewidth}{\raggedright $\eta$ $\in$ {1e-05, 3e-05, 0.0001, 0.001, 0.01, 0.1, 0.3, 1}\strut} & 8 & 40 & \parbox[t]{\linewidth}{\raggedright lr=0.001, weight\_decay=0.1\strut} \\
Polyak SGD & \parbox[t]{\linewidth}{\raggedright no swept hyperparameter\strut} & 1 & 5 & \parbox[t]{\linewidth}{\raggedright f\_star=0, max\_lr=1, eps=1e-08\strut} \\
RMSprop & \parbox[t]{\linewidth}{\raggedright $\eta$ $\in$ {1e-05, 3e-05, 0.0001, 0.001, 0.01, 0.1, 0.3, 1}\strut} & 8 & 40 & \parbox[t]{\linewidth}{\raggedright lr=0.001\strut} \\
SGD & \parbox[t]{\linewidth}{\raggedright $\eta$ $\in$ {1e-05, 3e-05, 0.0001, 0.001, 0.01, 0.1, 0.3, 1}\strut} & 8 & 40 & \parbox[t]{\linewidth}{\raggedright lr=0.3\strut} \\
SGD + momentum & \parbox[t]{\linewidth}{\raggedright $\eta$ $\in$ {1e-05, 3e-05, 0.0001, 0.001, 0.01, 0.1, 0.3, 1}\strut} & 8 & 40 & \parbox[t]{\linewidth}{\raggedright lr=0.1\strut} \\
SOAP & \parbox[t]{\linewidth}{\raggedright $\eta$ $\in$ {1e-05, 3e-05, 0.0001, 0.001, 0.01, 0.1, 0.3, 1}\strut} & 8 & 40 & \parbox[t]{\linewidth}{\raggedright lr=0.001\strut} \\
Shampoo & \parbox[t]{\linewidth}{\raggedright $\eta$ $\in$ {1e-05, 3e-05, 0.0001, 0.001, 0.01, 0.1, 0.3, 1}\strut} & 8 & 40 & \parbox[t]{\linewidth}{\raggedright lr=0.1\strut} \\
\bottomrule
\end{tabular}
}
}
\end{table}

\begin{table}[!ht]
  \centering
  \caption{Hyperparameter grid for polynomial regression. Conventions as in Table~\ref{tab:grid-toy}.}
  \label{tab:grid-poly}
  \svnfittable{\scriptsize\resizebox{\linewidth}{!}{%
\begin{tabular}{lp{2.2in}rrp{1.1in}}
\toprule
\textbf{Method} & \parbox[t]{\linewidth}{\raggedright \textbf{Grid swept}\strut} & \textbf{Configs} & \textbf{Runs} & \parbox[t]{\linewidth}{\raggedright \textbf{Selected}\strut} \\
\midrule
Sven & \parbox[t]{\linewidth}{\raggedright $\eta$ $\in$ {0.05, 0.1, 0.5, 1}; $k$ $\in$ {1, 2, 4, 8, 16, 32}; rtol $\in$ {1e-05, 0.0001, 0.001, 0.01, 0.03, 0.1}\strut} & 144 & 720 & \parbox[t]{\linewidth}{\raggedright k=16, lr=0.5, rtol=0.03\strut} \\
Adam & \parbox[t]{\linewidth}{\raggedright $\eta$ $\in$ {1e-06, 3e-06, 1e-05, 3e-05, 0.0001, 0.001, 0.01, 0.1, 0.3, 1}\strut} & 10 & 50 & \parbox[t]{\linewidth}{\raggedright lr=0.01\strut} \\
AdamW & \parbox[t]{\linewidth}{\raggedright $\eta$ $\in$ {1e-06, 3e-06, 1e-05, 3e-05, 0.0001, 0.001, 0.01, 0.1, 0.3, 1}\strut} & 10 & 50 & \parbox[t]{\linewidth}{\raggedright lr=0.01, weight\_decay=0.01\strut} \\
HIG & \parbox[t]{\linewidth}{\raggedright $\eta$ $\in$ {0.005, 0.015, 0.05, 0.1, 0.25}; $\tau$ $\in$ {1e-08, 1e-06, 0.0001, 0.01, 0.03, 0.1}\strut} & 30 & 150 & \parbox[t]{\linewidth}{\raggedright lr=0.1, tau=1e-08$^{\downarrow}$\strut} \\
JD (UPGrad) & \parbox[t]{\linewidth}{\raggedright $\eta$ $\in$ {1e-05, 3e-05, 0.0001, 0.001, 0.01, 0.1}\strut} & 6 & 30 & \parbox[t]{\linewidth}{\raggedright lr=0.001, aggregator=UPGrad, inner\_optimizer=Adam\strut} \\
KFAC & \parbox[t]{\linewidth}{\raggedright $\eta$ $\in$ {1e-06, 3e-06, 1e-05, 3e-05, 0.0001, 0.001, 0.01, 0.1, 0.3, 1}\strut} & 10 & 50 & \parbox[t]{\linewidth}{\raggedright lr=1e-05\strut} \\
Stochastic L-BFGS & \parbox[t]{\linewidth}{\raggedright $\eta$ $\in$ {0.1, 0.5, 1, 2, 4}; max iter $\in$ {1, 2, 3}; history $\in$ {2, 5, 10}\strut} & 45 & 225 & \parbox[t]{\linewidth}{\raggedright lr=1, max\_iter=1, history\_size=2\strut} \\
Muon & \parbox[t]{\linewidth}{\raggedright $\eta$ $\in$ {1e-06, 3e-06, 1e-05, 3e-05, 0.0001, 0.001, 0.01, 0.1, 0.3, 1}\strut} & 10 & 50 & \parbox[t]{\linewidth}{\raggedright lr=0.01\strut} \\
MuonW & \parbox[t]{\linewidth}{\raggedright $\eta$ $\in$ {1e-06, 3e-06, 1e-05, 3e-05, 0.0001, 0.001, 0.01, 0.1, 0.3, 1}\strut} & 10 & 50 & \parbox[t]{\linewidth}{\raggedright lr=0.01, weight\_decay=0.1\strut} \\
Polyak SGD & \parbox[t]{\linewidth}{\raggedright no swept hyperparameter\strut} & 1 & 5 & \parbox[t]{\linewidth}{\raggedright f\_star=0, max\_lr=1, eps=1e-08\strut} \\
RMSprop & \parbox[t]{\linewidth}{\raggedright $\eta$ $\in$ {1e-06, 3e-06, 1e-05, 3e-05, 0.0001, 0.001, 0.01, 0.1, 0.3, 1}\strut} & 10 & 50 & \parbox[t]{\linewidth}{\raggedright lr=0.01\strut} \\
SGD & \parbox[t]{\linewidth}{\raggedright $\eta$ $\in$ {1e-06, 3e-06, 1e-05, 3e-05, 0.0001, 0.001, 0.01, 0.1, 0.3, 1}\strut} & 10 & 50 & \parbox[t]{\linewidth}{\raggedright lr=0.1\strut} \\
SGD + momentum & \parbox[t]{\linewidth}{\raggedright $\eta$ $\in$ {1e-06, 3e-06, 1e-05, 3e-05, 0.0001, 0.001, 0.01, 0.1, 0.3, 1}\strut} & 10 & 50 & \parbox[t]{\linewidth}{\raggedright lr=0.01\strut} \\
SOAP & \parbox[t]{\linewidth}{\raggedright $\eta$ $\in$ {1e-06, 3e-06, 1e-05, 3e-05, 0.0001, 0.001, 0.01, 0.1, 0.3, 1}\strut} & 10 & 50 & \parbox[t]{\linewidth}{\raggedright lr=0.01\strut} \\
Shampoo & \parbox[t]{\linewidth}{\raggedright $\eta$ $\in$ {1e-06, 3e-06, 1e-05, 3e-05, 0.0001, 0.001, 0.01, 0.1, 0.3, 1}\strut} & 10 & 50 & \parbox[t]{\linewidth}{\raggedright lr=0.1\strut} \\
\bottomrule
\end{tabular}
}
}
\end{table}

\begin{table}[t]
  \centering
  \caption{Hyperparameter grids for MNIST label regression (top) and cross-entropy (bottom). Conventions as in Table~\ref{tab:grid-toy}.}
  \label{tab:grid-mnist}
  \svnfittable{\scriptsize\resizebox{\linewidth}{!}{%
\begin{tabular}{lp{2.2in}rrp{1.1in}}
\toprule
\textbf{Method} & \parbox[t]{\linewidth}{\raggedright \textbf{Grid swept}\strut} & \textbf{Configs} & \textbf{Runs} & \parbox[t]{\linewidth}{\raggedright \textbf{Selected}\strut} \\
\midrule
Sven & \parbox[t]{\linewidth}{\raggedright $\eta$ $\in$ {0.05, 0.1, 0.5, 1}; $k$ $\in$ {1, 2, 4, 8, 16, 32, 48, 64}; rtol $\in$ {0.0001, 0.001, 0.01, 0.1}\strut} & 128 & 640 & \parbox[t]{\linewidth}{\raggedright k=64$^{\uparrow}$, lr=0.5, rtol=0.001\strut} \\
Adam & \parbox[t]{\linewidth}{\raggedright $\eta$ $\in$ {1e-05, 3e-05, 0.0001, 0.001, 0.01, 0.1, 0.3, 1}\strut} & 8 & 40 & \parbox[t]{\linewidth}{\raggedright lr=0.001\strut} \\
AdamW & \parbox[t]{\linewidth}{\raggedright $\eta$ $\in$ {1e-05, 3e-05, 0.0001, 0.001, 0.01, 0.1, 0.3, 1}\strut} & 8 & 40 & \parbox[t]{\linewidth}{\raggedright lr=0.001, weight\_decay=0.01\strut} \\
HIG & \parbox[t]{\linewidth}{\raggedright $\eta$ $\in$ {0.005, 0.015, 0.05, 0.1, 0.25}; $\tau$ $\in$ {1e-08, 1e-06, 0.0001, 0.01, 0.03, 0.1}\strut} & 30 & 150 & \parbox[t]{\linewidth}{\raggedright lr=0.05, tau=0.03\strut} \\
JD (UPGrad) & \parbox[t]{\linewidth}{\raggedright $\eta$ $\in$ {1e-05, 3e-05, 0.0001, 0.001, 0.01, 0.1}\strut} & 6 & 30 & \parbox[t]{\linewidth}{\raggedright lr=0.001, aggregator=UPGrad, inner\_optimizer=Adam\strut} \\
Stochastic L-BFGS & \parbox[t]{\linewidth}{\raggedright $\eta$ $\in$ {0.1, 0.5, 1}; max iter $\in$ {1, 2, 3}; history $\in$ {2, 5, 10}\strut} & 27 & 135 & \parbox[t]{\linewidth}{\raggedright lr=0.1$^{\downarrow}$, max\_iter=3, history\_size=2\strut} \\
Muon & \parbox[t]{\linewidth}{\raggedright $\eta$ $\in$ {1e-05, 3e-05, 0.0001, 0.001, 0.01, 0.1, 0.3, 1}\strut} & 8 & 40 & \parbox[t]{\linewidth}{\raggedright lr=0.001\strut} \\
MuonW & \parbox[t]{\linewidth}{\raggedright $\eta$ $\in$ {1e-05, 3e-05, 0.0001, 0.001, 0.01, 0.1, 0.3, 1}\strut} & 8 & 40 & \parbox[t]{\linewidth}{\raggedright lr=0.001, weight\_decay=0.1\strut} \\
Polyak SGD & \parbox[t]{\linewidth}{\raggedright no swept hyperparameter\strut} & 1 & 5 & \parbox[t]{\linewidth}{\raggedright f\_star=0, max\_lr=1, eps=1e-08\strut} \\
RMSprop & \parbox[t]{\linewidth}{\raggedright $\eta$ $\in$ {1e-05, 3e-05, 0.0001, 0.001, 0.01, 0.1, 0.3, 1}\strut} & 8 & 40 & \parbox[t]{\linewidth}{\raggedright lr=0.001\strut} \\
SGD & \parbox[t]{\linewidth}{\raggedright $\eta$ $\in$ {1e-05, 3e-05, 0.0001, 0.001, 0.01, 0.1, 0.3, 1}\strut} & 8 & 40 & \parbox[t]{\linewidth}{\raggedright lr=0.1\strut} \\
SGD + momentum & \parbox[t]{\linewidth}{\raggedright $\eta$ $\in$ {1e-05, 3e-05, 0.0001, 0.001, 0.01, 0.1, 0.3, 1}\strut} & 8 & 40 & \parbox[t]{\linewidth}{\raggedright lr=0.01\strut} \\
SOAP & \parbox[t]{\linewidth}{\raggedright $\eta$ $\in$ {1e-05, 3e-05, 0.0001, 0.001, 0.01, 0.1, 0.3, 1}\strut} & 8 & 40 & \parbox[t]{\linewidth}{\raggedright lr=0.01\strut} \\
Shampoo & \parbox[t]{\linewidth}{\raggedright $\eta$ $\in$ {1e-05, 3e-05, 0.0001, 0.001, 0.01, 0.1, 0.3, 1}\strut} & 8 & 40 & \parbox[t]{\linewidth}{\raggedright lr=0.1\strut} \\
\bottomrule
\end{tabular}
}
}

  \smallskip
  \svnfittable{\scriptsize\resizebox{\linewidth}{!}{%
\begin{tabular}{lp{2.2in}rrp{1.1in}}
\toprule
\textbf{Method} & \parbox[t]{\linewidth}{\raggedright \textbf{Grid swept}\strut} & \textbf{Configs} & \textbf{Runs} & \parbox[t]{\linewidth}{\raggedright \textbf{Selected}\strut} \\
\midrule
Sven & \parbox[t]{\linewidth}{\raggedright $\eta$ $\in$ {0.05, 0.1, 0.5, 1}; $k$ $\in$ {1, 2, 4, 8, 16, 32, 48, 64}; rtol $\in$ {0.0001, 0.001, 0.01, 0.1, 0.3}\strut} & 160 & 800 & \parbox[t]{\linewidth}{\raggedright k=32, lr=0.5, rtol=0.3$^{\uparrow}$\strut} \\
Adam & \parbox[t]{\linewidth}{\raggedright $\eta$ $\in$ {1e-05, 3e-05, 0.0001, 0.001, 0.01, 0.1, 0.3, 1}\strut} & 8 & 40 & \parbox[t]{\linewidth}{\raggedright lr=0.0001\strut} \\
AdamW & \parbox[t]{\linewidth}{\raggedright $\eta$ $\in$ {1e-05, 3e-05, 0.0001, 0.001, 0.01, 0.1, 0.3, 1}\strut} & 8 & 40 & \parbox[t]{\linewidth}{\raggedright lr=0.001, weight\_decay=0.01\strut} \\
HIG & \parbox[t]{\linewidth}{\raggedright $\eta$ $\in$ {0.005, 0.015, 0.05, 0.1, 0.25}; $\tau$ $\in$ {1e-08, 1e-06, 0.0001, 0.01, 0.03, 0.1}\strut} & 30 & 150 & \parbox[t]{\linewidth}{\raggedright lr=0.05, tau=0.03\strut} \\
JD (UPGrad) & \parbox[t]{\linewidth}{\raggedright $\eta$ $\in$ {1e-05, 3e-05, 0.0001, 0.001, 0.01, 0.1}\strut} & 6 & 30 & \parbox[t]{\linewidth}{\raggedright lr=0.0001, aggregator=UPGrad, inner\_optimizer=Adam\strut} \\
Stochastic L-BFGS & \parbox[t]{\linewidth}{\raggedright $\eta$ $\in$ {0.1, 0.5, 1, 2, 4}; max iter $\in$ {1, 2, 3}; history $\in$ {2, 5, 10}\strut} & 45 & 225 & \parbox[t]{\linewidth}{\raggedright lr=0.5, max\_iter=1, history\_size=2\strut} \\
Muon & \parbox[t]{\linewidth}{\raggedright $\eta$ $\in$ {1e-05, 3e-05, 0.0001, 0.001, 0.01, 0.1, 0.3, 1}\strut} & 8 & 40 & \parbox[t]{\linewidth}{\raggedright lr=0.0001\strut} \\
MuonW & \parbox[t]{\linewidth}{\raggedright $\eta$ $\in$ {1e-05, 3e-05, 0.0001, 0.001, 0.01, 0.1, 0.3, 1}\strut} & 8 & 40 & \parbox[t]{\linewidth}{\raggedright lr=0.0001, weight\_decay=0.1\strut} \\
Polyak SGD & \parbox[t]{\linewidth}{\raggedright no swept hyperparameter\strut} & 1 & 5 & \parbox[t]{\linewidth}{\raggedright f\_star=0, max\_lr=1, eps=1e-08\strut} \\
RMSprop & \parbox[t]{\linewidth}{\raggedright $\eta$ $\in$ {1e-05, 3e-05, 0.0001, 0.001, 0.01, 0.1, 0.3, 1}\strut} & 8 & 40 & \parbox[t]{\linewidth}{\raggedright lr=0.0001\strut} \\
SGD & \parbox[t]{\linewidth}{\raggedright $\eta$ $\in$ {1e-05, 3e-05, 0.0001, 0.001, 0.01, 0.1, 0.3, 1}\strut} & 8 & 40 & \parbox[t]{\linewidth}{\raggedright lr=0.01\strut} \\
SGD + momentum & \parbox[t]{\linewidth}{\raggedright $\eta$ $\in$ {1e-05, 3e-05, 0.0001, 0.001, 0.01, 0.1, 0.3, 1}\strut} & 8 & 40 & \parbox[t]{\linewidth}{\raggedright lr=0.001\strut} \\
SOAP & \parbox[t]{\linewidth}{\raggedright $\eta$ $\in$ {1e-05, 3e-05, 0.0001, 0.001, 0.01, 0.1, 0.3, 1}\strut} & 8 & 40 & \parbox[t]{\linewidth}{\raggedright lr=0.0001\strut} \\
Shampoo & \parbox[t]{\linewidth}{\raggedright $\eta$ $\in$ {1e-05, 3e-05, 0.0001, 0.001, 0.01, 0.1, 0.3, 1}\strut} & 8 & 40 & \parbox[t]{\linewidth}{\raggedright lr=0.1\strut} \\
\bottomrule
\end{tabular}
}
}
\end{table}

\begin{table}[t]
  \centering
  \caption{Hyperparameter grid for nanoGPT on character-level Shakespeare. We only use one $k$ and \texttt{rtol} for Sven, so the scan is over $\eta$. Conventions as in Table~\ref{tab:grid-toy}.}
  \label{tab:grid-nanogpt}
  \svnfittable{\scriptsize\resizebox{\linewidth}{!}{%
\begin{tabular}{lp{2.2in}rrp{1.1in}}
\toprule
\textbf{Method} & \parbox[t]{\linewidth}{\raggedright \textbf{Grid swept}\strut} & \textbf{Configs} & \textbf{Runs} & \parbox[t]{\linewidth}{\raggedright \textbf{Selected}\strut} \\
\midrule
Sven & \parbox[t]{\linewidth}{\raggedright $\eta$ $\in$ {0.05, 0.1, 0.5, 1}\strut} & 4 & 20 & \parbox[t]{\linewidth}{\raggedright k=64, lr=0.1, rtol=0.001\strut} \\
AdamW & \parbox[t]{\linewidth}{\raggedright $\eta$ $\in$ {1e-05, 3e-05, 0.0001, 0.0003, 0.001, 0.003}\strut} & 6 & 30 & \parbox[t]{\linewidth}{\raggedright lr=0.0003, weight\_decay=0.01\strut} \\
Muon & \parbox[t]{\linewidth}{\raggedright $\eta$ $\in$ {1e-05, 3e-05, 0.0001, 0.0003, 0.001, 0.003}\strut} & 6 & 30 & \parbox[t]{\linewidth}{\raggedright lr=0.0001\strut} \\
MuonW & \parbox[t]{\linewidth}{\raggedright $\eta$ $\in$ {1e-05, 3e-05, 0.0001, 0.0003, 0.001, 0.003}\strut} & 6 & 30 & \parbox[t]{\linewidth}{\raggedright lr=0.0001, weight\_decay=0.1\strut} \\
SOAP & \parbox[t]{\linewidth}{\raggedright $\eta$ $\in$ {1e-05, 3e-05, 0.0001, 0.0003, 0.001, 0.003}\strut} & 6 & 30 & \parbox[t]{\linewidth}{\raggedright lr=0.0001\strut} \\
\bottomrule
\end{tabular}
}
}
\end{table}

\begin{table}[t]
  \centering
  \caption{Hyperparameter grids for CIFAR-10 label regression (top) and cross-entropy (bottom). The cross-entropy \texttt{rtol} axis was extended after the first pass placed the optimum at its edge, which is why that grid is ragged. Conventions as in Table~\ref{tab:grid-toy}.}
  \label{tab:grid-cifar}
  \svnfittable{\scriptsize\resizebox{\linewidth}{!}{%
\begin{tabular}{lp{2.2in}rrp{1.1in}}
\toprule
\textbf{Method} & \parbox[t]{\linewidth}{\raggedright \textbf{Grid swept}\strut} & \textbf{Configs} & \textbf{Runs} & \parbox[t]{\linewidth}{\raggedright \textbf{Selected}\strut} \\
\midrule
Sven & \parbox[t]{\linewidth}{\raggedright $\eta$ $\in$ {0.1, 0.5, 1}; $k$ $\in$ {64, 128}; rtol $\in$ {0.0001, 0.001, 0.01}\strut} & 18 & 90 & \parbox[t]{\linewidth}{\raggedright k=128$^{\uparrow}$, lr=0.5, rtol=0.001\strut} \\
Adam & \parbox[t]{\linewidth}{\raggedright $\eta$ $\in$ {1e-05, 3e-05, 0.0001, 0.001, 0.01, 0.1, 0.3}\strut} & 7 & 35 & \parbox[t]{\linewidth}{\raggedright lr=0.01\strut} \\
AdamW & \parbox[t]{\linewidth}{\raggedright $\eta$ $\in$ {1e-05, 3e-05, 0.0001, 0.001, 0.01, 0.1, 0.3}\strut} & 7 & 35 & \parbox[t]{\linewidth}{\raggedright lr=0.01, weight\_decay=0.01\strut} \\
Stochastic L-BFGS & \parbox[t]{\linewidth}{\raggedright $\eta$ $\in$ {0.1, 0.5, 1, 2, 4}; max iter $\in$ {1, 2, 3}; history $\in$ {2, 5, 10}\strut} & 45 & 225 & \parbox[t]{\linewidth}{\raggedright lr=2, max\_iter=3, history\_size=2\strut} \\
Muon & \parbox[t]{\linewidth}{\raggedright $\eta$ $\in$ {1e-05, 3e-05, 0.0001, 0.001, 0.01, 0.1, 0.3}\strut} & 7 & 35 & \parbox[t]{\linewidth}{\raggedright lr=0.01\strut} \\
MuonW & \parbox[t]{\linewidth}{\raggedright $\eta$ $\in$ {1e-05, 3e-05, 0.0001, 0.001, 0.01, 0.1, 0.3}\strut} & 7 & 35 & \parbox[t]{\linewidth}{\raggedright lr=0.01, weight\_decay=0.1\strut} \\
Polyak SGD & \parbox[t]{\linewidth}{\raggedright no swept hyperparameter\strut} & 1 & 5 & \parbox[t]{\linewidth}{\raggedright f\_star=0, max\_lr=1, eps=1e-08\strut} \\
RMSprop & \parbox[t]{\linewidth}{\raggedright $\eta$ $\in$ {1e-05, 3e-05, 0.0001, 0.001, 0.01, 0.1, 0.3}\strut} & 7 & 35 & \parbox[t]{\linewidth}{\raggedright lr=0.001\strut} \\
SGD & \parbox[t]{\linewidth}{\raggedright $\eta$ $\in$ {1e-05, 3e-05, 0.0001, 0.001, 0.01, 0.1, 0.3}\strut} & 7 & 35 & \parbox[t]{\linewidth}{\raggedright lr=0.001\strut} \\
SGD + momentum & \parbox[t]{\linewidth}{\raggedright $\eta$ $\in$ {1e-05, 3e-05, 0.0001, 0.001, 0.01, 0.1, 0.3}\strut} & 7 & 35 & \parbox[t]{\linewidth}{\raggedright lr=0.001\strut} \\
SOAP & \parbox[t]{\linewidth}{\raggedright $\eta$ $\in$ {1e-05, 3e-05, 0.0001, 0.001, 0.01, 0.1, 0.3}\strut} & 7 & 35 & \parbox[t]{\linewidth}{\raggedright lr=0.01\strut} \\
\bottomrule
\end{tabular}
}
}

  \smallskip
  \svnfittable{\scriptsize\resizebox{\linewidth}{!}{%
\begin{tabular}{lp{2.2in}rrp{1.1in}}
\toprule
\textbf{Method} & \parbox[t]{\linewidth}{\raggedright \textbf{Grid swept}\strut} & \textbf{Configs} & \textbf{Runs} & \parbox[t]{\linewidth}{\raggedright \textbf{Selected}\strut} \\
\midrule
Sven & \parbox[t]{\linewidth}{\raggedright $\eta$ $\in$ {0.02, 0.05, 0.1, 0.5, 1}; $k$ $\in$ {64, 128}; rtol $\in$ {0.0001, 0.001, 0.01, 0.03, 0.1, 0.3}\strut} & 39 & 195 & \parbox[t]{\linewidth}{\raggedright k=128$^{\uparrow}$, lr=0.5, rtol=0.3$^{\uparrow}$\strut} \\
Adam & \parbox[t]{\linewidth}{\raggedright $\eta$ $\in$ {1e-05, 3e-05, 0.0001, 0.001, 0.01, 0.1, 0.3, 1, 3}\strut} & 9 & 45 & \parbox[t]{\linewidth}{\raggedright lr=0.001\strut} \\
AdamW & \parbox[t]{\linewidth}{\raggedright $\eta$ $\in$ {1e-05, 3e-05, 0.0001, 0.001, 0.01, 0.1, 0.3, 1, 3}\strut} & 9 & 45 & \parbox[t]{\linewidth}{\raggedright lr=0.01, weight\_decay=0.01\strut} \\
Stochastic L-BFGS & \parbox[t]{\linewidth}{\raggedright $\eta$ $\in$ {0.1, 0.5, 1, 2, 4}; max iter $\in$ {1, 2, 3}; history $\in$ {2, 5, 10}\strut} & 45 & 225 & \parbox[t]{\linewidth}{\raggedright lr=2, max\_iter=2, history\_size=5\strut} \\
Muon & \parbox[t]{\linewidth}{\raggedright $\eta$ $\in$ {1e-05, 3e-05, 0.0001, 0.001, 0.01, 0.1, 0.3, 1, 3}\strut} & 9 & 45 & \parbox[t]{\linewidth}{\raggedright lr=0.01\strut} \\
MuonW & \parbox[t]{\linewidth}{\raggedright $\eta$ $\in$ {1e-05, 3e-05, 0.0001, 0.001, 0.01, 0.1, 0.3, 1, 3}\strut} & 9 & 45 & \parbox[t]{\linewidth}{\raggedright lr=0.01, weight\_decay=0.1\strut} \\
Polyak SGD & \parbox[t]{\linewidth}{\raggedright no swept hyperparameter\strut} & 1 & 5 & \parbox[t]{\linewidth}{\raggedright f\_star=0, max\_lr=1, eps=1e-08\strut} \\
RMSprop & \parbox[t]{\linewidth}{\raggedright $\eta$ $\in$ {1e-05, 3e-05, 0.0001, 0.001, 0.01, 0.1, 0.3, 1, 3}\strut} & 9 & 45 & \parbox[t]{\linewidth}{\raggedright lr=0.001\strut} \\
SGD & \parbox[t]{\linewidth}{\raggedright $\eta$ $\in$ {1e-05, 3e-05, 0.0001, 0.001, 0.01, 0.1, 0.3, 1, 3}\strut} & 9 & 45 & \parbox[t]{\linewidth}{\raggedright lr=3$^{\uparrow}$\strut} \\
SGD + momentum & \parbox[t]{\linewidth}{\raggedright $\eta$ $\in$ {1e-05, 3e-05, 0.0001, 0.001, 0.01, 0.1, 0.3, 1, 3}\strut} & 9 & 45 & \parbox[t]{\linewidth}{\raggedright lr=0.3\strut} \\
SOAP & \parbox[t]{\linewidth}{\raggedright $\eta$ $\in$ {1e-05, 3e-05, 0.0001, 0.001, 0.01, 0.1, 0.3, 1, 3}\strut} & 9 & 45 & \parbox[t]{\linewidth}{\raggedright lr=0.01\strut} \\
\bottomrule
\end{tabular}
}
}
\end{table}

\clearpage
\section{Efficient Implementation of Sven}
\label{app:gram}

The cost of a naive Sven step is dominated by forming the $B \times P$ Jacobian, which is tens of gigabytes at ResNet18 scale. This appendix describes the optimized implementation we develop for this paper, which avoids fully materializing $M$ in memory. It rests on the observation that the truncated pseudoinverse update depends on $M$ only through the $B \times B$ Gram matrix $M M^T$ and through one vector--Jacobian product $M^T w$, both of which can be obtained from a constant number of ordinary forward and backward passes. The classic path --- fully materializing $M$ and performing a randomized truncated SVD (Appendix~\ref{app:profiling}, Algorithms~\ref{alg:rsvd-v1} and~\ref{alg:rsvd-v2}) --- was implemented but is not used in practice.

\textbf{Note:} The efficient Gram implementation is only possible for architectures that do not mix between samples in a batch (e.g.\ BatchNorm) or reuse components in multiple places (e.g.\ transformer input/output weight tying).

\subsection{The Gram identity}
\label{app:gram_identity}

Write the thin SVD of the batch Jacobian as $M = U s V^T$, with $U \in \mathbb{R}^{B\times B}$, $s = \mathrm{diag}(\sigma_1 \geq \dots \geq \sigma_B)$ and $V \in \mathbb{R}^{P \times B}$. Truncating to an index set $\mathcal{K}$ of retained directions, the update of \Eqref{eqn:update} is
\begin{equation}
    \label{eqn:gram_update}
    M^{+}\mathcal{R} \;=\; \sum_{i \in \mathcal{K}} v_i\, \sigma_i^{-1}\, (u_i^T \mathcal{R})
    \;=\; M^T \underbrace{\left[\sum_{i \in \mathcal{K}} u_i\, \sigma_i^{-2}\,
    (u_i^T \mathcal{R})\right]}_{\textstyle w \,\in\, \mathbb{R}^{B}} ,
\end{equation}
using $M^T u_i = \sigma_i v_i$. The right singular vectors $v_i$, the only objects of dimension $P$, have canceled. The pairs $(\sigma_i, u_i)$ are the square roots of the eigenvalues and the eigenvectors of the Gram matrix
\begin{equation}
    \label{eqn:gram_def}
    G \;=\; M M^T \;=\; U s^2 U^T \;\in\; \mathbb{R}^{B \times B}.
\end{equation}
This requires only one eigendecomposition of a symmetric matrix of size $B \ll P$. The product $M^T w$, which is the gradient of the scalar $\sum_\alpha w_\alpha \mathcal{R}^\alpha(\theta)$ with respect to $\theta$, requires one ordinary backward
pass of a weighted residual sum.
Both $G$ and its eigendecomposition are kept in double precision regardless of the precision the model trains in (discussed in Appendix~\ref{app:gram_trunc}).

\subsection{Assembling the Gram matrix without per-sample gradients}
\label{app:gram_capture}

\Eqref{eqn:gram_def} is a sum of inner products of per-sample gradients, and the default Sven capture (\texttt{hooks}) obtains all of them from one forward and one backward pass. Forward hooks on every parameter-carrying layer $l$ record that layer's input $x_l$ and output $z_l$. A single backward pass then differentiates the \emph{sum} of the residual rows with respect to the captured outputs,
\begin{equation}
    \label{eqn:gram_gl}
    g_l \;=\; \frac{\partial}{\partial z_l} \sum_{\alpha} \mathcal{R}^\alpha ,
\end{equation}
and because distinct conditions are decoupled --- $\mathcal{R}^\alpha$ depends only on the activations of sample $\alpha$ --- the slice of $g_l$ belonging to sample $\alpha$ is exactly $\partial \mathcal{R}^\alpha / \partial z_l^\alpha$. One backward pass therefore yields all $B$ per-sample output gradients, and it targets layer \emph{outputs}, so no parameter gradient is ever formed. Because the rows enter only through \Eqref{eqn:gram_gl}, the chain factors of the $\kappa$ power, of the signed-residual row definition and of the micro-batch mean (Appendix~\ref{app:gram_knobs}) are already inside $g_l$ and need no special treatment.

For a layer applied once per sample with a two-dimensional input, the per-sample gradient with respect to the weight is the outer product $\partial \mathcal{R}^\alpha / \partial W_l = g_l^\alpha (x_l^\alpha)^T$, so its contribution to $G$ factorizes into the Hadamard product of two Gram matrices,
\begin{equation}
    \label{eqn:gram_layer}
    G^{(l)} \;=\; \big(g_l g_l^T\big) \odot \big(x_l x_l^T\big)
    \;+\; \big(g_l g_l^T\big)\big|_{\text{bias}},
    \qquad G \;=\; \sum_l G^{(l)} ,
\end{equation}
where the second term is present when the layer has a bias. Only $B \times B$ objects are built; the per-sample gradient itself is never stored. This factorization is standard in the per-example-gradient literature~\citep{goodfellow2015efficient}; what is new here is only that the Gram matrix, and not the per-sample gradients, is what we need out of it.
Three cases need more than \Eqref{eqn:gram_layer}:
\begin{itemize}
    \item \textbf{Layers applied at many positions per sample} (a projection inside a transformer block, a convolution). The per-sample gradient is then a sum of outer products over positions and no longer factorizes. The implementation forms the flattened per-sample gradient blocks explicitly --- $\sum_{\rm pos} g_l^{\alpha}\,(x_l^{\alpha})^T$ for a projection, $\texttt{unfold}(x_l)$ contracted against $g_l$ for a \texttt{Conv2d} --- and contracts them into $G$ immediately. Each block is $\mathcal{O}(B\, d_{\rm out} d_{\rm in})$ rather than $\mathcal{O}(B P)$, and the batch is processed in chunks whenever a block would exceed a fixed element budget, one side being recomputed for off-diagonal chunk pairs rather than held, so the transient is bounded independently of $B$.
    \item \textbf{Embeddings}, whose per-sample gradient is a sparse scatter of $g_l$ into the rows the sample's tokens select. These are densely scattered into blocks and contracted under the same budget. 
    \item \textbf{Normalization affine parameters}. With the statistics held fixed the per-sample gradients are $\sum_{\rm pos} g_l \odot \hat{x}_l$ and $\sum_{\rm pos} g_l$, each contributing $A A^T$ with $A$ the resulting $(B, C)$ matrix. \texttt{LayerNorm} normalizes over trailing dimensions at each position, so it is per-sample by construction and $\hat{x}_l$ is recomputed from the captured input; layers that normalize with batch statistics are treated in Appendix~\ref{app:gram_fallback}.
\end{itemize}
The capture therefore costs one forward and one backward pass plus the per-layer contractions. The memory cost is what an ordinary backward pass already holds plus the double-precision contractions and $B \times B$ accumulator; much smaller than the $B \times P$ Jacobian. A complete step is two forward--backward pairs (capture and $M^T w$) and one $B \times B$ eigendecomposition.

\subsection{Architectural restrictions and fallbacks}
\label{app:gram_fallback}

\Eqref{eqn:gram_gl} is exact only where the batch elements are decoupled. The efficient implementation requires that every captured layer is applied exactly once per forward pass, that no parameters be tied between layers, and that nothing resample between the capture and the update. Train-mode dropout, repeated modules, tied weights, grouped or non-zero-padded convolutions, a normalization layer with no running statistics to hold fixed, and any parameter-carrying layer outside the supported set (linear, \texttt{Conv2d} with a single group, normalization, \texttt{LayerNorm}, embedding) are not supported.

The important case is BatchNorm, where activations of sample $\alpha$ depend on the whole batch. This couples batch elements, and what the hooks recover is not $\partial\mathcal{R}^\alpha/\partial\theta$. Freezing the statistics severely degrades performance for ResNet18 on CIFAR-10, so the ResNet18 experiments (Appendix~\ref{app:cifar_10}) use a different capture. Two exact fallbacks compute $G$ from Jacobian blocks of bounded size, both by reverse-mode differentiation of the residual rows with respect to a \emph{subset} of the parameters at a time:
\begin{itemize}
    \item \texttt{full}: one group holding every parameter, i.e.\ the dense $(B, P)$ Jacobian materialized once and contracted into $G$. It needs the fewest passes and the most memory. We use this form for the ResNet18 studies.
    \item \texttt{chunked}: the parameter tensors are partitioned into groups of at most a fixed number of elements; each group's $(B, P_{\mathrm{grp}})$ Jacobian block is formed, contracted into $G$ and discarded, and a single tensor larger than the budget forms its own group whose rows are themselves computed in batches. Peak memory is set by the largest group rather than by $P$, at the price of one differentiation pass per group. We do not use this form in this paper, but it would be necessary for very large models that could not use the efficient Gram capture.
\end{itemize}

\subsection{Truncation and numerical resolution}
\label{app:gram_trunc}

The retained set $\mathcal{K}$ of \Eqref{eqn:gram_update} is formed in the Gram eigenbasis, reproducing the truncation of Algorithm~\ref{alg:update}: the eigenvalues are sorted in descending order, $\sigma_i = \sqrt{\max(\lambda_i, 0)}$, the ones with $\sigma_i \leq \texttt{rtol}\cdot\sigma_1$ are dropped, and of those that survive the $k$ largest are kept.
An empty $\mathcal{K}$ means $G$ is zero or non-finite, and raises an error (runs counted as ``diverged''). Forming $G$ squares the condition number, $\mathrm{cond}(G) = \mathrm{cond}(M)^2$, which sets the resolution of the spectrum. 
We thus accumulate $G$ in float64 regardless of what the model uses.

\subsection{Micro-batches and parameter masks}
\label{app:gram_knobs}
To reduce memory usage in cases where the efficient Gram capture cannot be used, we introduce two options: micro-batching and parameter batching. We describe them here, and include experimental studies in Appendix~\ref{app:lims}. 

\paragraph{Micro-batches} Instead of treating a full batch elementwise, we group it into ``micro-batches" $\mathcal{D}_b$ of size $1 < \mu \leq B$ and sum- or mean-aggregate sample-wise losses in a micro-batch into a single scalar loss. Mathematically, 
\begin{equation}
    L(\theta) = \sum_{i=1}^B L_i(\theta) = \sum_{b=1}^{B / \mu}\underbrace{\sum_{\alpha\in\mathcal D_{b}}\ell^\alpha(\theta)}_{\tilde{\ell}_b(\theta)},
\end{equation}
where the $\mathcal{D}_b$ are a partition of the batch. We then apply Sven at the level of micro-batches, using $\tilde{\ell}_b$ as the scalar losses. This requires instantiating only a $(B/\mu) \times P$ Jacobian, saving a factor of $\mu$ in memory. As the decomposition becomes coarser (fewer, larger micro-batches), our method converges to traditional SGD.

\paragraph{Batched parameters.} 
While batching \textit{data} is standard practice to reduce computational overhead and introduce beneficial stochasticity, batching \textit{parameters} is rarely, if ever, considered. 
The batched parameter approach simply involves updating a randomly selected subset of parameters at each optimizer step, and is implemented using masks.
In principle this should reduce memory overhead, but in practice we find that PyTorch and JAX still appear to instantiate full intermediate gradients ``under the hood'', and the memory savings are not realized. 
It is also worth noting that if we allow for sufficiently batched parameters, we can instead effectively learn in the underparametrized limit, and use the analytic form for the Moore-Penrose inverse in \Eqref{eqn:MPUnder}.

\subsection{The algorithm}
\label{app:gram_alg}

\begin{algorithm}[t]
  \caption{Sven update step, Gram implementation. Lines 1--4 are the \texttt{hooks} capture;
  the \texttt{full} and \texttt{chunked} captures replace them by Jacobian blocks contracted into
  the same $G$ (Appendix~\ref{app:gram_fallback}). 
  }
  \label{alg:gram}
  \begin{algorithmic}[1]
    \Require batch of conditions $\{\alpha\}$ with $B$ rows, parameters $\theta$,
      learning rate $\eta$, rank $k$, tolerance \texttt{rtol}, exponent $\kappa$
    \State $\{x_l, z_l\},\, \mathcal{R}^\alpha \leftarrow$ one forward pass, recording each
      layer's input and output \Comment{$\mathcal{R}^\alpha = (\ell^\alpha)^{\kappa/2}$}
    \State $g_l \leftarrow \partial \big(\sum_\alpha \mathcal{R}^\alpha\big) / \partial z_l$
      \Comment{one backward; rows decoupled}
    \State $G \leftarrow \sum_l \big[(g_l g_l^T) \odot (x_l x_l^T) + (g_l g_l^T)|_{\rm bias}\big]$
      \Comment{$(B,B)$, float64}
    \State $\lambda, U \leftarrow \mathrm{eigh}(G)$, sorted descending; $\;\sigma_i \leftarrow
      \sqrt{\max(\lambda_i, 0)}$ \Comment{independent of $k$}
    \State $\mathcal{K} \leftarrow \{ i \leq k \;:\; \sigma_i > \texttt{rtol}\cdot\sigma_1 \}$
      \Comment{\Eqref{eqn:rank_law}}
    \If{$\mathcal{K} = \emptyset$}
      \State \Return failure \Comment{$G$ zero or non-finite: the run has diverged}
    \EndIf
    \State $w \leftarrow \sum_{i \in \mathcal{K}} u_i \, \sigma_i^{-2} \, (u_i^T R)$
    \State $\delta\theta \leftarrow -\eta \, \partial \big(\sum_\alpha w_\alpha
      \mathcal{R}^\alpha\big) / \partial \theta$
      \Comment{$= -\eta\, M^T w$; one forward, one backward}
    \State \Return $\theta + \delta\theta$
  \end{algorithmic}
\end{algorithm}

\begin{table}[t]
  \centering
  \caption{Step time and peak memory of Sven's four backends: \texttt{hooks},
  \texttt{full} and \texttt{chunked} Gram captures and the classic explicit-Jacobian path with a randomized truncated SVD. Column 4 reports the step time ratio relative to Adam. The final column reports the spread of the step time across $k$. Missing rows are configurations that cannot be run: \texttt{hooks} is unavailable on ResNet18 because it cannot handle BatchNorm.}
  \label{tab:gram_cost}
  \svnfittable{\begin{tabular}{lrrrrrrrr}
\toprule
\textbf{Architecture} & \textbf{Backend} & \textbf{Step (ms)} & \textbf{x Adam} & \textbf{Capture (ms)} & \textbf{Capture share} & \textbf{Solve+apply (ms)} & \textbf{Peak (MB)} & \textbf{k spread} \\
\midrule
Toy 1D MLP & Sven (Gram, hooks) & 11.5 & 3.11 & 4.78 & 42\% & 6.72 & 18.86 & 1.02 \\
 & Sven (Gram, full $J$) & 15.4 & 4.16 & 7.74 & 50\% & 7.67 & 18.86 & 1 \\
 & Sven (Gram, chunked) & 159.7 & 43.2 & 152 & 95\% & 8.11 & 18.86 & -- \\
 & Sven (classic, rand. SVD) & 10.28 & 2.78 & 7.39 & 72\% & 2.89 & 21.41 & 1.01 \\
\midrule
Polynomial MLP & Sven (Gram, hooks) & 11.71 & 3.15 & 4.86 & 41\% & 6.86 & 18.87 & 1.01 \\
 & Sven (Gram, full $J$) & 15.39 & 4.15 & 7.72 & 50\% & 7.66 & 18.87 & 1 \\
 & Sven (Gram, chunked) & 163.6 & 44.1 & 155 & 95\% & 8.08 & 18.87 & -- \\
 & Sven (classic, rand. SVD) & 10.5 & 2.83 & 7.46 & 71\% & 3.04 & 21.44 & 1.01 \\
\midrule
MNIST MLP & Sven (Gram, hooks) & 10.28 & 2.62 & 3.79 & 37\% & 6.49 & 20.89 & 1 \\
 & Sven (Gram, full $J$) & 14.31 & 3.65 & 6.64 & 46\% & 7.66 & 48.07 & 1.01 \\
 & Sven (Gram, chunked) & 155.9 & 39.7 & 147 & 94\% & 9.09 & 45.53 & -- \\
 & Sven (classic, rand. SVD) & 10.13 & 2.58 & 6.69 & 66\% & 3.44 & 48.15 & 1.02 \\
\midrule
nanoGPT & Sven (Gram, hooks) & 59.53 & 2.53 & 32.3 & 54\% & 27.3 & 614.2 & 1.02 \\
 & Sven (Gram, full $J$) & 229.2 & 9.74 & 204 & 89\% & 25.6 & 2892 & 1 \\
 & Sven (Gram, chunked) & 423.2 & 18 & 398 & 94\% & 25.7 & 2624 & -- \\
 & Sven (classic, rand. SVD) & 239.2 & 10.2 & 234 & 98\% & 5.05 & 2892 & 1.01 \\
\midrule
CIFAR-10 ResNet18 & Sven (Gram, full $J$) & 188 & 7.89 & 165 & 88\% & 22.8 & 22976 & 1 \\
 & Sven (Gram, chunked) & 207.9 & 8.72 & 182 & 88\% & 25.9 & 6921 & -- \\
 & Sven (classic, rand. SVD) & 1041 & 43.7 & 907 & 87\% & 134 & 23030 & 1.11 \\
\bottomrule
\end{tabular}
}
\end{table}

Algorithm~\ref{alg:gram} summarizes the memory-optimized Sven step. Unlike Algorithm~\ref{alg:update}, the cost is independent of $k$ since the whole spectrum of $G \in \mathbb{R}^{B\times B}$ comes out of the eigendecomposition.

\subsection{Measured cost}
\label{app:gram_cost}

Table~\ref{tab:gram_cost} gives Sven's step time and peak memory for the four backends across the architectures of this paper (see Appendix~\ref{app:profiling} for a full profiling study against the full set of baseline optimizers). The quoted numbers come from the dedicated timing runs discussed in Appendix~\ref{app:exp_details}. When available, the \texttt{hooks} Gram capture is fastest and costs only 2-3$\times$ as much time as an Adam step. For the small MLPs (toy 1D, polynomial) all three Gram capture methods use a similar amount of memory, but for larger models the \texttt{hooks} capture is significantly lighter. In the case of CIFAR-10 + ResNet18, where the hooked capture is unavailable, the \texttt{chunked} capture saves a lot of memory relative to the full Jacobian instantiation. Chunked capture is also much less time efficient for the smaller models because the chunking introduces unnecessary overhead, but catches up to \texttt{full} for ResNet18. Finally, we observe that the ``classic'' algorithm -- which computes a randomized SVD of the full $B \times P$ Jacobian instead of eigendecomposing its Gram matrix -- is the clear loser in both memory and time efficiency for all of the larger models.

\subsection{Remaining Challenges}
\label{app:gram_future}

The largest barriers to scaling Sven are the cases where efficient Gram capture is not possible. When it \textit{is} possible, Sven is relatively close to parity with Adam. A capture that handles BatchNorm, weight tying, etc.\ without materializing blocks, or a training-time policy that removes the coupling without changing what is optimized, would be a natural direction to explore next.

\section{Additional Results}
\label{app:additional_results}

This appendix includes the full details of the scans presented in Sec.~\ref{sec:experiments}: every method's confirmation table, the ranking summary across all seven scans, the all-seed trajectories, and the 1D regression scan in full. The $k$ and \texttt{rtol} sweeps at each scan's selected configuration are in Appendix~\ref{app:spectra}, next to the spectra that explain them.

\begin{figure}[!ht]
    \centering
    \includegraphics[width=\linewidth]{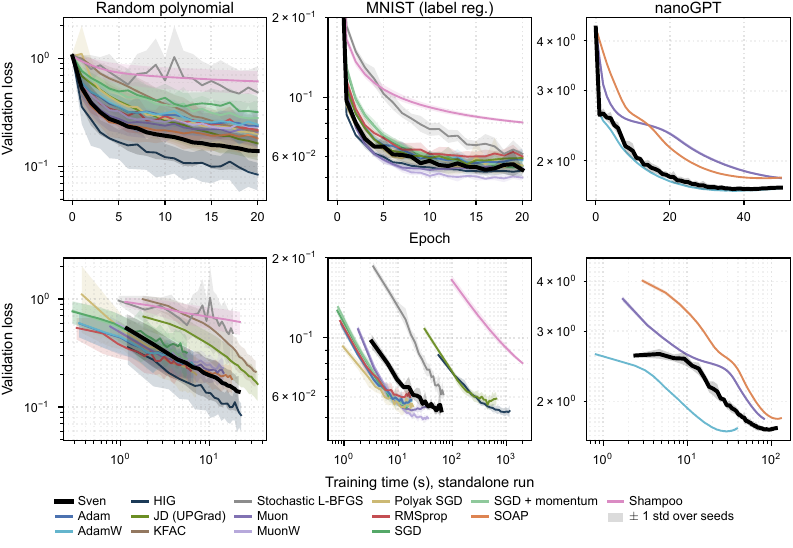}
    \caption{Reproduction of Fig.~\ref{fig:main_results} with all baselines shown: validation loss against epoch (top) and against standalone wall time (bottom). Bands are $\pm 1$ std over the \numNConfirmSeeds{} confirmation seeds. SOAP is omitted for MNIST because it does not converge well.}
    \label{fig:headline_curves_all}
\end{figure}

\begin{figure}[!ht]
    \centering
    \includegraphics[width=\linewidth]{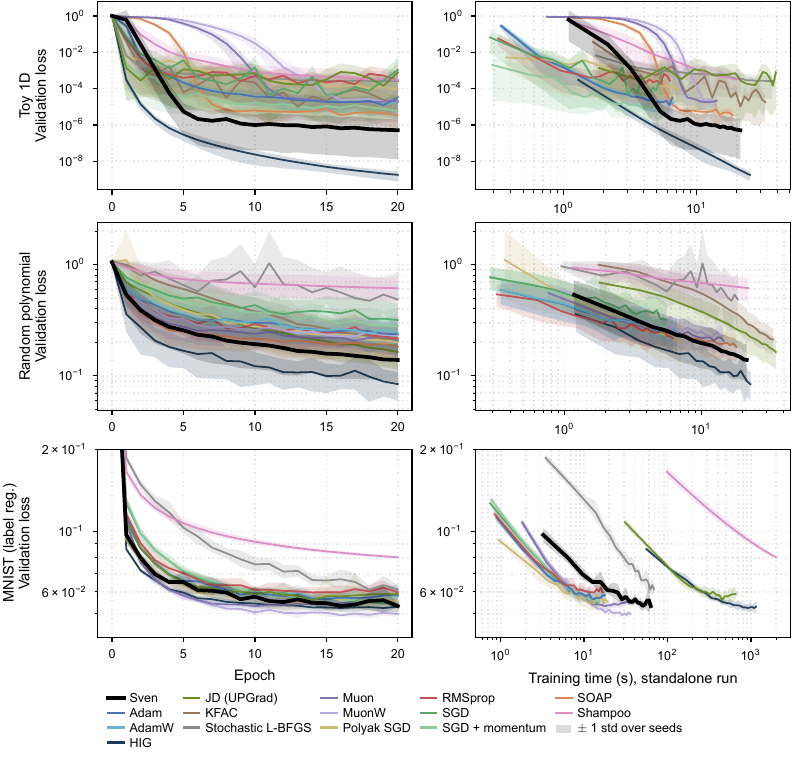}
    \caption{Confirmation trajectories for 1D regression (top), polynomial regression (middle) and MNIST label regression (bottom). We plot validation loss against epoch (left) and against standalone training time (right). Bands are mean $\pm 1$ std over the \numNConfirmSeeds{} confirmation seeds. SOAP is omitted from MNIST because it does not converge well.}
    \label{fig:main_results_with_errorBars}
\end{figure}

\paragraph{All-seed trajectories.} Fig.~\ref{fig:headline_curves_all} redraws Fig.~\ref{fig:main_results} with every method in each field rather than the five best. Fig.~\ref{fig:main_results_with_errorBars} shows similar results for the best configurations on the three \textit{regression} tasks (1D, polynomial, MNIST), with a third column added for training loss. The 1D regression results are included here, rather than in the main text, because they do not differ significantly from the polynomial regression results (and have the largest between-seed variance).

\begin{figure}[!ht]
    \centering
    \includegraphics[width=\linewidth]{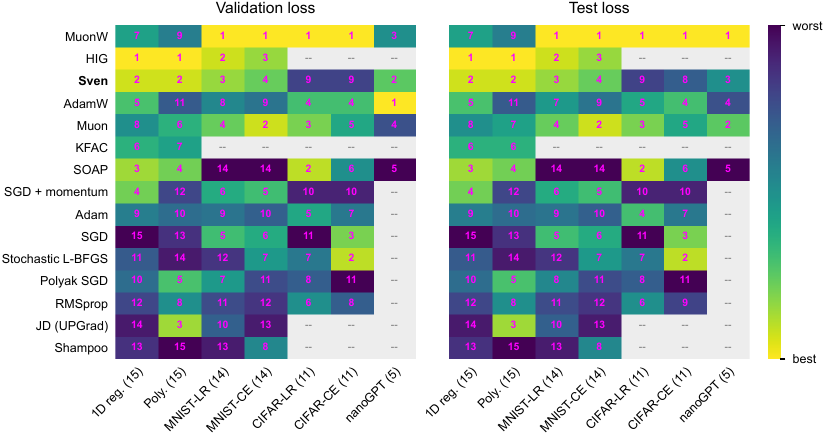}
    \caption{Ranking of the best-performing configurations of each optimizer on each scan using validation loss (left) and test loss (right). Numbers indicate rank, and colors are normalized by the number of contenders for each scan (indicated in the bottom row). A gray cell means the method was not run on that scan or produced no finite confirmation value.}
    \label{fig:ranking}
\end{figure}

\paragraph{Summary tables.} Table~\ref{tab:headline} compactly summarizes the results of Sec.~\ref{sec:experiments}: Sven and seven baselines on the three headline tasks, with the cost columns from the standalone timing pass. The per-scan tables at the end of this appendix give the same for every method.  Figure~\ref{fig:ranking} gives every method's rank on every scan, on validation loss (which selection used) and on test loss (an outcome). Sven is consistently among the top 2 or 3 optimizers, except for the case of CIFAR-10 where it falls far behind the rest.

\begin{table}[!ht]
  \centering
  \caption{Headline results on the confirmation seeds: final validation and test loss (mean $\pm 1$ std over seeds), test accuracy where defined, finished/attempted runs, seconds per epoch and peak memory. Losses are from the confirmation pass; the two cost columns are the standalone timing pass.}
  \label{tab:headline}
  \resizebox{\linewidth}{!}{%
\begin{tabular}{lp{0.95in}rrrrrrr}
\toprule
\textbf{Method} & \parbox[t]{\linewidth}{\raggedright \textbf{Configuration}\strut} & \textbf{Val. loss} & \textbf{Val. (tune)} & \textbf{Test loss} & \textbf{Test acc.} & \textbf{fin./att.} & \textbf{s/ep.} & \textbf{MB} \\
\midrule
\multicolumn{9}{l}{\textit{Random Polynomial}} \\
Sven & \parbox[t]{\linewidth}{\raggedright $\eta$=0.5, $k$=16, rtol=0.03\strut} & \textbf{0.1388 $\boldsymbol{\pm}$ 0.055} & 0.1095 & 0.136 $\pm$ 0.055 & -- & 15/15 & 1.33 & 18.86 \\
HIG & \parbox[t]{\linewidth}{\raggedright $\eta$=0.1, tau=1e-08\strut} & \textbf{0.08406 $\boldsymbol{\pm}$ 0.024} & 0.07149 & 0.08028 $\pm$ 0.02 & -- & 15/15 & 1.4 & 21.59 \\
MuonW & \parbox[t]{\linewidth}{\raggedright $\eta$=0.01, wd=0.1\strut} & 0.2313 $\pm$ 0.07 & 0.2061 & 0.2226 $\pm$ 0.06 & -- & 15/15 & 0.967 & 18.44 \\
Muon & \parbox[t]{\linewidth}{\raggedright $\eta$=0.01\strut} & 0.207 $\pm$ 0.062 & 0.1651 & 0.202 $\pm$ 0.058 & -- & 15/15 & 0.978 & 18.44 \\
SOAP & \parbox[t]{\linewidth}{\raggedright $\eta$=0.01\strut} & 0.1818 $\pm$ 0.05 & 0.1496 & 0.1756 $\pm$ 0.042 & -- & 15/15 & 1.15 & 21.29 \\
AdamW & \parbox[t]{\linewidth}{\raggedright $\eta$=0.01, wd=0.01\strut} & 0.2442 $\pm$ 0.092 & 0.2054 & 0.2403 $\pm$ 0.09 & -- & 15/15 & 0.58 & 18.44 \\
Adam & \parbox[t]{\linewidth}{\raggedright $\eta$=0.01\strut} & 0.2377 $\pm$ 0.087 & 0.1852 & 0.2283 $\pm$ 0.079 & -- & 15/15 & 0.58 & 18.44 \\
SGD & \parbox[t]{\linewidth}{\raggedright $\eta$=0.1\strut} & 0.3182 $\pm$ 0.12 & 0.2263 & 0.315 $\pm$ 0.13 & -- & 15/15 & 0.528 & 18.43 \\
\midrule
\multicolumn{9}{l}{\textit{MNIST (label reg.)}} \\
Sven & \parbox[t]{\linewidth}{\raggedright $\eta$=0.5, $k$=64, rtol=0.001\strut} & 0.05328 $\pm$ 0.0007 & 0.05215 & 0.05363 $\pm$ 0.0024 & 96.7\% & 5/5 & 3.74 & 31.36 \\
HIG & \parbox[t]{\linewidth}{\raggedright $\eta$=0.05, tau=0.03\strut} & 0.05298 $\pm$ 0.0026 & 0.05076 & 0.05324 $\pm$ 0.0023 & 96.6\% & 5/5 & 58.3 & 404.3 \\
MuonW & \parbox[t]{\linewidth}{\raggedright $\eta$=0.001, wd=0.1\strut} & \textbf{0.04982 $\boldsymbol{\pm}$ 0.0013} & 0.04999 & 0.04959 $\pm$ 0.0025 & 97.0\% & 5/5 & 2.4 & 31.59 \\
Muon & \parbox[t]{\linewidth}{\raggedright $\eta$=0.001\strut} & 0.05473 $\pm$ 0.0017 & 0.0558 & 0.05424 $\pm$ 0.0038 & 96.7\% & 5/5 & 2.34 & 31.59 \\
SOAP & \parbox[t]{\linewidth}{\raggedright $\eta$=0.01\strut} & 0.5808 $\pm$ 0.44 & 0.5708 & 0.5732 $\pm$ 0.45 & 44.5\% & 5/5 & 3.34 & 41.83 \\
AdamW & \parbox[t]{\linewidth}{\raggedright $\eta$=0.001, wd=0.01\strut} & 0.05784 $\pm$ 0.0048 & 0.05686 & 0.05613 $\pm$ 0.004 & 96.7\% & 5/5 & 1.46 & 31.7 \\
Adam & \parbox[t]{\linewidth}{\raggedright $\eta$=0.001\strut} & 0.05844 $\pm$ 0.0055 & 0.05683 & 0.057 $\pm$ 0.004 & 96.6\% & 5/5 & 1.48 & 31.7 \\
SGD & \parbox[t]{\linewidth}{\raggedright $\eta$=0.1\strut} & 0.05524 $\pm$ 0.0025 & 0.05338 & 0.05439 $\pm$ 0.0019 & 96.9\% & 5/5 & 1.33 & 31.48 \\
\midrule
\multicolumn{9}{l}{\textit{nanoGPT (tiny-shakespeare)}} \\
Sven & \parbox[t]{\linewidth}{\raggedright $\eta$=0.1, $k$=64, rtol=0.001\strut} & \textbf{1.715 $\boldsymbol{\pm}$ 0.008} & 1.724 & 1.977 $\pm$ 0.024 & -- & 5/5 & 2.4 & 615.4 \\
MuonW & \parbox[t]{\linewidth}{\raggedright $\eta$=0.0001, wd=0.1\strut} & 1.81 $\pm$ 0.0025 & 1.818 & 1.973 $\pm$ 0.0061 & -- & 5/5 & 1.75 & 326.5 \\
Muon & \parbox[t]{\linewidth}{\raggedright $\eta$=0.0001\strut} & 1.812 $\pm$ 0.0023 & 1.819 & 1.976 $\pm$ 0.0058 & -- & 5/5 & 1.76 & 326.5 \\
SOAP & \parbox[t]{\linewidth}{\raggedright $\eta$=0.0001\strut} & 1.817 $\pm$ 0.0062 & 1.822 & 2.033 $\pm$ 0.0092 & -- & 5/5 & 2.73 & 352.3 \\
AdamW & \parbox[t]{\linewidth}{\raggedright $\eta$=0.0003, wd=0.01\strut} & \textbf{1.713 $\boldsymbol{\pm}$ 0.018} & 1.714 & 1.978 $\pm$ 0.02 & -- & 5/5 & 0.914 & 329.7 \\
\bottomrule
\end{tabular}
}

\end{table}

\paragraph{Data-seed replicates.} The two synthetic tasks are each three different random problems, and the confirmation pass runs all three. Table~\ref{tab:dataseeds} separates the two spreads: the standard deviation of the three per-instance means is variation of the target function, while the seed band within an instance is variation of the optimizer's luck. The ratio measures the between-problem variation against the within-problem variation (optimizer luck). For polynomial regression it is $>1$, indicating that the problem instance dominates variance. In this paper we always quote the pooled losses across dataset and model seeds.

\begin{table}[t]
  \centering
  \caption{Data-seed replicates on the two synthetic tasks. Each data seed is a different random problem, so the standard deviation of the three per-instance means is variation of the target and not of the optimizer. A ratio above one says the choice of instance moves the answer more than the seed does.}
  \label{tab:dataseeds}
  \resizebox{\linewidth}{!}{%
\begin{tabular}{lrrrrr}
\toprule
\textbf{Method} & \textbf{Val. (pooled)} & \textbf{Val. (inst. mean)} & \textbf{Betw.-inst. std} & \textbf{Within-inst. std} & \textbf{Ratio} \\
\midrule
\multicolumn{6}{l}{\textit{Toy 1D}} \\
HIG & $1.775\times 10^{-9}$ & $1.775\times 10^{-9}$ & $6.39\times 10^{-10}$ & $8.08\times 10^{-10}$ & 0.791 \\
Sven & $5.086\times 10^{-7}$ & $5.086\times 10^{-7}$ & $6.45\times 10^{-7}$ & $9.45\times 10^{-7}$ & 0.682 \\
SOAP & $3.372\times 10^{-6}$ & $3.372\times 10^{-6}$ & $9.89\times 10^{-7}$ & $3.11\times 10^{-6}$ & 0.318 \\
SGD + momentum & $6.027\times 10^{-6}$ & $6.027\times 10^{-6}$ & $2.06\times 10^{-6}$ & $3.17\times 10^{-6}$ & 0.652 \\
AdamW & $1.735\times 10^{-5}$ & $1.735\times 10^{-5}$ & $9.65\times 10^{-6}$ & $1.47\times 10^{-5}$ & 0.657 \\
KFAC & $1.802\times 10^{-5}$ & $1.802\times 10^{-5}$ & $1.30\times 10^{-5}$ & $1.48\times 10^{-5}$ & 0.878 \\
MuonW & $1.928\times 10^{-5}$ & $1.928\times 10^{-5}$ & $1.82\times 10^{-6}$ & $9.34\times 10^{-6}$ & 0.195 \\
Muon & $2.159\times 10^{-5}$ & $2.159\times 10^{-5}$ & $8.43\times 10^{-6}$ & $1.70\times 10^{-5}$ & 0.495 \\
Adam & $3.264\times 10^{-5}$ & $3.264\times 10^{-5}$ & $2.66\times 10^{-5}$ & $4.53\times 10^{-5}$ & 0.587 \\
Polyak SGD & $8.792\times 10^{-5}$ & $8.792\times 10^{-5}$ & $2.68\times 10^{-5}$ & 0.000104 & 0.257 \\
Stochastic L-BFGS & 0.0002659 & 0.0002644 & $6.93\times 10^{-5}$ & 0.000305 & 0.227 \\
RMSprop & 0.0002746 & 0.0002746 & 0.000262 & 0.000529 & 0.495 \\
Shampoo & 0.0002956 & 0.0002956 & 0.000155 & 0.000346 & 0.449 \\
JD (UPGrad) & 0.0007893 & 0.0007893 & 0.00103 & 0.00153 & 0.675 \\
SGD & 0.001084 & 0.001084 & 0.00177 & 0.00233 & 0.761 \\
\midrule
\multicolumn{6}{l}{\textit{Random Polynomial}} \\
HIG & 0.08406 & 0.08406 & 0.0268 & 0.00816 & 3.28 \\
Sven & 0.1388 & 0.1388 & 0.057 & 0.0259 & 2.2 \\
JD (UPGrad) & 0.1635 & 0.1635 & 0.0524 & 0.0218 & 2.4 \\
SOAP & 0.1818 & 0.1818 & 0.0507 & 0.0243 & 2.09 \\
Polyak SGD & 0.1997 & 0.1997 & 0.0954 & 0.0333 & 2.87 \\
Muon & 0.207 & 0.207 & 0.0682 & 0.0237 & 2.88 \\
KFAC & 0.2118 & 0.2118 & 0.0605 & 0.0241 & 2.51 \\
RMSprop & 0.2173 & 0.2173 & 0.0608 & 0.0225 & 2.71 \\
MuonW & 0.2313 & 0.2313 & 0.0782 & 0.022 & 3.56 \\
Adam & 0.2377 & 0.2377 & 0.0912 & 0.0373 & 2.45 \\
AdamW & 0.2442 & 0.2442 & 0.0808 & 0.0579 & 1.4 \\
SGD + momentum & 0.2805 & 0.2805 & 0.137 & 0.0447 & 3.07 \\
SGD & 0.3182 & 0.3182 & 0.133 & 0.0518 & 2.56 \\
Stochastic L-BFGS & 0.4855 & 0.4855 & 0.359 & 0 & -- \\
Shampoo & 0.6137 & 0.6137 & 0.166 & 0.0531 & 3.12 \\
\bottomrule
\end{tabular}
}

\end{table}

\begin{table}[!t]
  \centering
  \caption{Every method's selected configuration on the confirmation seeds, 1D regression (top) and random-polynomial regression (bottom). Rows are sorted by validation loss, so the row order is the ranking; the band is $\pm 1$ std over seeds.}
  \label{tab:confirm-synth}
  \resizebox{\linewidth}{!}{%
\begin{tabular}{lp{1.15in}rrr}
\toprule
\textbf{Method} & \parbox[t]{\linewidth}{\raggedright \textbf{Selected configuration}\strut} & \textbf{Val. loss} & \textbf{Test loss} & \textbf{Train (eval)} \\
\midrule
HIG & \parbox[t]{\linewidth}{\raggedright lr=0.005, tau=1e-06\strut} & $(1.775 \pm 0.99)\times 10^{-9}$ & $(1.886 \pm 1.1)\times 10^{-9}$ & $(1.811 \pm 1.0)\times 10^{-9}$ \\
Sven & \parbox[t]{\linewidth}{\raggedright k=32, lr=0.01, rtol=0.0001\strut} & $(5.086 \pm 15)\times 10^{-7}$ & $(5.358 \pm 16)\times 10^{-7}$ & $(5.075 \pm 15)\times 10^{-7}$ \\
SOAP & \parbox[t]{\linewidth}{\raggedright lr=0.001\strut} & $(3.372 \pm 3.1)\times 10^{-6}$ & $(3.347 \pm 3.1)\times 10^{-6}$ & $(3.370 \pm 3.1)\times 10^{-6}$ \\
SGD + momentum & \parbox[t]{\linewidth}{\raggedright lr=0.1\strut} & $(6.027 \pm 3.8)\times 10^{-6}$ & $(6.040 \pm 3.7)\times 10^{-6}$ & $(6.040 \pm 3.8)\times 10^{-6}$ \\
AdamW & \parbox[t]{\linewidth}{\raggedright lr=0.001, weight\_decay=0.01\strut} & $(1.735 \pm 1.7)\times 10^{-5}$ & $(1.736 \pm 1.8)\times 10^{-5}$ & $(1.738 \pm 1.7)\times 10^{-5}$ \\
KFAC & \parbox[t]{\linewidth}{\raggedright lr=0.001\strut} & $(1.802 \pm 1.9)\times 10^{-5}$ & $(1.795 \pm 1.9)\times 10^{-5}$ & $(1.798 \pm 1.9)\times 10^{-5}$ \\
MuonW & \parbox[t]{\linewidth}{\raggedright lr=0.001, weight\_decay=0.1\strut} & $(1.928 \pm 0.90)\times 10^{-5}$ & $(1.919 \pm 0.89)\times 10^{-5}$ & $(1.923 \pm 0.90)\times 10^{-5}$ \\
Muon & \parbox[t]{\linewidth}{\raggedright lr=0.001\strut} & $(2.159 \pm 2.0)\times 10^{-5}$ & $(2.178 \pm 2.1)\times 10^{-5}$ & $(2.165 \pm 2.0)\times 10^{-5}$ \\
Adam & \parbox[t]{\linewidth}{\raggedright lr=0.001\strut} & $(3.264 \pm 5.8)\times 10^{-5}$ & $(3.320 \pm 6.0)\times 10^{-5}$ & $(3.290 \pm 5.9)\times 10^{-5}$ \\
Polyak SGD & \parbox[t]{\linewidth}{\raggedright f\_star=0, max\_lr=1, eps=1e-08\strut} & $(8.792 \pm 11)\times 10^{-5}$ & $(8.806 \pm 11)\times 10^{-5}$ & $(8.748 \pm 11)\times 10^{-5}$ \\
Stochastic L-BFGS & \parbox[t]{\linewidth}{\raggedright lr=0.03, max\_iter=3, history\_size=10\strut} & $(2.659 \pm 2.9)\times 10^{-4}$ & $(2.667 \pm 2.9)\times 10^{-4}$ & $(2.670 \pm 3.0)\times 10^{-4}$ \\
RMSprop & \parbox[t]{\linewidth}{\raggedright lr=0.001\strut} & $(2.746 \pm 6.6)\times 10^{-4}$ & $(2.750 \pm 6.6)\times 10^{-4}$ & $(2.759 \pm 6.6)\times 10^{-4}$ \\
Shampoo & \parbox[t]{\linewidth}{\raggedright lr=0.1\strut} & $(2.956 \pm 4.5)\times 10^{-4}$ & $(3.016 \pm 4.7)\times 10^{-4}$ & $(2.991 \pm 4.7)\times 10^{-4}$ \\
JD (UPGrad) & \parbox[t]{\linewidth}{\raggedright lr=0.01, aggregator=UPGrad, inner\_optimizer=Adam\strut} & $(7.893 \pm 24)\times 10^{-4}$ & $(7.829 \pm 23)\times 10^{-4}$ & $(7.922 \pm 24)\times 10^{-4}$ \\
SGD & \parbox[t]{\linewidth}{\raggedright lr=0.3\strut} & $(1.084 \pm 4.0)\times 10^{-3}$ & $(1.083 \pm 3.9)\times 10^{-3}$ & $(1.095 \pm 4.0)\times 10^{-3}$ \\
\bottomrule
\end{tabular}
}

  \smallskip
  \resizebox{\linewidth}{!}{%
\begin{tabular}{lp{1.15in}rrr}
\toprule
\textbf{Method} & \parbox[t]{\linewidth}{\raggedright \textbf{Selected configuration}\strut} & \textbf{Val. loss} & \textbf{Test loss} & \textbf{Train (eval)} \\
\midrule
HIG & \parbox[t]{\linewidth}{\raggedright lr=0.1, tau=1e-08\strut} & 0.08406 $\pm$ 0.024 & 0.08028 $\pm$ 0.02 & 0.05783 $\pm$ 0.014 \\
Sven & \parbox[t]{\linewidth}{\raggedright k=16, lr=0.5, rtol=0.03\strut} & 0.1388 $\pm$ 0.055 & 0.136 $\pm$ 0.055 & 0.1106 $\pm$ 0.042 \\
JD (UPGrad) & \parbox[t]{\linewidth}{\raggedright lr=0.001, aggregator=UPGrad, inner\_optimizer=Adam\strut} & 0.1635 $\pm$ 0.049 & 0.1529 $\pm$ 0.044 & 0.1234 $\pm$ 0.038 \\
SOAP & \parbox[t]{\linewidth}{\raggedright lr=0.01\strut} & 0.1818 $\pm$ 0.05 & 0.1756 $\pm$ 0.042 & 0.1283 $\pm$ 0.031 \\
Polyak SGD & \parbox[t]{\linewidth}{\raggedright f\_star=0, max\_lr=1, eps=1e-08\strut} & 0.1997 $\pm$ 0.089 & 0.1924 $\pm$ 0.081 & 0.1457 $\pm$ 0.067 \\
Muon & \parbox[t]{\linewidth}{\raggedright lr=0.01\strut} & 0.207 $\pm$ 0.062 & 0.202 $\pm$ 0.058 & 0.1474 $\pm$ 0.04 \\
KFAC & \parbox[t]{\linewidth}{\raggedright lr=1e-05\strut} & 0.2118 $\pm$ 0.056 & 0.2013 $\pm$ 0.048 & 0.1626 $\pm$ 0.041 \\
RMSprop & \parbox[t]{\linewidth}{\raggedright lr=0.01\strut} & 0.2173 $\pm$ 0.056 & 0.2108 $\pm$ 0.047 & 0.1539 $\pm$ 0.032 \\
MuonW & \parbox[t]{\linewidth}{\raggedright lr=0.01, weight\_decay=0.1\strut} & 0.2313 $\pm$ 0.07 & 0.2226 $\pm$ 0.06 & 0.1873 $\pm$ 0.051 \\
Adam & \parbox[t]{\linewidth}{\raggedright lr=0.01\strut} & 0.2377 $\pm$ 0.087 & 0.2283 $\pm$ 0.079 & 0.1696 $\pm$ 0.061 \\
AdamW & \parbox[t]{\linewidth}{\raggedright lr=0.01, weight\_decay=0.01\strut} & 0.2442 $\pm$ 0.092 & 0.2403 $\pm$ 0.09 & 0.1872 $\pm$ 0.078 \\
SGD + momentum & \parbox[t]{\linewidth}{\raggedright lr=0.01\strut} & 0.2805 $\pm$ 0.13 & 0.2723 $\pm$ 0.11 & 0.2154 $\pm$ 0.1 \\
SGD & \parbox[t]{\linewidth}{\raggedright lr=0.1\strut} & 0.3182 $\pm$ 0.12 & 0.315 $\pm$ 0.13 & 0.259 $\pm$ 0.12 \\
Stochastic L-BFGS & \parbox[t]{\linewidth}{\raggedright lr=1, max\_iter=1, history\_size=2\strut} & 0.4855 $\pm$ 0.36 & 0.4485 $\pm$ 0.33 & 0.4123 $\pm$ 0.34 \\
Shampoo & \parbox[t]{\linewidth}{\raggedright lr=0.1\strut} & 0.6137 $\pm$ 0.15 & 0.6064 $\pm$ 0.13 & 0.5699 $\pm$ 0.14 \\
\bottomrule
\end{tabular}
}

\end{table}

\begin{table}[!t]
  \centering
  \caption{Every method's selected configuration on the confirmation seeds, MNIST label regression (top) and MNIST cross-entropy (bottom). Conventions as in Table~\ref{tab:confirm-synth}. K-FAC is absent from both: it has no eligible configuration on either scan (Appendix~\ref{app:robustness}).}
  \label{tab:confirm-mnist}
  \resizebox{\linewidth}{!}{%
\begin{tabular}{lp{1.15in}rrrr}
\toprule
\textbf{Method} & \parbox[t]{\linewidth}{\raggedright \textbf{Selected configuration}\strut} & \textbf{Val. loss} & \textbf{Test loss} & \textbf{Test acc.} & \textbf{Train (eval)} \\
\midrule
MuonW & \parbox[t]{\linewidth}{\raggedright lr=0.001, weight\_decay=0.1\strut} & 0.04982 $\pm$ 0.0013 & 0.04959 $\pm$ 0.0025 & 97.03\% & 0.02206 $\pm$ 0.00041 \\
HIG & \parbox[t]{\linewidth}{\raggedright lr=0.05, tau=0.03\strut} & 0.05298 $\pm$ 0.0026 & 0.05324 $\pm$ 0.0023 & 96.60\% & 0.02293 $\pm$ 0.0019 \\
Sven & \parbox[t]{\linewidth}{\raggedright k=64, lr=0.5, rtol=0.001\strut} & 0.05328 $\pm$ 0.0007 & 0.05363 $\pm$ 0.0024 & 96.72\% & 0.02804 $\pm$ 0.0012 \\
Muon & \parbox[t]{\linewidth}{\raggedright lr=0.001\strut} & 0.05473 $\pm$ 0.0017 & 0.05424 $\pm$ 0.0038 & 96.72\% & 0.01652 $\pm$ 0.0011 \\
SGD & \parbox[t]{\linewidth}{\raggedright lr=0.1\strut} & 0.05524 $\pm$ 0.0025 & 0.05439 $\pm$ 0.0019 & 96.89\% & 0.02968 $\pm$ 0.0019 \\
SGD + momentum & \parbox[t]{\linewidth}{\raggedright lr=0.01\strut} & 0.05524 $\pm$ 0.0015 & 0.05454 $\pm$ 0.0017 & 96.95\% & 0.02863 $\pm$ 0.0011 \\
Polyak SGD & \parbox[t]{\linewidth}{\raggedright f\_star=0, max\_lr=1, eps=1e-08\strut} & 0.05532 $\pm$ 0.004 & 0.05694 $\pm$ 0.0031 & 96.59\% & 0.02319 $\pm$ 0.0026 \\
AdamW & \parbox[t]{\linewidth}{\raggedright lr=0.001, weight\_decay=0.01\strut} & 0.05784 $\pm$ 0.0048 & 0.05613 $\pm$ 0.004 & 96.69\% & 0.02526 $\pm$ 0.0029 \\
Adam & \parbox[t]{\linewidth}{\raggedright lr=0.001\strut} & 0.05844 $\pm$ 0.0055 & 0.057 $\pm$ 0.004 & 96.59\% & 0.02432 $\pm$ 0.0033 \\
JD (UPGrad) & \parbox[t]{\linewidth}{\raggedright lr=0.001, aggregator=UPGrad, inner\_optimizer=Adam\strut} & 0.05902 $\pm$ 0.0025 & 0.05738 $\pm$ 0.0042 & 96.46\% & 0.02461 $\pm$ 0.0029 \\
RMSprop & \parbox[t]{\linewidth}{\raggedright lr=0.001\strut} & 0.05989 $\pm$ 0.0015 & 0.05897 $\pm$ 0.0018 & 96.43\% & 0.02781 $\pm$ 0.0024 \\
Stochastic L-BFGS & \parbox[t]{\linewidth}{\raggedright lr=0.1, max\_iter=3, history\_size=2\strut} & 0.06127 $\pm$ 0.0025 & 0.0609 $\pm$ 0.0026 & 96.64\% & 0.04508 $\pm$ 0.0025 \\
Shampoo & \parbox[t]{\linewidth}{\raggedright lr=0.1\strut} & 0.0804 $\pm$ 0.00099 & 0.08082 $\pm$ 0.0012 & 95.79\% & 0.0612 $\pm$ 0.0018 \\
SOAP & \parbox[t]{\linewidth}{\raggedright lr=0.01\strut} & 0.5808 $\pm$ 0.44 & 0.5732 $\pm$ 0.45 & 44.45\% & 0.555 $\pm$ 0.47 \\
\bottomrule
\end{tabular}
}

  \smallskip
  \resizebox{\linewidth}{!}{%
\begin{tabular}{lp{1.15in}rrrr}
\toprule
\textbf{Method} & \parbox[t]{\linewidth}{\raggedright \textbf{Selected configuration}\strut} & \textbf{Val. loss} & \textbf{Test loss} & \textbf{Test acc.} & \textbf{Train (eval)} \\
\midrule
MuonW & \parbox[t]{\linewidth}{\raggedright lr=0.0001, weight\_decay=0.1\strut} & 0.1046 $\pm$ 0.0038 & 0.1032 $\pm$ 0.0049 & 96.90\% & 0.06315 $\pm$ 0.0024 \\
Muon & \parbox[t]{\linewidth}{\raggedright lr=0.0001\strut} & 0.1067 $\pm$ 0.004 & 0.1054 $\pm$ 0.0052 & 96.88\% & 0.05821 $\pm$ 0.0023 \\
HIG & \parbox[t]{\linewidth}{\raggedright lr=0.05, tau=0.03\strut} & 0.1119 $\pm$ 0.0031 & 0.1094 $\pm$ 0.0039 & 96.73\% & 0.04896 $\pm$ 0.0032 \\
Sven & \parbox[t]{\linewidth}{\raggedright k=32, lr=0.5, rtol=0.3\strut} & 0.1134 $\pm$ 0.0051 & 0.1128 $\pm$ 0.0045 & 96.64\% & 0.05013 $\pm$ 0.0057 \\
SGD + momentum & \parbox[t]{\linewidth}{\raggedright lr=0.001\strut} & 0.1249 $\pm$ 0.0076 & 0.1206 $\pm$ 0.0052 & 96.22\% & 0.08399 $\pm$ 0.0071 \\
SGD & \parbox[t]{\linewidth}{\raggedright lr=0.01\strut} & 0.1258 $\pm$ 0.013 & 0.1219 $\pm$ 0.014 & 96.24\% & 0.08522 $\pm$ 0.014 \\
Stochastic L-BFGS & \parbox[t]{\linewidth}{\raggedright lr=0.5, max\_iter=1, history\_size=2\strut} & 0.1307 $\pm$ 0.0088 & 0.125 $\pm$ 0.0056 & 96.36\% & 0.08887 $\pm$ 0.012 \\
Shampoo & \parbox[t]{\linewidth}{\raggedright lr=0.1\strut} & 0.1317 $\pm$ 0.0031 & 0.1271 $\pm$ 0.0063 & 96.13\% & 0.06502 $\pm$ 0.005 \\
AdamW & \parbox[t]{\linewidth}{\raggedright lr=0.001, weight\_decay=0.01\strut} & 0.1368 $\pm$ 0.0027 & 0.1285 $\pm$ 0.0063 & 96.64\% & 0.03307 $\pm$ 0.0032 \\
Adam & \parbox[t]{\linewidth}{\raggedright lr=0.0001\strut} & 0.1456 $\pm$ 0.0068 & 0.145 $\pm$ 0.0081 & 95.73\% & 0.1221 $\pm$ 0.0063 \\
Polyak SGD & \parbox[t]{\linewidth}{\raggedright f\_star=0, max\_lr=1, eps=1e-08\strut} & 0.1471 $\pm$ 0.0015 & 0.145 $\pm$ 0.0088 & 96.81\% & 0.02024 $\pm$ 0.003 \\
RMSprop & \parbox[t]{\linewidth}{\raggedright lr=0.0001\strut} & 0.153 $\pm$ 0.01 & 0.1496 $\pm$ 0.0093 & 95.51\% & 0.1301 $\pm$ 0.01 \\
JD (UPGrad) & \parbox[t]{\linewidth}{\raggedright lr=0.0001, aggregator=UPGrad, inner\_optimizer=Adam\strut} & 0.1591 $\pm$ 0.011 & 0.1568 $\pm$ 0.011 & 95.62\% & 0.122 $\pm$ 0.0086 \\
SOAP & \parbox[t]{\linewidth}{\raggedright lr=0.0001\strut} & 0.646 $\pm$ 1.1 & 0.6406 $\pm$ 1.1 & 77.88\% & 0.6019 $\pm$ 1.1 \\
\bottomrule
\end{tabular}
}

\end{table}

\section{Analyzing Singular-Value Spectra}
\label{app:spectra}

Sven's approximation is the truncation controlled by $k$ and \texttt{rtol}, so inspecting the spectrum of $M$ during training illuminates how Sven is working. In this appendix we analyze spectra and their interaction with $k$ / \texttt{rtol} in two settings: ``online'', which logs untruncated spectra periodically during training; and ``probe'' measurements, which use cached checkpoints and recompute Jacobians on a fixed held-out probe set. The online spectrum reflects states the optimizer saw during training, while the probe spectrum reflects the ``general problem'' -- a held-out test set -- rather than training instances. The results shown in this section expand upon Fig.~\ref{fig:spectrum_truncation} in the main text.

\subsection{The spectra}

\begin{figure}[!ht]
    \centering
    \includegraphics[width=\linewidth]{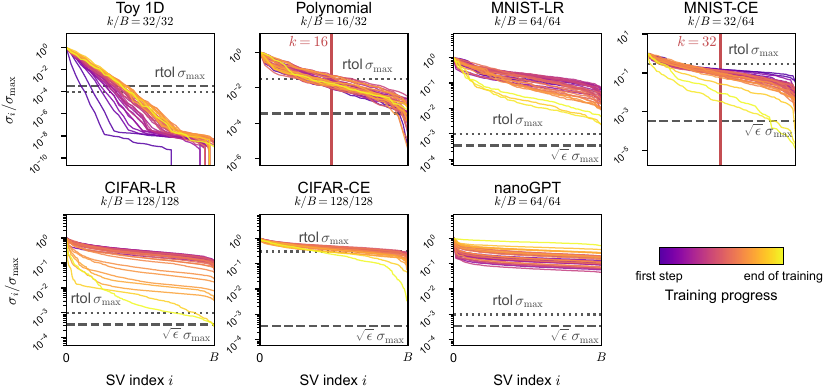}
    \caption{Full-width online Gram spectra over training (one model seed per scan). Each line is from one logged batch and normalized by its $\sigma_1$; the vertical line indicates $k$ when $k < B$, the dotted line the \texttt{rtol} cut, and the dashed line the float32 round-off floor.}
    \label{fig:online-spectra}
\end{figure}

Figure~\ref{fig:online-spectra} draws each scan's spectrum over training for the best-performing configuration, highlighting the broad differences in shape. On the 1D task it falls steeply and lies mostly below the \texttt{rtol} floor, becoming slightly less steep over the course of training. The polynomial spectrum is less steep and retains its shape. MNIST CE and label regression (LR) become steeper over training, but the best \texttt{rtol} setting means the spectrum is heavily truncated for CE and barely truncated for LR. A similar trend is true for CIFAR, though its CE spectrum is relatively stable. This suggests that retaining directions with lower SVs is more important for performance on regression problems than on cross-entropy, but it is unclear why.

\subsection{Effect of $k$ and \texttt{rtol}}

\paragraph{Truncation.} Two knobs restrict the update: $k$ and  \texttt{rtol}. Implemented as in Appendix~\ref{app:gram_trunc}, the number of directions the step inverts is
\begin{equation}
    \label{eqn:rank_law}
    r_{\rm used} \;=\; \min\big(k,\ \#\{i:\sigma_i > \texttt{rtol}\cdot\sigma_1\}\big).
\end{equation}
It follows that the best-performing $k$ cannot be read as a count of significant directions, because the spectra in Fig.~\ref{fig:online-spectra} indicate it is often \texttt{rtol} that controls $r_{\rm used}$ (see Table~\ref{tab:spectra-mechanism} for a detailed accounting). 
Whether $k$ or \texttt{rtol} truncates can also change during training: on CIFAR-10 label regression neither does at step 0 and \texttt{rtol} does at the end. A corollary for the tuning grids: a grid point whose binding cut is $k$ produces the same trajectory as every tighter \texttt{rtol} at the same $(k, \eta)$, and one whose binding cut is \texttt{rtol} the same as every larger $k$, so Sven's grid holds fewer distinct trajectories than points --- \numBudgetSvenDistinctFracMin{} to \numBudgetSvenDistinctFracMax{} of the nominal grid across the four MLP scans --- where a single-knob baseline has one trajectory per point.

\begin{figure}[!ht]
    \centering
    \includegraphics[width=\linewidth]{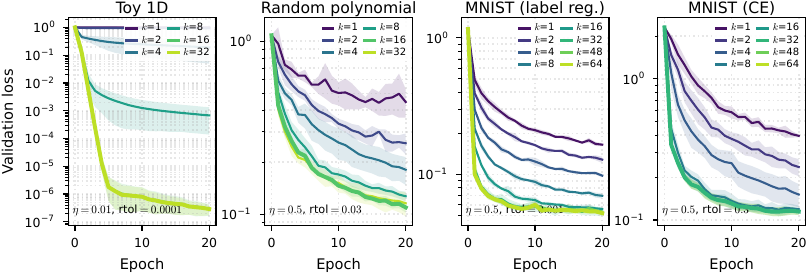}
    \caption{Validation loss against epoch over the rank cutoff $k$, with $\eta$ and \texttt{rtol} pinned at each scan's selected values; the thick curve is the selected $k$.}
    \label{fig:k_and_sv_plots}
\end{figure}

\begin{figure}[!ht]
    \centering
    \includegraphics[width=\linewidth]{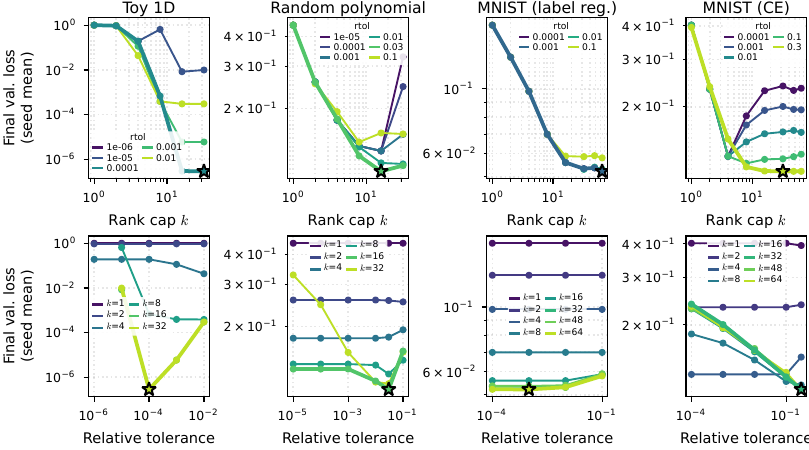}
    \caption{Mean final validation loss over $k$ at fixed \texttt{rtol} (top) and over \texttt{rtol} at fixed $k$ (bottom), both at each scan's selected learning rate; the star is the selection of record and ineligible configurations are omitted.}
    \label{fig:hparam_landscape}
\end{figure}

\paragraph{Sweeps} Fig.~\ref{fig:k_and_sv_plots} sweeps $k$ with $\eta$ and \texttt{rtol} pinned at each scan's selected values, and Fig.~\ref{fig:hparam_landscape} shows the seed-mean final loss over $k$ at fixed \texttt{rtol} and over \texttt{rtol} at fixed $k$. Improvement saturates at $k = \numToySvenKSaturate{}$ (1D), \numPolySvenKSaturate{} (polynomial), \numMnistLRSvenKSaturate{} (MNIST label regression) and \numMnistCESvenKSaturate{} (MNIST cross-entropy): a different fraction of $B$ on each scan corresponding to the point where \texttt{rtol} starts to truncate, past which raising $k$ changes nothing. The \texttt{rtol} panels read the same way against the spectra: on the 1D task, whose spectrum is steepest, the loss moves by orders of magnitude across the \texttt{rtol} grid; on MNIST LR, whose spectrum is flatter, most \texttt{rtol} lines coincide. On the polynomial task a looser cut \emph{helps}, up to the selected \texttt{rtol}, and the apparent preference for a large $k$ at a tight \texttt{rtol} is the interaction of the two cuts rather than the value of additional directions.

\begin{figure}[!ht]
    \centering
    \includegraphics[width=\linewidth]{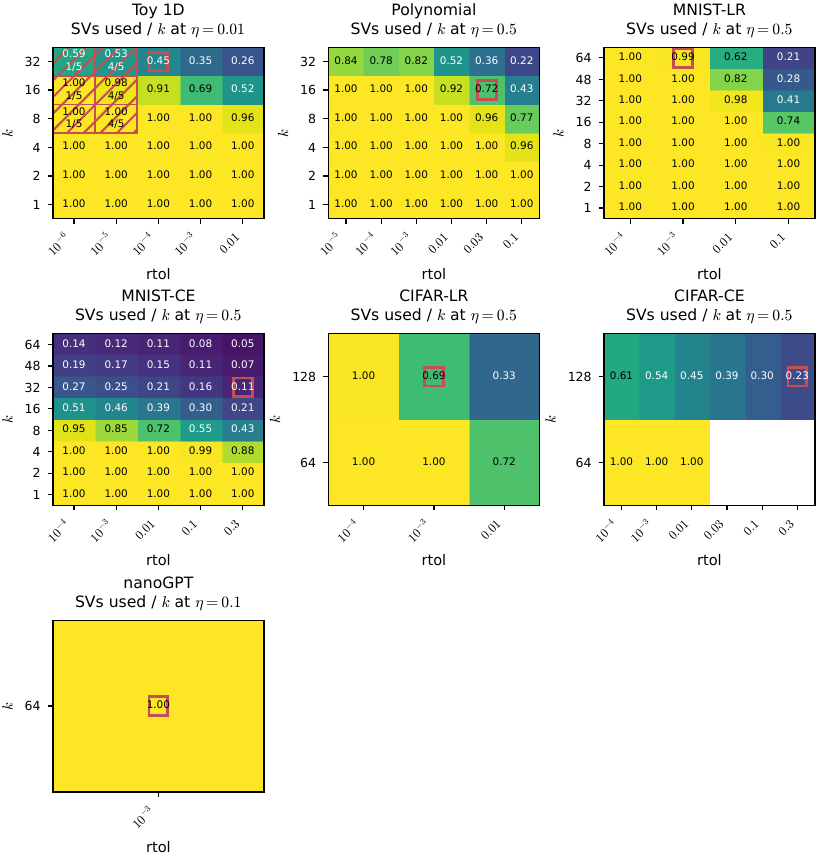}
    \caption{The rank the update uses as a fraction of the cap $k$, over the tuning $(k, \texttt{rtol})$ grid at each scan's selected learning rate, final epoch. A cell at $1$ is a cell where $k$ truncates; anything below it is a cell where \texttt{rtol} truncates. The red square is the selection of record and a hatched cell indicates fewer seeds than the grid intended (failed runs).}
    \label{fig:used-rank-grid}
\end{figure}

Fig.~\ref{fig:used-rank-grid} extends the same measurement to the whole tuning grid: at each $(k,\texttt{rtol})$ it shows the used rank as a fraction of $k$ at the selected learning rate, which makes the region where $k$ truncates and the region where \texttt{rtol} truncates visible at once.

\subsection{Effect of truncation}

\begin{figure}[!ht]
    \centering
    \includegraphics[width=\linewidth]{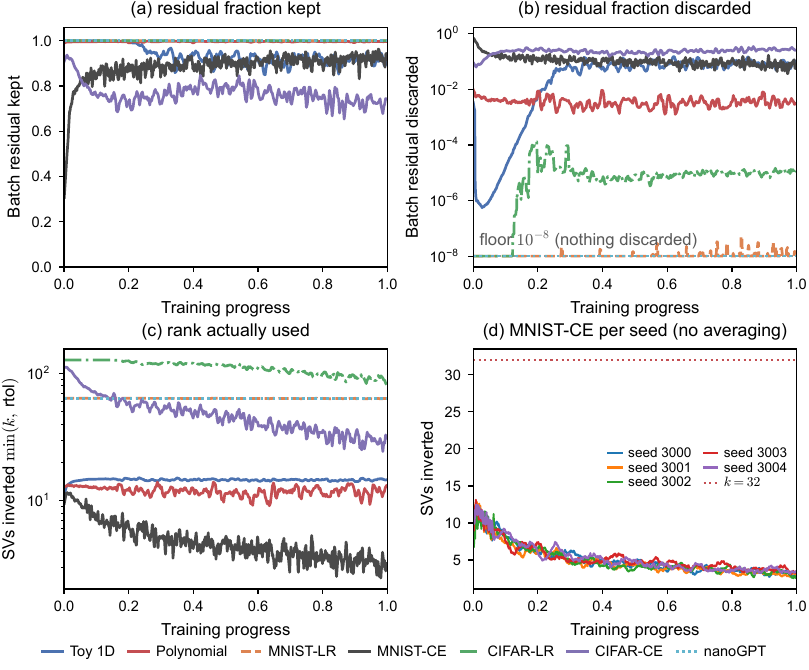}
    \caption{(a)--(c) Kept and discarded fraction of the batch residual and the rank actually inverted, against training progress, seed mean over the diagnostics seeds with a 60-step moving average. (d) The five seeds of the MNIST cross-entropy scan drawn separately: the collapse to $\sim 3$ directions happens on every seed.}
    \label{fig:online-mechanism}
\end{figure}

For each logged batch we compute the fraction of the residuals $\|\tilde{U}^T\mathcal{R}\|^2$ carried by the directions the truncation \textit{keeps}, and track it over training: Fig.~\ref{fig:online-mechanism}(a) draws it along with the used rank (c), and Table~\ref{tab:spectra-mechanism} gives the first- and last-step values. The fraction is consistently 100\% on MNIST LR, CIFAR LR, and nanoGPT, indicating that the discarded directions carry essentially none of the residual. MNIST LR and nanoGPT use the entire spectrum throughout training, while for the others \texttt{rtol} does reduce the rank. The truncation thus discards something tangible on some scans, but with an inconsistent effect on the fraction of retained residuals. It \emph{rises} on the polynomial scan (\numPolySvenEnergyFirst{}\% to \numPolySvenEnergyFinal{}\%) and MNIST CE (\numMnistCESvenEnergyFirst{}\% to \numMnistCESvenEnergyFinal{}\%), i.e.\ the residual concentrates into the directions that survive the truncation. It falls marginally on the 1D scan (\numToySvenEnergyFirst{}\% to \numToySvenEnergyFinal{}\%) and substantially on CIFAR CE (\numCifarCESvenEnergyFirst{}\% to \numCifarCESvenEnergyFinal{}\%), where by the end of training roughly a quarter of the batch residual lies in directions the update throws away. This effect is clearly dataset-dependent, and the scan where we truncate the highest residual fraction is where Sven performs the worst (CIFAR-10).

\begin{table}[t]
  \centering
  \caption{What the truncation does, per scan. ``Used'' is the number of directions actually inverted, $\min(k, \texttt{rtol}\text{-rank})$, at the first and last logged step; ``Kept'' is the fraction of that batch's residual they account for. ``Binds'' is read at the last logged step. Seed means over the diagnostics pass.}
  \label{tab:spectra-mechanism}
  \resizebox{\linewidth}{!}{%
\begin{tabular}{lrrccc}
\toprule
\textbf{Scan} & \textbf{$k$ / $B$} & \textbf{rtol} & \textbf{Binds} & \textbf{Used, first $\to$ final} & \textbf{Kept, first $\to$ final} \\
\midrule
Toy 1D & $32$ / $32$ & $10^{-4}$ & rtol & 4.2 $\to$ 14.4 & 92.08\% $\to$ 91.53\% \\
Polynomial & $16$ / $32$ & $0.03$ & rtol & 7.0 $\to$ 13.4 & 92.25\% $\to$ 99.59\% \\
MNIST-LR & $64$ / $64$ & $10^{-3}$ & neither ($k=B$) & 64.0 $\to$ 63.6 & 100.00\% $\to$ 100.00\% \\
MNIST-CE & $32$ / $64$ & $0.3$ & rtol & 10.0 $\to$ 2.8 & 31.01\% $\to$ 95.17\% \\
CIFAR-LR & $128$ / $128$ & $10^{-3}$ & rtol & 128.0 $\to$ 81.6 & 100.00\% $\to$ 100.00\% \\
CIFAR-CE & $128$ / $128$ & $0.3$ & rtol & 106.0 $\to$ 36.6 & 90.13\% $\to$ 76.69\% \\
nanoGPT & $64$ / $64$ & $10^{-3}$ & neither ($k=B$) & 64.0 $\to$ 64.0 & 100.00\% $\to$ 100.00\% \\
\bottomrule
\end{tabular}
}

\end{table}

\begin{figure}[!ht]
    \centering
    \includegraphics[width=\linewidth]{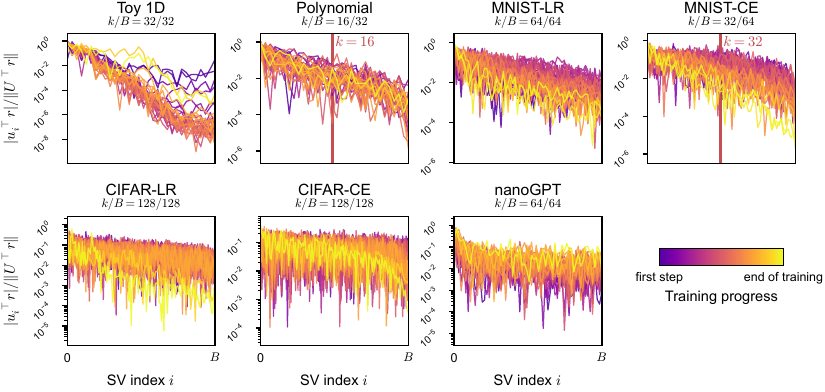}
    \caption{Where the batch residual sits: $|u_i\cdot R|/\|U^TR\|$ against singular-value index over training, one seed per panel.}
    \label{fig:online-utr}
\end{figure}

\subsection{Probe-Set Analysis}

As discussed above, we save checkpoints from the training trajectories of four optimizers (Sven, Adam, MuonW, and HIG), and compute double-precision Jacobians on a probe dataset and analyze their singular value spectra. The spectrum width is $\min(\text{probe rows}, P)$, and the settings for the four probed datasets are shown in Table~\ref{tab:spectra-probe-widths}. 

\begin{table}[t]
  \centering
  \caption{The cached probe sets. The spectrum width is $\min(\text{probe rows}, P)$ and only one of the two bounds is ever active. ``Resolved rank'' counts the singular values above the double-precision threshold on Sven's final checkpoint, averaged over 5 model seeds.}
  \label{tab:spectra-probe-widths}
  \begin{tabular}{lrrrr}
\toprule
\textbf{Scan} & \textbf{Probe rows} & \textbf{$P$} & \textbf{Spectrum width} & \textbf{Resolved rank, final step} \\
\midrule
Toy 1D & 10{,}000 & 593 & 593 & 45.0 \\
Random Polynomial & 10{,}000 & 673 & 673 & 673.0 \\
MNIST (label reg.) & 512 & 27{,}562 & 512 & 512.0 \\
MNIST (CE) & 512 & 27{,}562 & 512 & 512.0 \\
\bottomrule
\end{tabular}

\end{table}

\begin{figure}[!ht]
    \centering
    \includegraphics[width=\linewidth]{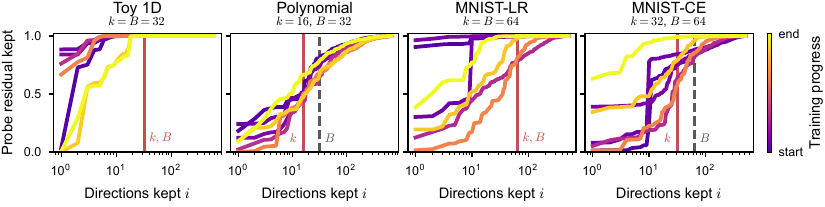}
    \caption{Cumulative fraction of the reachable probe residual against the number of leading probe directions kept, one seed per scan, colored by checkpoint. Two cuts are marked: $k$ is the rank a step inverts, $B$ the row space a step lives in, so even an untruncated step reaches no more than the curve's value at $B$. The $x$ axis runs to $\min(\text{probe rows}, P)$.}
    \label{fig:probe-energy}
\end{figure}

\paragraph{Probe-set Residuals} Fig.~\ref{fig:probe-energy} and Table~\ref{tab:spectra-probe-energy} summarize the probe results for Sven: the fraction of the reachable probe residual that lies in the leading directions of the probe Jacobian, with the selected $k$ marked. On the 1D task and both MNIST scans the top-$k$ directions carry nearly all of it; on the polynomial task they carry around 50\%. In the online measurement, which measured the kept fraction of a \textit{batch-level} residual, we saw that the selected $k$ kept essentially 100\% of the residual. The probe measurement answers a different question: how much of the whole problem's residual any rank-$k$ update can address. In this case, the batch size is the upper truncation limit, as the step lives in the batch's row space, so even at $k=B$ it can reach no more than the curve's value at $B$. The top-$k$ directions of the probe Jacobian are also the best $k$ for the \textit{whole} set, not the directions one might obtain from a randomly sampled batch. The curve in Fig.~\ref{fig:probe-energy} at $k$ is thus an upper bound on what a step addresses. What the probe measurement establishes is therefore not that the truncation discards signal at each step but that a rank-$k$, batch-$B$ update is a local correction to a residual that generally needs more directions to describe. This is most evident on the polynomial task, which is also where Sven has a larger margin over the first-order optimizers.

\begin{table}[t]
  \centering
  \caption{Residual fraction inside the top-$k$ probe directions, offline and in double precision, averaged over 5 model seeds; fractions are of the part of the probe residual the full Jacobian column space can reach, which is $\ge 99.5\%$ on every scan. An asterisk marks the selected $k$.}
  \label{tab:spectra-probe-energy}
  \svnfittable{\begin{tabular}{llrrr}
\toprule
\textbf{Scan} & \textbf{Checkpoint} & \textbf{$k$} & \textbf{Top-$k$ fraction (\%)} & \textbf{Seed std (\%)} \\
\midrule
Toy 1D & first & 8 & 99.6 & 0.2 \\
Toy 1D & first & 16 & 100.0 & 0.0 \\
Toy 1D & first & 32\,$^{\ast}$ & 100.0 & 0.0 \\
\midrule
Toy 1D & last & 8 & 64.2 & 13.0 \\
Toy 1D & last & 16 & 76.9 & 10.6 \\
Toy 1D & last & 32\,$^{\ast}$ & 100.0 & 0.0 \\
\midrule\midrule
Random Polynomial & first & 8 & 36.9 & 4.3 \\
Random Polynomial & first & 16\,$^{\ast}$ & 55.1 & 5.9 \\
Random Polynomial & first & 32 & 79.1 & 1.2 \\
\midrule
Random Polynomial & last & 8 & 43.6 & 13.8 \\
Random Polynomial & last & 16\,$^{\ast}$ & 58.8 & 10.2 \\
Random Polynomial & last & 32 & 73.7 & 5.9 \\
\midrule\midrule
MNIST (label reg.) & first & 8 & 91.8 & 2.7 \\
MNIST (label reg.) & first & 16 & 99.7 & 0.2 \\
MNIST (label reg.) & first & 32 & 99.9 & 0.1 \\
MNIST (label reg.) & first & 64\,$^{\ast}$ & 99.9 & 0.0 \\
\midrule
MNIST (label reg.) & last & 8 & 60.9 & 11.7 \\
MNIST (label reg.) & last & 16 & 78.6 & 14.7 \\
MNIST (label reg.) & last & 32 & 93.8 & 8.0 \\
MNIST (label reg.) & last & 64\,$^{\ast}$ & 99.9 & 0.1 \\
\midrule\midrule
MNIST (CE) & first & 8 & 4.7 & 1.5 \\
MNIST (CE) & first & 16 & 64.9 & 6.1 \\
MNIST (CE) & first & 32\,$^{\ast}$ & 76.5 & 1.8 \\
MNIST (CE) & first & 64 & 83.9 & 1.2 \\
\midrule
MNIST (CE) & last & 8 & 79.8 & 9.9 \\
MNIST (CE) & last & 16 & 89.0 & 7.5 \\
MNIST (CE) & last & 32\,$^{\ast}$ & 96.1 & 3.2 \\
MNIST (CE) & last & 64 & 99.6 & 0.2 \\
\bottomrule
\end{tabular}
}
\end{table}

\begin{figure}[!ht]
    \centering
    \includegraphics[width=\linewidth]{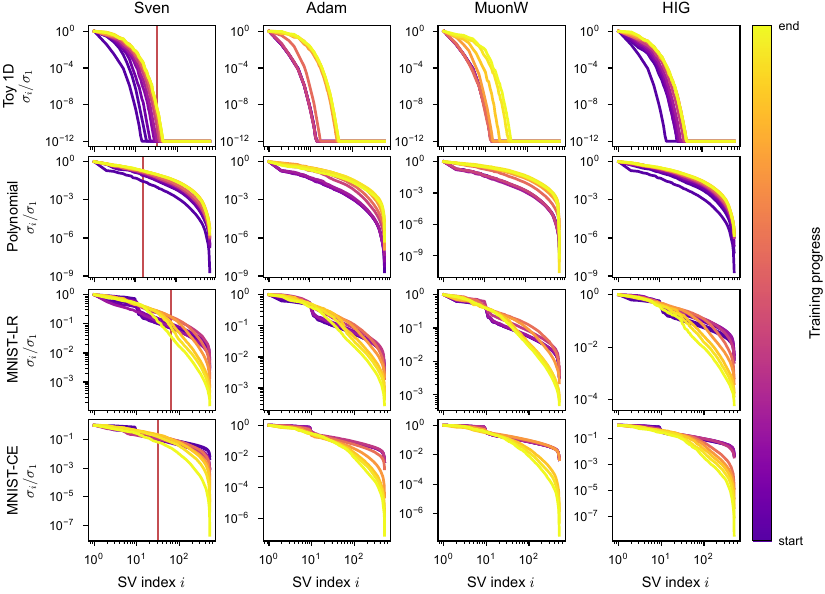}
    \caption{Probe-set spectra in double precision along the trajectories of four optimizers (columns) on every cached scan (rows), using the same probe rows at every checkpoint/optimizer. The red line is Sven's $k$. The spectrum width is $\min(\text{probe rows}, P)$.}
    \label{fig:probe-spectra-all}
\end{figure}

\begin{figure}[!ht]
    \centering
    \includegraphics[width=\linewidth]{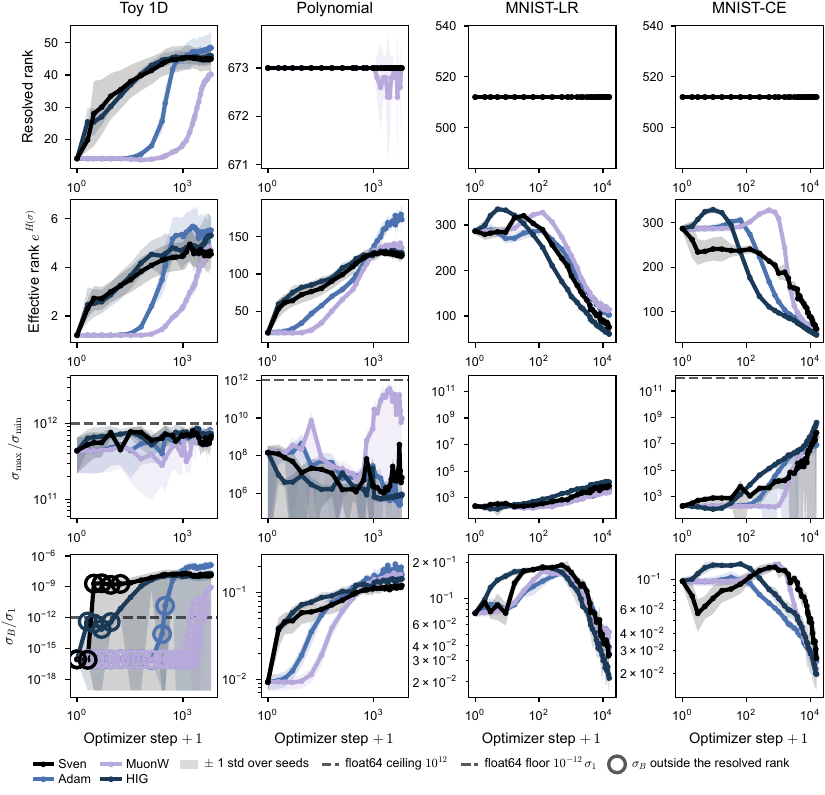}
    \caption{Resolved rank, effective rank, condition number and $\sigma_B/\sigma_1$ of the probe Jacobian along the trajectories of Sven, Adam, MuonW and HIG, seed mean $\pm$ 1 std over five cached seeds. The condition number is taken over the resolved part of the spectrum and cannot exceed \numSpectraProbeCeiling{}, the inverse of the float64 threshold; a hollow marker means index $B$ lies outside the resolved rank, so the ratio beside it is a threshold artifact. On the 1D task only \numToySvenProbeRankFirst{} of \numToySvenProbeWidth{} directions are resolved at initialization, which is why that scan's condition number and $\sigma_B/\sigma_1$ are artifacts throughout.}
    \label{fig:probe-metrics-all}
\end{figure}

\begin{figure}[!ht]
    \centering
    \includegraphics[width=\linewidth]{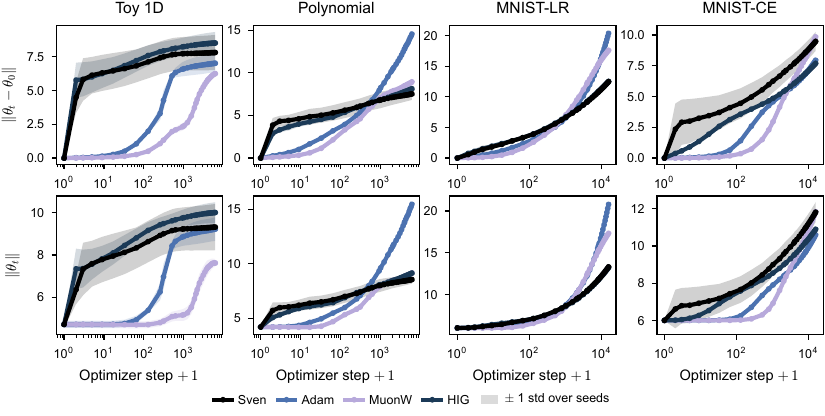}
    \caption{Distance from initialization and parameter norm along the four cached trajectories, seed mean $\pm$ 1 std.}
    \label{fig:probe-norms}
\end{figure}

\paragraph{Other optimizers} Fig.~\ref{fig:probe-spectra-all} shows the probe spectra along the trajectories of Sven, Adam, MuonW and HIG on every cached scan, one named seed per scan. Fig.~\ref{fig:probe-metrics-all} plots the resolved rank, the effective rank, the condition number and $\sigma_B/\sigma_1$ against training step for every cached scan, and Fig.~\ref{fig:probe-norms} the distance from initialization and the parameter norm. Two notions of rank appear in these figures and tables. The \emph{resolved rank} is the number of singular values above \numSpectraProbeFloor{}$\,\sigma_{\max}$, the limit for a float64 spectrum computation; it is a count at a threshold, and the condition number is taken over the resolved part only. The \emph{effective rank} is the entropy rank $\exp(-\sum_i p_i \log p_i)$ with $p_i = \sigma_i / \sum_j \sigma_j$: it equals $r$ for a flat spectrum of $r$ equal singular values and falls smoothly as the spectrum concentrates. For a very hierarchical spectrum, it is far below the resolved rank.

We observe that for the toy 1D dataset, the resolved rank is much smaller than $\min(\text{probe rows}, P)$ and that the condition number and $\sigma_B/\sigma_1$ get very near the float64 cutoff. This reflects the problem's strongly hierarchical spectrum. For the other datasets, the resolved rank typically equals $\min(\text{probe rows}, P)$. In all cases the effective rank is significantly smaller, reflecting varying degrees of hierarchy in the spectra. The trends differ for 1D/polynomial versus MNIST, however: the effective rank trends upwards in the synthetic task, indicating a flattening spectrum, while it trends down for both MNIST tasks, reflecting steepening. The overall trends, measured by beginning and endpoint values, are remarkably similar for all the optimizers.

Fig.~\ref{fig:probe-norms} examines the impact of Sven's update being the minimum-norm solution of the linearized system at each step, which invites the conjecture that it might stay closer to its initialization, or reach a smaller-norm solution than a first-order method. This does not appear to be true in general, with Sven's parameter norm ending larger than Adam's in some cases and smaller in others. Sven's parameter norm does tend to grow faster early in training, however. Table~\ref{tab:spectra-low4} compares Sven's parameter norm with other baselines at the last saved checkpoint. It lists Sven's mean parameter norm over seeds, compares it to the best (smallest) norm obtained by other optimizers, and shows the mean norm difference between Sven and the best other baseline. As in Fig.~\ref{fig:probe-norms}, the results are inconclusive, but it is clear that Sven does not generally result in a lower parameter norm. 

\begin{table}[t]
  \centering
  \caption{Parameter norm change (top panel) and absolute value (bottom) for Sven and the lowest of the other three optimizers, paired by model seed. A bold paired difference is one whose 95\% interval excludes zero.}
  \label{tab:spectra-low4}
  \svnfittable{\begin{tabular}{lrlcrcc}
\toprule
\textbf{Scan} & \textbf{Sven} & \textbf{Best other mean} & \textbf{Rank} & \textbf{Paired diff.} & \textbf{95\% interval} & \textbf{Seeds lower} \\
\midrule
\multicolumn{7}{l}{\textit{$\|\theta_t-\theta_0\|$}} \\
Toy 1D & 7.82 $\pm$ 1.3 & 6.259 (MuonW) & 3 of 4 & 1.56 $\pm$ 1.4 & $[-0.196, 3.32]$ & 0/5 \\
Random Polynomial & 7.524 $\pm$ 0.71 & 8.12 (HIG) & 1 of 4 & -0.597 $\pm$ 0.58 & $[-1.32, 0.127]$ & 4/5 \\
MNIST (label reg.) & 12.48 $\pm$ 0.27 & 12.52 (HIG) & 1 of 4 & -0.0423 $\pm$ 0.16 & $[-0.239, 0.155]$ & 3/5 \\
MNIST (CE) & 9.475 $\pm$ 0.77 & 7.69 (HIG) & 3 of 4 & \textbf{1.79 $\boldsymbol{\pm}$ 0.8} & $[0.786, 2.78]$ & 0/5 \\
\midrule
\multicolumn{7}{l}{\textit{$\|\theta_t\|$}} \\
Toy 1D & 9.316 $\pm$ 1.1 & 7.62 (MuonW) & 3 of 4 & \textbf{1.7 $\boldsymbol{\pm}$ 1.2} & $[0.263, 3.13]$ & 0/5 \\
Random Polynomial & 8.545 $\pm$ 0.37 & 9.134 (HIG) & 1 of 4 & \textbf{-0.59 $\boldsymbol{\pm}$ 0.27} & $[-0.923, -0.256]$ & 5/5 \\
MNIST (label reg.) & 13.33 $\pm$ 0.21 & 13.24 (HIG) & 2 of 4 & 0.0848 $\pm$ 0.18 & $[-0.142, 0.311]$ & 1/5 \\
MNIST (CE) & 11.8 $\pm$ 0.53 & 10.59 (Adam) & 3 of 4 & \textbf{1.21 $\boldsymbol{\pm}$ 0.57} & $[0.501, 1.92]$ & 0/5 \\
\bottomrule
\end{tabular}
}
\end{table}

\section{Paired Comparisons}
\label{app:budget}

\begin{figure}[!ht]
    \centering
    \includegraphics[width=\linewidth]{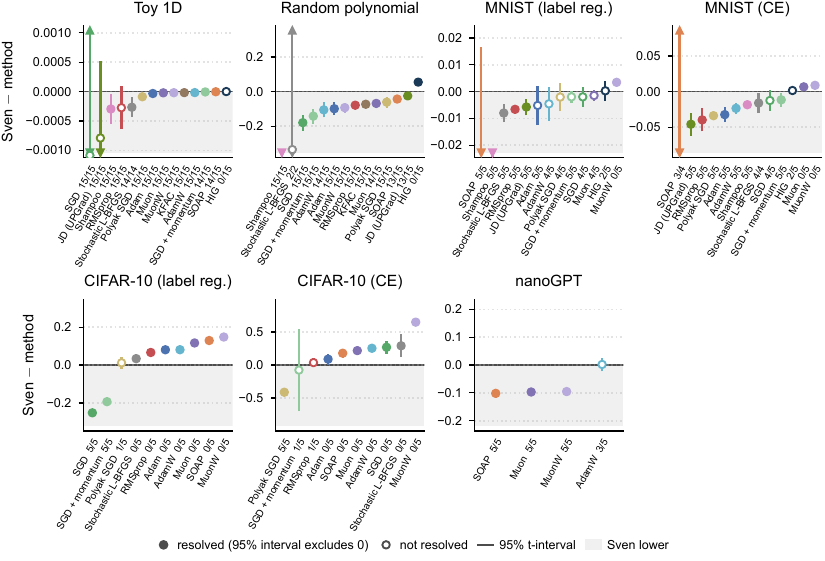}
    \caption{Paired per-seed differences in final validation loss. Differences are taken within a pair (same model seed, and data seed on the two synthetic tasks). Error bars are Student-$t$ 95\% intervals on the mean difference and a filled marker is an interval that excludes zero. The count beside each name is the pairs Sven won out of the pairs that finished. An interval wider than the panel is clipped and its end drawn as an arrow.}
    \label{fig:paired}
\end{figure}

This appendix reports the per-seed paired differences between Sven and baseline optimizers.
Every optimizer at a given model seed starts from the same initialization and sees the same batches in the same order, so the difference between two methods can be taken within a seed, which removes the seed-to-seed variation. Fig.~\ref{fig:paired} shows the mean paired difference in final validation between Sven and the baseline optimizers, along with its 95\% $t$-interval. This elucidates where Sven truly is a better or worse choice than the baselines, and supports the observations in the main text from Figs.~\ref{fig:main_results} and \ref{fig:paired_main}.

\section{Dataset-Level Overparametrization}
\label{app:overparam}

Section~\ref{sec:meth} distinguishes two notions of overparametrization: $P > B$, which holds everywhere, and $P > N$, a property of the dataset. The theory that motivates Sven concerns the singularity of the natural-gradient metric, which is a $P$ versus $N$ statement. In this appendix we explicitly vary $P/N$, where $N$ is the size of the training set (the validation and test set fixed).

We study 1D regression, polynomial regression, and MNIST label regression at varying levels of $P/N$ spanning under- and overparametrized. The synthetic datasets are trained full-batch, while MNIST subsamples the training set and keeps its batch size from earlier experiments\footnote{The MNIST MLP is much larger than the MLP for the synthetic datasets, so full-batch training is not feasible in e.g.\ the overparametrized setting.}. Thirteen optimizers\footnote{Jacobian Descent and K-FAC are omitted due to previously observed training instability.} are trained for 200 epochs on a similar hyperparameter grid to the studies in the main text, with best configurations selected via validation loss.

\begin{figure}[t]
    \centering
    \includegraphics[width=\linewidth]{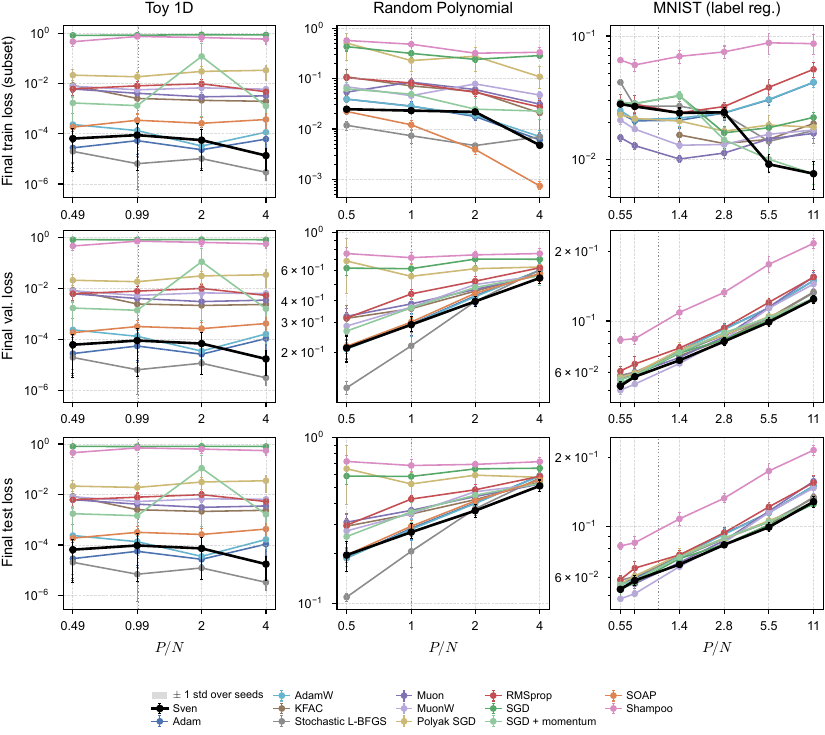}
    \caption{Final loss on the training subset (top), validation (middle) and test (bottom) against $P/N$ for thirteen methods on the 1D, polynomial and MNIST label-regression tasks, with the configuration selected inside each arm on validation loss and a band of $\pm 1$ std over seeds. SOAP is omitted from the MNIST panels, as in Fig.~\ref{fig:main_results}.}
    \label{fig:overparam}
\end{figure}

Fig.~\ref{fig:overparam} plots train, validation, and test loss versus $P/N$ for the best-performing configuration of each optimizer. Sven generally follows the same patterns as the baselines and more or less maintains its relative rank; there is no clear evidence that it has an advantage in dataset-level overparametrized settings.

\section{Batch-Size Sensitivity}
\label{app:batchsize}

Sven is clearly sensitive to batch size, as it controls the rank of the Jacobian used in computing the step. In Appendix~\ref{app:spectra} we observed that the batch-size-limited rank is sometimes significantly smaller than the rank of the residuals evaluated on a large test set, so it may be beneficial to run Sven on as large a batch size as possible. In this appendix we study this dependence, sweeping batch size and comparing Sven to other baselines at each setting. We focus on polynomial regression and scan over $B = \numBatchsizeGrid{}$ at a fixed \numPolyEpochs{} training epochs, tuning baselines at each $B$ as before. We omit HIG and JD here, focusing mainly on Sven's performance relative to the more standard baselines.

\begin{figure}[t]
    \centering
    \includegraphics[width=0.55\linewidth]{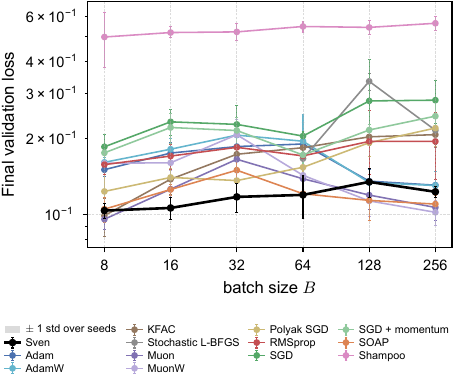}
    \caption{Batch-size sweep on random-polynomial regression: selected final validation loss against $B$ for. Error bars are $\pm 1$ std over seeds.}
    \label{fig:batchsize}
\end{figure}

Figure~\ref{fig:batchsize} shows the best-performing validation loss versus batch size for Sven and the other baselines. Almost no optimizers have a monotonic trend, Sven included. This suggests that batch size is not particularly influential in overall performance for the polynomial dataset, despite what Fig.~\ref{fig:probe-energy} suggests about the rank gap between the batch-constrained residuals and the full-dataset residuals.

\section{Classification with Cross-Entropy}
\label{app:mnist_ce}

The headline MNIST experiment in Sec.~\ref{sec:experiments} uses a label-regression loss, $L(\mathbf{x}_i;\theta) = \|f_\theta(\mathbf{x}_i) - \mathbf{y}_i\|^2$, which puts it on the same footing as the synthetic datasets and is consistent with the mean-squared-error derivation of Sec.~\ref{sec:meth}. This appendix examines performance using cross-entropy loss (still with $\kappa = 2$). We run optimizer hyperparameter scans using the same models and protocols as the main experiments, described in Appendix~\ref{app:exp_details}.

\begin{figure}[t]
    \centering
    \includegraphics[width=\linewidth]{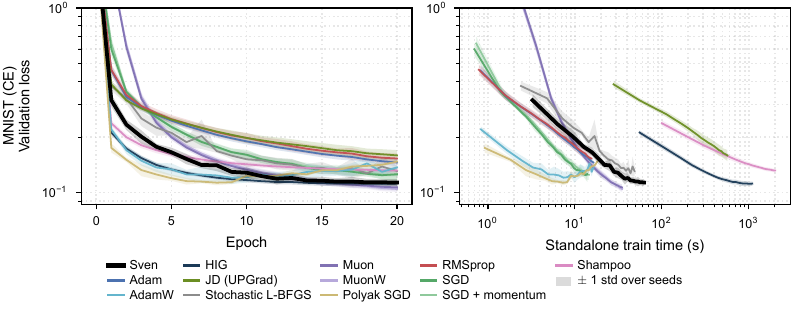}
    \caption{Confirmation-seed trajectories on MNIST with cross-entropy: validation loss against epoch (left) and against synchronized standalone training time (right); bands of $\pm 1$ std over seeds. SOAP is omitted, as in Fig.~\ref{fig:main_results}.}
    \label{fig:mnist_ce}
\end{figure}

Figure~\ref{fig:mnist_ce} plots validation loss versus epoch (left) and wall-time (right) for the best-performing optimizer settings, with wall-time again measured from a dedicated run. Sven performs strongly, outperforming most baselines except HIG and Muon, with HIG taking more than 10 times longer due to its heavier Jacobian computation. Sven thus appears to be competitive with popular baselines in the cross-entropy setting, and is not restricted to regression problems (this was already suggested by the nanoGPT results in Fig.~\ref{fig:main_results}). Referring back to Fig.~\ref{fig:online-spectra}, we can analyze MNIST cross-entropy vs label regression by looking at the Jacobian SV spectrum shapes during training (Fig.~\ref{fig:spectrum_truncation}(b) also overlays them at the end of training). The two roughly resemble each other at initialization, but over the course of training the CE spectrum becomes sharply hierarchical while the LR spectrum remains flatter (though it does become steeper). The CE spectrum is also aggressively truncated by the selected (best-performing) \texttt{rtol}, while the LR spectrum remains un-truncated throughout training. Since the two scans share everything but the loss, these differences are a property of the objective. A plausible conjecture is softmax saturation: once the model is confident, almost all per-sample gradients become small and nearly parallel, leaving few independent conditions in a batch. We do not study this in detail here, and leave it for future investigation.

\section{CIFAR-10 Experiments}
\label{app:cifar_10}

In this appendix we present results training Sven on CIFAR-10~\citep{krizhevsky2009learning} with a convolutional neural network classifier, where we find that it performs quite poorly. We train ResNet18~\citep{he2016deep} under both cross-entropy and label regression losses, and compare against 10 baselines\footnote{JD and HIG require output Jacobians and were omitted because they cannot scale to ResNet-18; K-FAC and Shampoo were omitted due to cost}. We train for 20 epochs with $B=\numCifarLRSvenB{}$ and the splits described in Appendix~\ref{app:datasets}. BatchNorm was found to be necessary to achieve good performance with \textit{any} optimizer, so we enable it for these studies. This prevents us from using the memory efficient Gram matrix capture for Sven, so we fall back to the costly full Jacobian capture described in Appendix~\ref{app:gram}. Sven is therefore much slower and more memory-hungry on CIFAR than on other baselines.

\begin{figure}[t]
    \centering
    \includegraphics[width=\linewidth]{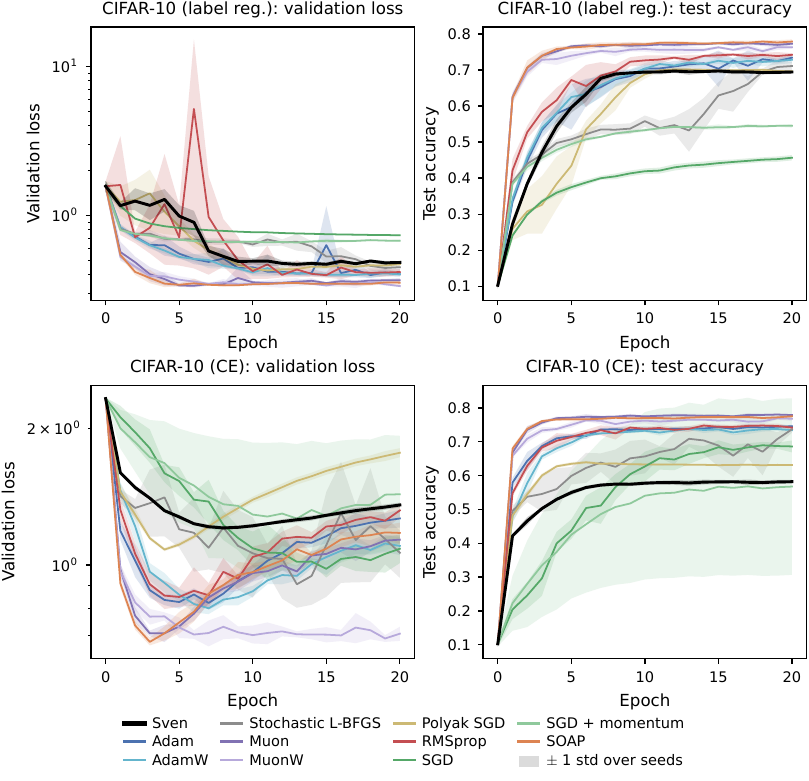}
    \caption{Validation loss (left) and test accuracy (right) for CIFAR-10 / ResNet18 with label regression (top) and cross-entropy (bottom).}
    \label{fig:cifar10_results}
\end{figure}

\begin{figure}[t]
    \centering
    \includegraphics[width=\linewidth]{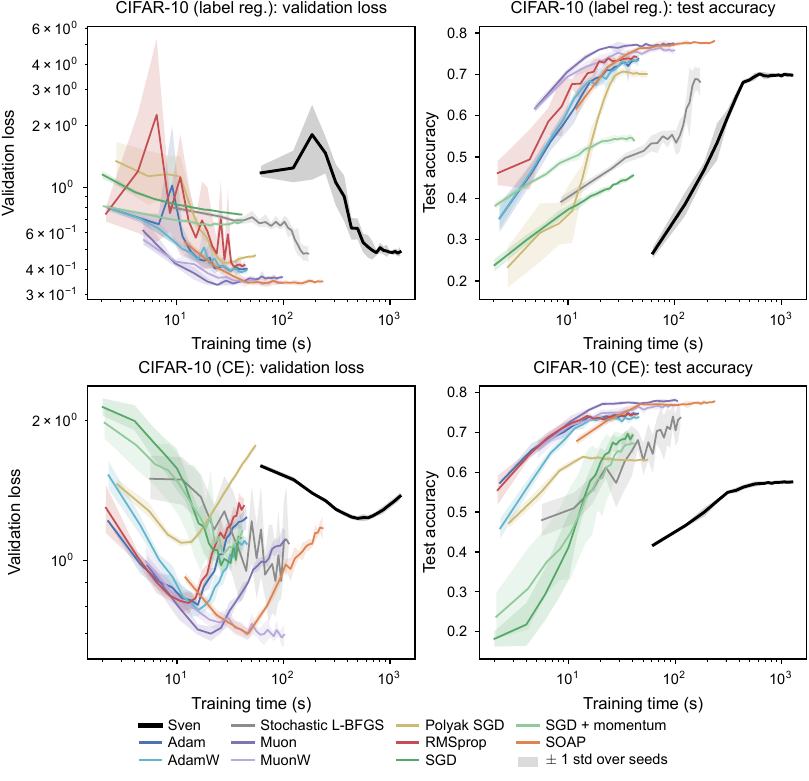}
    \caption{Validation loss (left) and test accuracy (right) plotted against training wall-time for CIFAR-10 / ResNet18 with label regression (top) and cross-entropy (bottom).}
    \label{fig:cifar_curves_time}
\end{figure}

\begin{figure}[t]
    \centering
    \includegraphics[width=\linewidth]{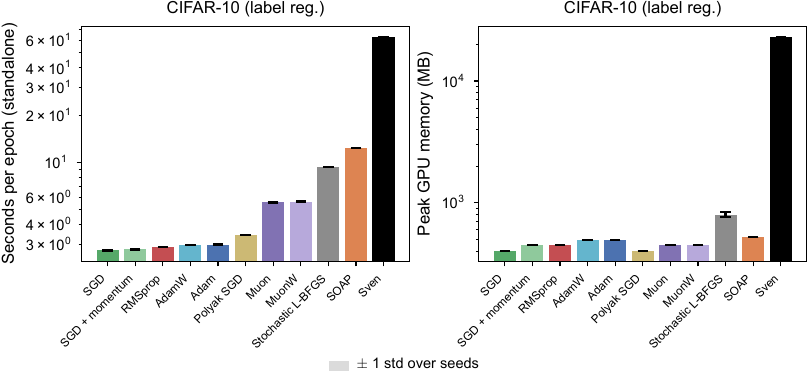}
    \caption{Seconds per epoch and peak GPU memory of each method's selected configuration on the label-regression scan.}
    \label{fig:cifar_cost}
\end{figure}

Table~\ref{tab:cifar} summarizes results for the best-performing optimizer variants on label regression and cross-entropy. Fig.~\ref{fig:cifar10_results} summarizes validation loss/accuracy as a function of epoch, while Fig.~\ref{fig:cifar_curves_time} plots them against wall time. Sven performs poorly in both settings, especially so for CE, though it has comparatively very low cross-seed spread in this case. Curiously, most methods overfit strongly for CE while they do not for LR. As expected, the wall-time plots show that Sven takes \textit{much} longer to train due to its inefficient Jacobian capture. Fig.~\ref{fig:cifar_cost} plots epoch wall-time and peak GPU memory, confirming this. Looking back to Fig.~\ref{fig:online-mechanism} in Appendix~\ref{app:spectra}, we recall that CIFAR-10 CE appears to discard a substantial fraction of the residual norm by the end of training, which may partially explain the degraded performance here\footnote{The same is not true of CIFAR label regression, so we hesitate to make a strong causal claim}.

\begin{table}[t]
  \centering
  \caption{CIFAR-10 summary table for label regression (top) and cross-entropy (bottom).}
  \label{tab:cifar}
  \svnfittable{\begin{tabular}{lrrrrrrr}
\toprule
\textbf{Objective} & \textbf{Method} & \textbf{Configuration} & \textbf{Val} & \textbf{Test} & \textbf{Val acc} & \textbf{Test acc} & \textbf{Train (eval)} \\
\midrule
CIFAR-10 (label reg.) & MuonW & lr=0.01, weight\_decay=0.1 & 0.3357 $\pm$ 0.005 & 0.3437 $\pm$ 0.0046 & 76.8\% & 76.3\% & 0.2069 \\
 & SOAP & lr=0.01 & 0.3542 $\pm$ 0.017 & 0.3625 $\pm$ 0.0094 & 78.3\% & 77.9\% & 0.03345 \\
 & Muon & lr=0.01 & 0.367 $\pm$ 0.013 & 0.3808 $\pm$ 0.015 & 78.0\% & 77.3\% & 0.03828 \\
 & AdamW & lr=0.01, weight\_decay=0.01 & 0.4027 $\pm$ 0.0075 & 0.4015 $\pm$ 0.011 & 72.6\% & 72.8\% & 0.1456 \\
 & Adam & lr=0.01 & 0.403 $\pm$ 0.015 & 0.4005 $\pm$ 0.011 & 73.1\% & 73.4\% & 0.1189 \\
 & RMSprop & lr=0.001 & 0.4173 $\pm$ 0.019 & 0.4165 $\pm$ 0.013 & 74.2\% & 74.2\% & 0.07515 \\
 & Stochastic L-BFGS & lr=2, max\_iter=3, history\_size=2 & 0.4499 $\pm$ 0.026 & 0.4509 $\pm$ 0.026 & 71.4\% & 71.1\% & 0.2051 \\
 & Polyak SGD & f\_star=0, max\_lr=1, eps=1e-08 & 0.472 $\pm$ 0.026 & 0.4687 $\pm$ 0.013 & 69.9\% & 69.6\% & 0.04596 \\
 & Sven & k=128, lr=0.5, rtol=0.001 & 0.4838 $\pm$ 0.011 & 0.4938 $\pm$ 0.0095 & 69.9\% & 69.4\% & 0.06737 \\
 & SGD + momentum & lr=0.001 & 0.6771 $\pm$ 0.013 & 0.6814 $\pm$ 0.0079 & 54.4\% & 54.5\% & 0.1879 \\
 & SGD & lr=0.001 & 0.7361 $\pm$ 0.0099 & 0.7336 $\pm$ 0.0091 & 44.9\% & 45.6\% & 0.6141 \\
\midrule
CIFAR-10 (CE) & MuonW & lr=0.01, weight\_decay=0.1 & 0.7054 $\pm$ 0.025 & 0.7073 $\pm$ 0.025 & 76.9\% & 76.8\% & 0.3106 \\
 & Stochastic L-BFGS & lr=2, max\_iter=2, history\_size=5 & 1.063 $\pm$ 0.13 & 1.09 $\pm$ 0.12 & 73.6\% & 73.7\% & 0.1519 \\
 & SGD & lr=3 & 1.085 $\pm$ 0.08 & 1.107 $\pm$ 0.074 & 68.9\% & 68.6\% & 0.3555 \\
 & AdamW & lr=0.01, weight\_decay=0.01 & 1.101 $\pm$ 0.05 & 1.108 $\pm$ 0.026 & 73.6\% & 73.5\% & 0.09618 \\
 & Muon & lr=0.01 & 1.137 $\pm$ 0.024 & 1.18 $\pm$ 0.034 & 78.2\% & 77.9\% & 0.03619 \\
 & SOAP & lr=0.01 & 1.176 $\pm$ 0.063 & 1.249 $\pm$ 0.062 & 78.5\% & 77.6\% & 0.02897 \\
 & Adam & lr=0.001 & 1.266 $\pm$ 0.058 & 1.304 $\pm$ 0.044 & 75.0\% & 74.6\% & 0.05469 \\
 & RMSprop & lr=0.001 & 1.318 $\pm$ 0.049 & 1.366 $\pm$ 0.045 & 74.2\% & 74.2\% & 0.06282 \\
 & Sven & k=128, lr=0.5, rtol=0.3 & 1.355 $\pm$ 0.016 & 1.339 $\pm$ 0.016 & 57.9\% & 58.2\% & 0.1586 \\
 & SGD + momentum & lr=0.3 & 1.431 $\pm$ 0.49 & 1.454 $\pm$ 0.48 & 56.8\% & 56.8\% & 0.6511 \\
 & Polyak SGD & f\_star=0, max\_lr=1, eps=1e-08 & 1.766 $\pm$ 0.041 & 1.779 $\pm$ 0.025 & 63.1\% & 63.2\% & 0.004623 \\
\bottomrule
\end{tabular}
}
\end{table}

\section{Parameter Masking at ResNet Scale}
\label{app:fig5}

In this appendix we explore parameter masking, introduced in Appendix~\ref{app:gram_knobs}, as a theoretical option for reducing memory overhead for large models like ResNet18 where the efficient Gram capture cannot be used. This does not yet work in practice because, as discussed above, autograd backends still instantiate full intermediate gradient tensors even when the requested fraction $f < 1$. We can still study how $f < 1$ affects optimization performance, which we explore here.

\begin{figure}[t]
    \centering
    \includegraphics[width=\linewidth]{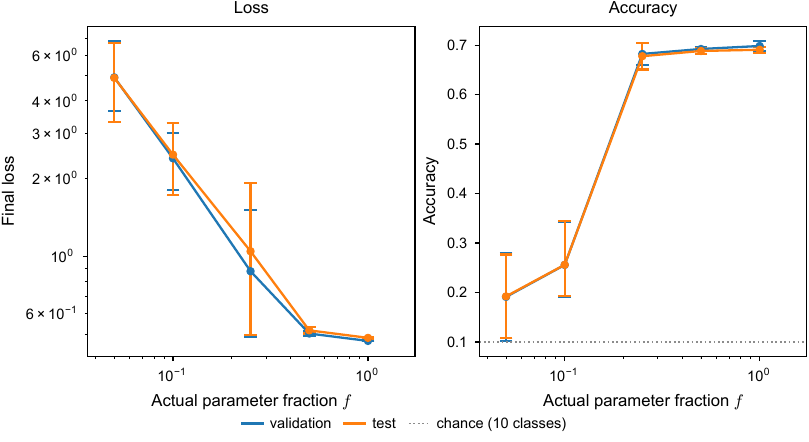}
    \caption{Validation/test loss (left) and accuracy (right) against the actual parameter fraction on ResNet18 / CIFAR-10 with label regression. Quality holds to $f \approx \numFigFiveHoldFrac{}$ and collapses at $f \le \numFigFiveCollapseFrac{}$.}
    \label{fig:fig5-quality}
\end{figure}

 We train CIFAR-10 + ResNet18 with label regression at $f \in \{0.05, 0.1, 0.25, 0.5, 1\}$, running three model seeds each. The parameter mask is an elementwise Bernoulli and is resampled at every training step; we choose a \textit{uniformly} random subset of parameters, not sub-units of the network (e.g. full weight matrices). All points use the best-performing Sven configuration from Appendix~\ref{app:cifar_10}. Fig.~\ref{fig:fig5-quality} shows validation loss and test accuracy as a function of the parameter fraction $f$. Surprisingly, performance is roughly the same at $f = 0.5$, and only modestly degraded in validation loss at $f = 0.25$ (accuracy remains flat). Accuracy drops fast below $f = 0.25$, but the results suggest that performance can be maintained with 75\% reduction in memory overhead.

\begin{figure}[t]
    \centering
    \includegraphics[width=\linewidth]{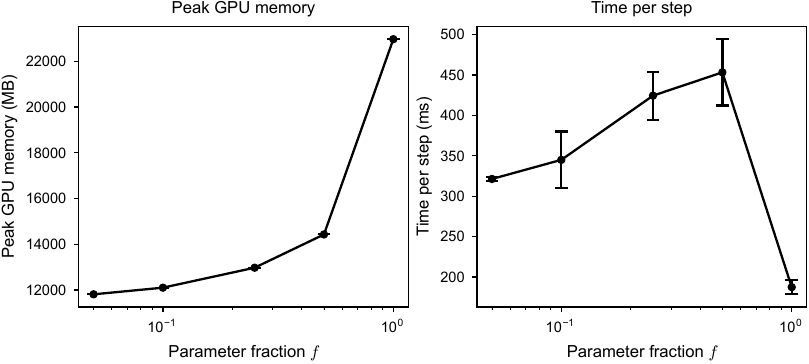}
    \caption{Peak memory (left) and time per optimizer step (right) against the actual parameter fraction; error bars are $\pm 1$ std over seeds.}
    \label{fig:fig5-cost}
\end{figure}

Fig.~\ref{fig:fig5-cost} plots peak memory usage and step time as a function of $f$. Peak memory does fall with $f$, but by less than 50\% and not following the theoretical $1/f$ profile. As stated above, this is because of how the Gram capture works here: ResNet18 must use the full capture, which is one reverse-mode pass over all $P$ parameters that materializes the $B \times P$ Jacobian before the mask is applied. The peak therefore has two parts: the backward pass component (activations + the Jacobian block) which is fixed at every $f$, and the mask-selected block that is gathered out and which scales with $f$. The former is the $\sim$12 GB floor the memory curve flattens onto. In the efficient capture with hooks, the floor does not exist and memory is proportional to the active fraction. 
Interestingly, step time \textit{increases} as $f$ is decreased.

\section{Language Modeling with Transformers}
\label{app:transformers}

This appendix studies Sven's performance on language modeling tasks: the nanoGPT tiny-shakespeare runs presented in the main paper, and a trial large-scale run on GPT-2-small. We observe poor performance on GPT-2-small, but with the major caveat that we did not have sufficient time or compute resources to run sensible hyperparameter sweeps or multiple seeds.

\subsection{nanoGPT on character-level Shakespeare}

\begin{figure}[t]
    \centering
    \includegraphics[width=\linewidth]{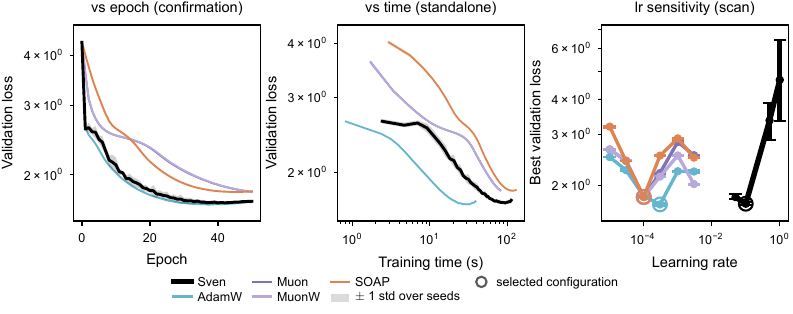}
    \caption{nanoGPT on character-level Shakespeare. Validation loss against epoch (left) and against training time (center),  $\pm 1$ std over 5 seeds; final validation loss against learning rate (right).}
    \label{fig:nanogpt}
\end{figure}

The training setup for the nanoGPT studies is described in Appendix~\ref{app:exp_details}. We compare five methods: AdamW, Muon, MuonW, SOAP, and Sven\footnote{We use a limited set of mainstream baselines for this study, as it is more computationally demanding.}, scanning hyperparameters for each and choosing the best-performing configuration. Sven is able to use the efficient \texttt{hooks} capture in this case. The results are summarized in Fig.~\ref{fig:nanogpt}, which shows Sven as a strong contender: it performs as well as AdamW and better than Muon or SOAP, while being about as time-efficient as Muon/SOAP. One significant caveat is that we only scan Sven's learning rate in this experiment (Fig.~\ref{fig:nanogpt}, right) due to resource constraints. It is possible that Sven performs even better under appropriate tuning, but this was beyond the scope of this study.

\subsection{GPT-2-small on FineWeb-edu}

\begin{figure}[t]
    \centering
    \includegraphics[width=\linewidth]{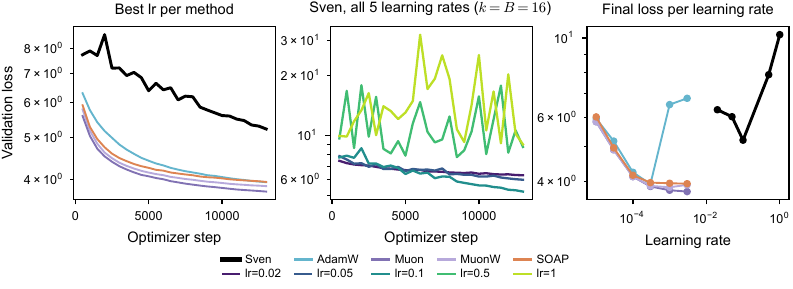}
    \caption{GPT-2-small on FineWeb-edu: validation loss against optimization step for the best learning rate of each method (left) and for each of Sven's learning rates (center), and the final validation loss against the learning rate for every method (right). Resource constraints prevented a full hyperparameter scan.}
    \label{fig:gpt2}
\end{figure}

In this study we train GPT-2-small~\citep{radford2019gpt2} for one epoch on FineWeb-edu~\citep{penedo2024fineweb} with $B = 16$ (see Appendix~\ref{app:exp_details}). We are again unable to perform a full hyperparameter scan due to resource constraints, so these results should be read as preliminary and subject to (potentially significant) change under re-tuning. Fig.~\ref{fig:gpt2} draws the trajectories and Table~\ref{tab:gpt2} gives each method's best run and its cost. Sven is clearly the worst performer, but it is unknown if this is due to poor hyperparameter settings. Further study is needed to truly assess its performance on large-scale language modeling. 

Interestingly, Sven's wall-time cost is \textit{significantly} higher on this task due to the structure of the language modeling task. The hooks capture of Appendix~\ref{app:gram_capture} adds each linear layer's contribution to the Gram as $(g g^{\top}) \odot (x x^{\top})$, which never forms a per-sample gradient --- but only when the layer's input has one row per sample. For a sequence model, the input is $B \times L \times d$ and the per-sample gradient is a sum over the $L$ positions. Our current implementation falls back to forming the flattened $B \times (d_{\rm out} d_{\rm in})$ per-sample gradient block explicitly and contracting it in chunks. This block is modestly sized for attention heads, but enormous for the output head (\texttt{lm\_head}): $d_{\rm out} d_{\rm in} = 50{,}304 \times 768 \approx 3.9 \times 10^{7}$. The head is thus captured one sample at a time and the off-diagonal Gram entries recompute their blocks ($\mathcal{O}(B^2)$ block formations). This sets the long step/epoch time and larger peak memory of Table~\ref{tab:gpt2}. The nanoGPT head is much smaller ($128 \times 65$) and does not cause the same problem. The fix is known but not used: the head's Gram factorizes over positions exactly as the 2D case does, $K_{ab} = \sum_{t,t'} (g_a[t] \cdot g_b[t'])(x_a[t] \cdot x_b[t'])$, and an implementation that contracts over token pairs without materializing the blocks removes the time and the memory cost. This was found to be slower at smaller model sizes and thus disabled, but will likely speed things up at GPT-2 scale.

\begin{table}[t]
  \centering
  \caption{Results summary table for GPT-2-small. Results use just one model seed due to resource constraints.}
  \label{tab:gpt2}
  \svnfittable{\begin{tabular}{lrrrrrr}
\toprule
\textbf{Method} & \textbf{Best lr} & \textbf{Val} & \textbf{Test} & \textbf{Wall (h)} & \textbf{ms/step} & \textbf{Peak MB} \\
\midrule
Muon & 0.003 & 3.741 & 3.632 & 2.95 & 793 & 24650 \\
MuonW & 0.001 & 3.846 & 3.742 & 2.95 & 793 & 24650 \\
AdamW & 0.0003 & 3.925 & 3.826 & 2.89 & 776 & 24990 \\
SOAP & 0.003 & 3.935 & 3.825 & 4.15 & 1120 & 27391 \\
Sven & 0.1 & 5.198 & 5.107 & 9.25 & 2517 & 36050 \\
\bottomrule
\end{tabular}
}
\end{table}

\section{Varying the $\kappa$ parameter}
\label{app:kappaScans}
In Section~\ref{sec:sven_general_case} we introduced the parameter $\kappa$, which defines the effective residual $\mathcal{R}_{\rm eff}^\alpha = (\ell^\alpha)^{\kappa/2}$. 
In this framing the Sven update step (\Eqref{eqn:update}) is directly analogous to the linear least-squares solution when $\kappa = 1$, and $\kappa = 2$ makes the effective residual the per-sample loss itself. We use $\kappa = 2$ as a default throughout the paper because it does not require taking fractional powers of the loss. In this appendix we explore different values of $\kappa$ and address some subtleties around using $\kappa \neq 2$.

First, we note that using the unsigned effective residual with $\kappa = 1$ is unstable, as it inevitably produces non-finite values within a few epochs. With $\mathcal{R}_{\rm eff} = \ell^{\kappa/2}$ and a scalar mean-squared-error loss, the row of $M$ belonging to a sample is proportional to $\ell^{\kappa/2 - 1}\,\partial\ell/\partial\theta$, which diverges as $\ell \to 0$ for $\kappa < 2$. Taking the residual with its sign, $\mathcal{R}_{\rm eff} = \mathrm{sign}(r)\,|r|^{\kappa}$ for a scalar residual $r$, removes the singularity: the row is proportional to $|r|^{\kappa-1}$, which is finite for $\kappa \ge 1$, and the sign cancels between $M^+$ and $\mathcal{R}$ so the update is unchanged wherever the old form was well defined. With signed residuals, we find that training with $\kappa = 1$ is stable.

Second, we note that $\kappa$ acts on $M$ as a per-sample row scaling: with $g_\alpha = \partial\ell^\alpha/\partial\theta$ the ordinary per-sample gradient, row $\alpha$ of $M$ is $(\kappa/2)\,(\ell^\alpha)^{\kappa/2-1}\, g_\alpha$, so $M = DG$ for a diagonal $D$ that depends on $\kappa$ and on each sample's own loss. Wherever $G$ has full row rank --- the untruncated overparametrized solve --- the pseudoinverse undoes any such scaling exactly, $(DG)^{+} = G^{+}D^{-1}$, and the factor combines with $\mathcal{R}_{\rm eff}^\alpha = (\ell^\alpha)^{\kappa/2}$ to give $\delta\theta = -\eta\,(2/\kappa)\,G^{+}\ell$: the $\kappa$ update is $2/\kappa$ times the $\kappa = 2$ update, for any per-sample loss, scalar or vector target alike. $\kappa$ can therefore only act through the truncation, which is applied to the singular values of $DG$ and so sees the reweighting. Comparing $\kappa$ values at a common $\eta$ therefore compares effective step sizes. We instead compare at matched \emph{effective step} $2\eta/\kappa$. The shared learning-rate list realizes \numKappaNMatched{} common effective steps, $\{\numKappaMatchedSteps{}\}$, at all three $\kappa$, in two arms: $k = 32$, where the truncation is active, and $k = \numKappaB{} = B$, where it is not.

\begin{figure}[t]
    \centering
    \includegraphics[width=\linewidth]{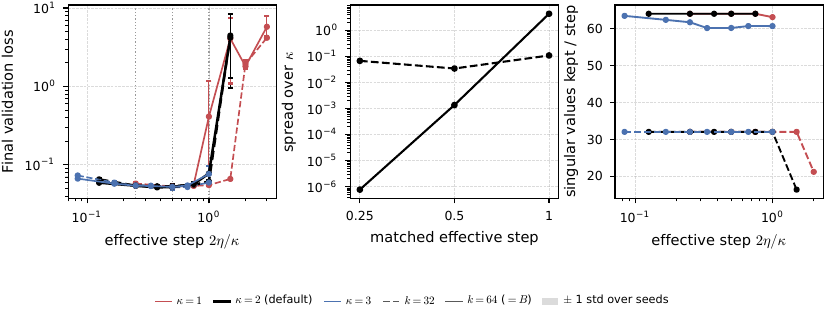}
    \caption{$\kappa \in \{\numKappaValues{}\}$ on MNIST label regression, plotted against the \emph{effective} step $2\eta/\kappa$ rather than against $\eta$, at $k=32$ (truncation active) and $k=\numKappaB{}=B$ (inactive). Left: final validation loss. Center: the spread over $\kappa$ at each matched step. Right: the number of singular values the update inverts against effective step. Bands are $\pm 1$ std over \numKappaNSeeds{} model seeds.}
    \label{fig:kappaScan}
\end{figure}

Fig.~\ref{fig:kappaScan} compares various settings of $\kappa$ on MNIST label regression. In the untruncated arm the scaling argument is confirmed to the precision of the experiment: below the largest effective step size the three $\kappa$ values agree with a small spread. In the truncated arm they do not agree, showing a significantly larger spread. So $\kappa$ acts only through the truncation, which is what Sec.~\ref{sec:sven_general_case} predicts, since $\kappa$ changes the row space of $M$ and therefore which directions survive the cut. 

\section{Decreasing Memory Requirements}\label{app:lims}
Appendix~\ref{app:gram_knobs} defines two knobs for reducing Sven's memory: micro-batching and parameter batching. In principle, both cut down the size of the Jacobian that must be instantiated, though the efficient Gram capture makes this cost saving less relevant\footnote{It is still important when the efficient capture cannot be used, e.g.\ the CIFAR-10 studies with ResNet18 in Appendix~\ref{app:cifar_10}.}. This appendix shows micro-batching and parameter batching results on a few of the small-scale problems (toy 1D, polynomial, and MNIST).

\begin{figure}[t]
    \centering
    \includegraphics[width=\linewidth]{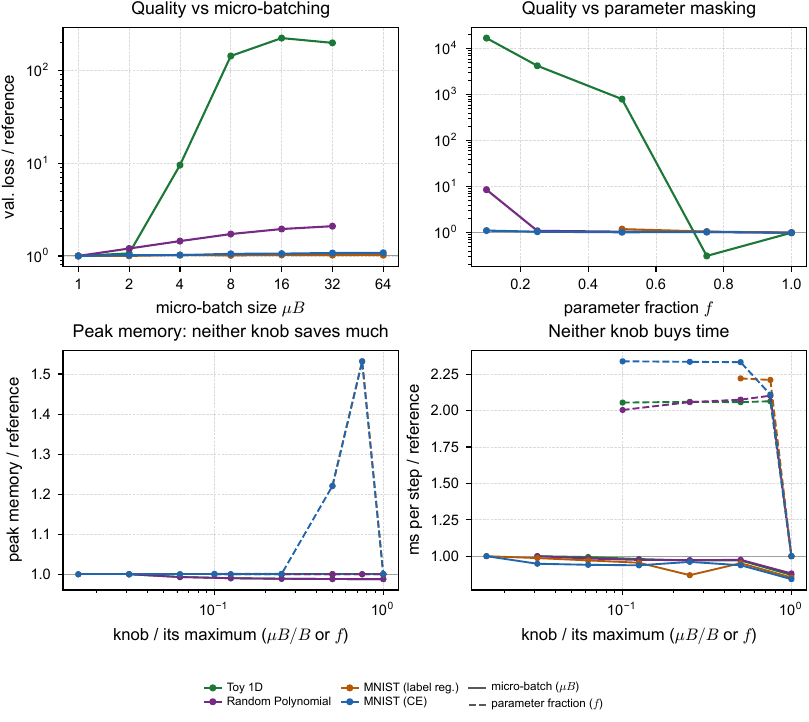}
    \caption{The two memory-mitigation knobs on the four MLP scans, each normalized by its own reference setting. Top: final validation loss against micro-batch size and against parameter fraction. Bottom: peak memory (left) and step time (right) against each knob scaled to its own maximum, relative to the reference setting (solid for micro-batching, dashed for parameter batching).}
    \label{fig:microbatch_paramFrac}
\end{figure}

We vary the micro-batch size $\mu B$ between $1$ and $B$, and the parameter fraction $f \in \{\numParamfracToyValues{}\}$, using Sven's selected best configuration from the large hyperparameter scans. Fig.~\ref{fig:microbatch_paramFrac} summarizes the results, showing a generally modest impact on performance as $\mu B$ and $f$ are varied\footnote{In Appendix~\ref{app:fig5} we found that CIFAR-10 was also fairly robust to reducing parameter fraction down to $f \sim 0.25$, but not below this.}. The notable exception is the 1D regression, which clearly begins to diverge outside of the default settings. This may be related to its strongly hierarchical singular value spectrum observed in earlier appendices. Memory-wise and timing-wise, micro-batching has little impact, as expected. Parameter batching also has little impact on memory but a modest impact on step time. This is consistent with what we observed in Appendix~\ref{app:fig5}, and is likely related to how autograd handles parameter masking.

\section{Robustness and Stability}
\label{app:robustness}

Sven's truncation (\texttt{rtol} and $k$) is designed to avoid inverting tiny singular values, but at tight \texttt{rtol} or large learning rates it can still become unstable, as can most of the baselines at large learning rates. Here we count failures on the four headline tuning grids (toy 1D, polynomial, MNIST label regression, MNIST cross-entropy), and show Sven's failures on the $P/N$ sweeps of Appendix~\ref{app:overparam} and the batch-size sweep of Appendix~\ref{app:batchsize} beside them as examples of where the diverged fraction gets high. A run is diverged if the runner recorded a numerical failure, or if its final validation loss is non-finite or exceeds ten times its initial value.

\begin{figure}[t]
    \centering
    \includegraphics[width=\linewidth]{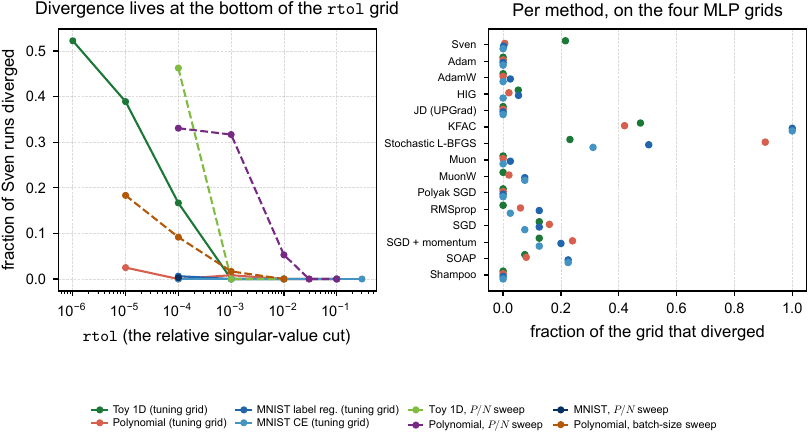}
    \caption{Left: fraction of Sven runs that diverged versus \texttt{rtol}, on the four headline tuning grids (solid) and on the $P/N$ and batch-size sweeps (dashed). Right: fraction of each grid that diverged, per method. Sven's grid includes the \texttt{rtol} axis, so the denominators are not equal across methods.}
    \label{fig:divergence}
\end{figure}

\begin{figure}[t]
    \centering
    \includegraphics[width=\linewidth]{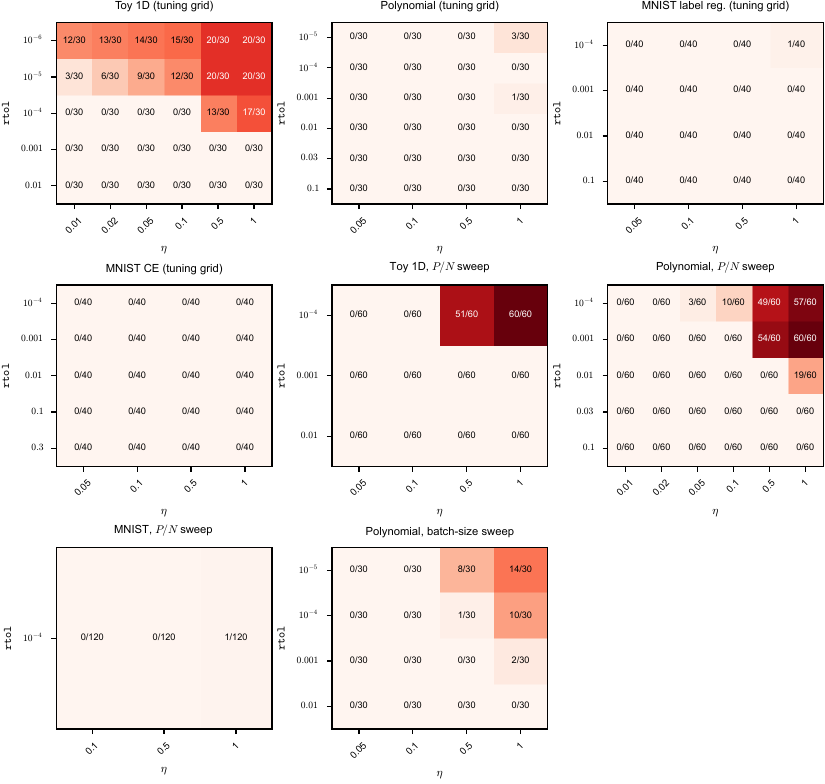}
    \caption{Sven's diverged runs over the $(\texttt{rtol}, \eta)$ plane, as diverged/attempted per cell: the four headline grids, then the three $P/N$ sweeps and the batch-size sweep.}
    \label{fig:divergence-grids}
\end{figure}

Fig.~\ref{fig:divergence} (left) shows Sven's diverged fraction against \texttt{rtol} and Fig.~\ref{fig:divergence-grids} its distribution over the full $(\texttt{rtol}, \eta)$ plane, per grid. On the headline grids the rates are relatively small, with failures concentrating at low \texttt{rtol} and large $\eta$. This is especially pronounced on toy 1D, where the failure rate hits $\sim$50\% at the lowest \texttt{rtol}. The failure rates are larger for the $P/N$ sweeps, likely because they probe unusual settings; the pattern in $(\texttt{rtol}, \eta)$ is broadly the same. 

Fig.~\ref{fig:divergence} (right) gives the diverged fraction per method on the four tuning grids; the denominators differ, since Sven sweeps more hyperparameter configurations. L-BFGS and K-FAC diverge on a substantial number of runs, with K-FAC failing on \textit{all} MNIST runs\footnote{This is due to failures inside the eigendecomposition of its Kronecker-factored Fisher, which is rank-deficient in this regime (Sec.~\ref{sec:related})}.

\section{Memory and Time Profiling}
\label{app:profiling}
This appendix measures Sven's memory and wall-time cost relative to the baseline optimizers. The cost of a Sven step splits into the \emph{capture} (retrieving the Jacobian and/or Jacobian Gram matrix) and the \emph{solve} (eigendecomposition of the Gram matrix or full pseudoinverse of the Jacobian, plus the application of the update). We use the memory-efficient implementation (denoted \texttt{hooks}) described in Appendix~\ref{app:gram} by default, which is generally the most time- and memory-efficient option. In this appendix we analyze that implementation alongside three others: full Jacobian capture + randomized pseudoinverse (\texttt{classic}), full Jacobian capture + Gram eigendecomposition (\texttt{full}), and chunked Gram accumulation where the full Jacobian cannot be instantiated in memory (\texttt{chunked}).

The \texttt{classic} backend is a naive implementation, which costs $\mathcal{O}(k P B)$ due to the explicit $B \times P$ pseudoinverse computation. This can be done with one of three techniques: a full SVD followed by truncation, or two randomized algorithms. The full SVD explicitly decomposes the Jacobian using \texttt{torch.svd} and truncates to the top $k$ singular values; it is the most expensive option, though for very small models and reasonable batch sizes we found it fast and well within memory. The two randomized algorithms share the same Halko--Martinsson--Tropp range-finding approach~\citep{halko2011finding} but differ in how they extract singular values and right singular vectors from the projected matrix. Both target rank $k$, oversample by $p$, and perform $q$ power iterations (in practice, we use $p = 5$ and $q = 1$).

Algorithms~\ref{alg:rsvd-v1} and~\ref{alg:rsvd-v2} describe the two randomized algorithms\footnote{In these algorithms, the symbol $B$ is the \emph{projected matrix} $Q^\top A$, not the batch size}. Given a matrix $A \in \mathbb{R}^{m\times n}$, the first variant (Algorithm~\ref{alg:rsvd-v1}) is simply the standard recipe: build an $m \times r$ orthonormal basis $Q$ of the dominant range of $A$ via QR, form the small projected matrix $B = Q^{\top} A \in \mathbb{R}^{r \times n}$, and call a dense SVD on $B$. This is numerically stable --- in particular, the singular values of $B$ inherit the conditioning of $A$ rather than its square --- but it requires the SVD backend to handle $B$, which has many more columns than rows.
The second variant (Algorithm~\ref{alg:rsvd-v2}) was introduced as a workaround for cuSolver's \texttt{gesvdj} failing with \texttt{CUSOLVER\_STATUS\_INVALID\_VALUE} when $n$ is very large (e.g.\ a large ResNet). After forming the same $B$, it instead constructs the $r \times r$ Gram matrix $G = B B^{\top}$, recovers $(S, U)$ from a symmetric eigendecomposition of $G$ via \texttt{torch.linalg.eigh}, and obtains the right singular vectors through the rectangular matmul $V_{h} = S^{-1} U^{\top} A$.  The price for this robustness is the explicit squaring of singular values: the condition number of $G$ is the square of that of $B$, so the smallest retained singular values lose roughly half their significant digits\footnote{The memory-optimized Gram formulation has the same limitation}. For our truncated regime, in which $k$ is chosen so that the retained spectrum is well separated from the discarded tail, this loss is acceptable, and this algorithm is the recommended choice whenever $P$ exceeds a few million.

\begin{algorithm}[t]
\caption{Randomized truncated SVD (v1) --- range-finder + dense SVD.}
\label{alg:rsvd-v1}
\begin{algorithmic}[1]
\Require $A \in \mathbb{R}^{m \times n}$, target rank $k$, oversampling $p$, power iterations $q$
\State $r \gets \min(k + p,\ \min(m, n))$
\State Draw $\Omega \in \mathbb{R}^{n \times r}$ with i.i.d.\ $\mathcal{N}(0,1)$ entries
\State $Y \gets A\,\Omega$ \Comment{$(m, r)$ random sketch}
\For{$i = 1, \dots, q$}
    \State $Y \gets A\,(A^{\top} Y)$ \Comment{Power iteration}
\EndFor
\State $Q, \_ \gets \operatorname{QR}(Y)$ \Comment{$Q \in \mathbb{R}^{m \times r}$}
\State $B \gets Q^{\top} A$ \Comment{$(r, n)$ projected matrix}
\State $U_{B}, S, V_{h} \gets \operatorname{SVD}(B)$
\State $U \gets Q\,U_{B}$
\State \Return $U_{:,:k},\ S_{:k},\ (V_{h})_{:k,:}$
\end{algorithmic}
\end{algorithm}

\begin{algorithm}[t]
\caption{Randomized truncated SVD (v2) --- Gram-matrix eigendecomposition.}
\label{alg:rsvd-v2}
\begin{algorithmic}[1]
\Require $A \in \mathbb{R}^{m \times n}$, target rank $k$, oversampling $p$, power iterations $q$
\State $r \gets \min(k + p,\ \min(m, n))$
\State Draw $\Omega \in \mathbb{R}^{n \times r}$ with i.i.d.\ $\mathcal{N}(0,1)$ entries
\State $Y \gets A\,\Omega$
\For{$i = 1, \dots, q$}
    \State $Y \gets A\,(A^{\top} Y)$
\EndFor
\State $Q, \_ \gets \operatorname{QR}(Y)$ \Comment{$Q \in \mathbb{R}^{m \times r}$}
\State $B \gets Q^{\top} A$ \Comment{$(r, n)$}
\State $G \gets B\,B^{\top}$ \Comment{$(r, r)$ Gram matrix}
\State $\Lambda, W \gets \operatorname{eigh}(G)$ \Comment{Sort descending}
\State $S \gets \sqrt{\max(\Lambda, 0)}$
\State $U \gets Q\,W$
\State $V_{h} \gets \operatorname{diag}(S^{-1})\,U^{\top} A$ \Comment{Single matmul; safe for any $n$}
\State \Return $U_{:,:k},\ S_{:k},\ (V_{h})_{:k,:}$
\end{algorithmic}
\end{algorithm}

Our profiling runs follow a standard form: initialize the model, the data and the optimizer, move \numProfStagedBatches{} pre-loaded batches to the device, record a resident memory baseline, take \numProfWarmupSteps{} warm-up steps and then measure \numProfMeasuredSteps{} optimizer steps. We report the peak memory during each step and, as the step time, the mean over the last \numProfAmortPct{}\% of them. The profiler measures \numProfNArchs{} architectures: three MLPs (1D, polynomial, MNIST), nanoGPT and ResNet18. We profile each of Sven's four backends as well as twelve baseline optimizers. For L-BFGS we fix \texttt{max\_iter} $=3$ throughout.

\begin{figure}[!ht]
    \centering
    \includegraphics[width=\linewidth]{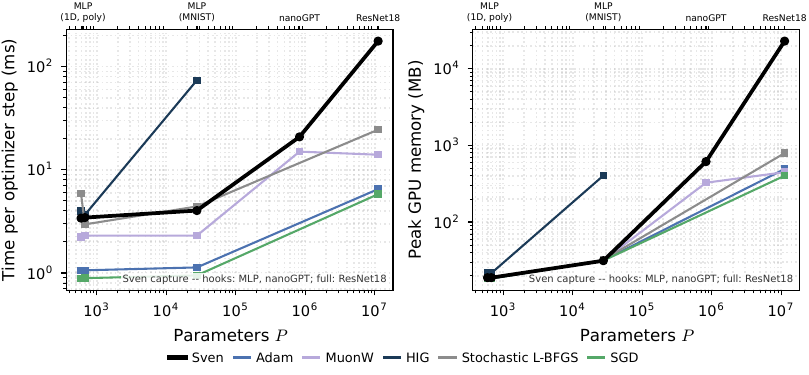}
    \caption{Time per optimizer step (left) and peak GPU memory (right) against parameter count over \numCostMemoryNArchs{} model scales spanning \numCostMemoryPMin{} to \numCostMemoryPMax{} parameters (\numCostMemoryPMaxArch{}), at each method's selected configuration.}
    \label{fig:cost_memory}
\end{figure}

Figure~\ref{fig:cost_memory} summarizes step time and peak memory of each method's selected configuration against parameter count, over five architectures. Sven uses \texttt{hooks} capture in all cases except for ResNet18, where it uses \texttt{full}. Sven uses nearly the same amount of memory as Adam for the MLPs, increasing somewhat for nanoGPT (for reasons discussed in Appendix~\ref{app:transformers}) and far exceeding it for ResNet18 due to the full Jacobian capture. Sven is roughly 2-3 times slower than Adam for the MLPs, becoming slower again for nanoGPT (same reasons) and ResNet18. 

\begin{figure}[!ht]
    \centering
    \includegraphics[width=\linewidth]{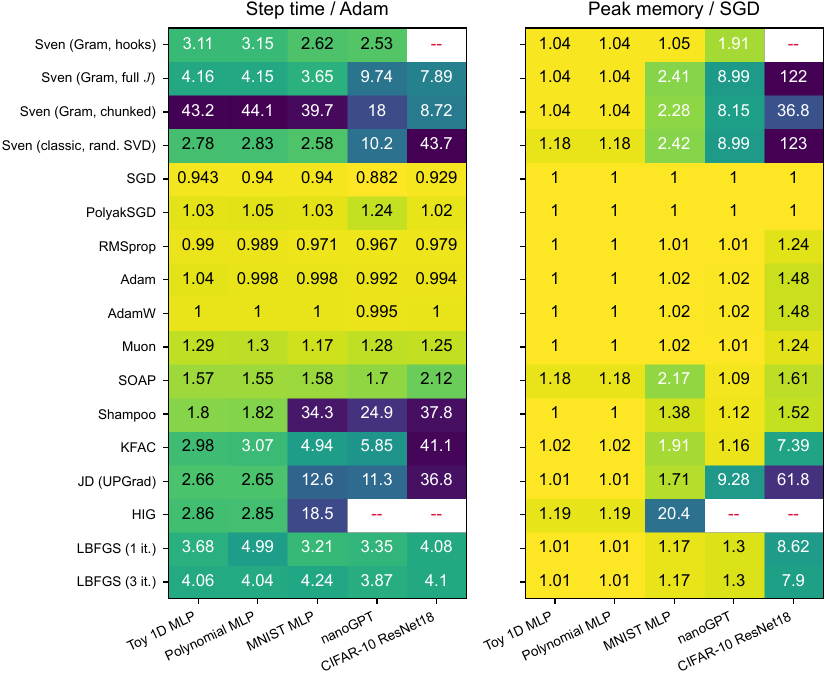}
    \caption{Step time relative to Adam (left) and peak memory relative to SGD (right) for every method at every architecture, at each architecture's profile set point.}
    \label{fig:memory_timing_profiling_comparisons}
\end{figure}

Figure~\ref{fig:memory_timing_profiling_comparisons} gives step time relative to Adam and peak memory relative to SGD for every method at every architecture. The \texttt{hooks} capture is the most efficient, with all other Sven backends using more memory and/or time per step across the board. When the \texttt{hooks} capture is unavailable for ResNet18, the \texttt{chunked} capture saves significant memory while sacrificing a little step time relative to a full capture. The \texttt{classic} backend is suboptimal for any reasonably large model, with the MLPs less affected due to their small parameter count.

\begin{figure}[!ht]
    \centering
    \includegraphics[width=\linewidth]{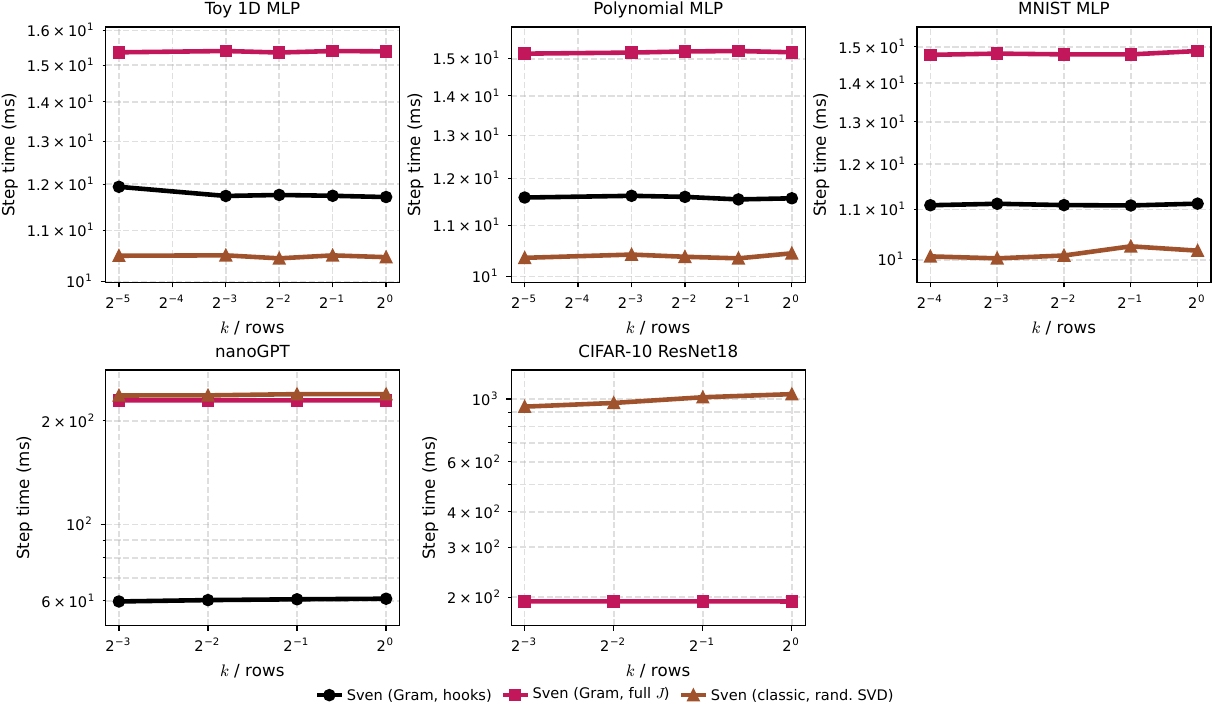}
    \caption{Step time against the rank cap $k$, per architecture and backend.}
    \label{fig:profile_k_sweep}
\end{figure}

\begin{figure}[!ht]
    \centering
    \includegraphics[width=\linewidth]{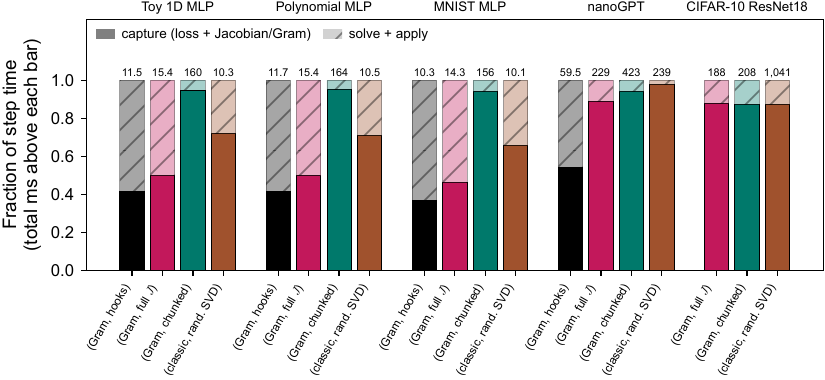}
    \caption{Share of a Sven step spent in the capture (solid) against the solve-plus-apply (hatched), per backend and architecture, with the total step time in ms above each bar.}
    \label{fig:profile_phases}
\end{figure}

Figure~\ref{fig:profile_k_sweep} measures step time as a function of $k$ across the different Sven backends. Aside from large time differences between backends, none varies substantially with $k$. 
Figure~\ref{fig:profile_phases} splits each Sven step into the capture and the solve-plus-apply across backends. The capture is cheapest for \texttt{hooks}, as expected, and most expensive for the \texttt{chunked} capture which serially processes portions of the full architecture. The \texttt{full} and \texttt{classic} backends both materialize the Jacobian, and their captures cost the same; \texttt{full}'s smaller capture \emph{fraction} comes from a larger solve. At this scale a whole step is ${\sim}10$~ms and the Gram solve is dominated by fixed latencies: a float64 \texttt{eigh} through cuSolver costs a few ms however small the matrix, and recovering $J^\top w$ takes one extra forward and backward pass, whereas the classic path already holds $J$ and applies it as one matmul. The Gram solve is therefore near-constant on MLPs, whichever capture fed it. The split only measures compute once the model is large enough for the solve to have real work in it: on ResNet18 the classic solve is a randomized SVD over a $128 \times \numCifarLRParams{}$ Jacobian at \numProfCifarClassicSolveMs{}~ms against \numProfCifarFullSolveMs{}~ms for the Gram solve, and the two backends' capture fractions agree.

\begin{figure}[!ht]
    \centering
    \includegraphics[width=\linewidth]{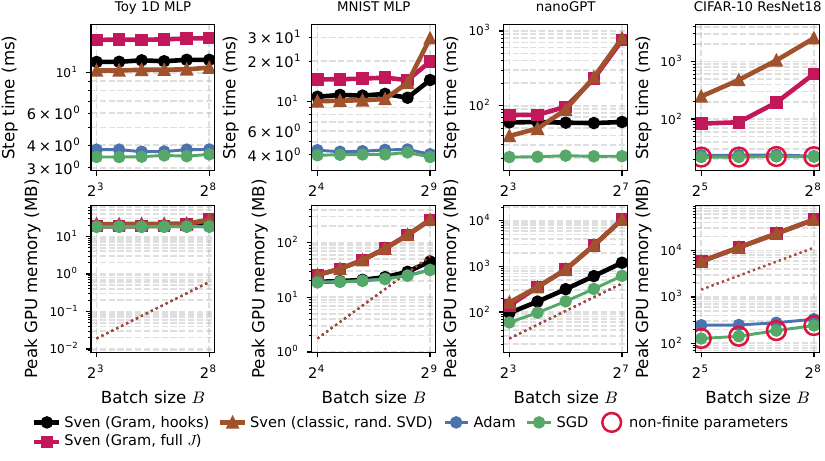}
    \caption{Step time (top) and peak memory (bottom) against the batch size $B$, per architecture. The dotted line on the memory panels is the memory size of the dense Jacobian.}
    \label{fig:mnist_profiling_memory_scans}
\end{figure}

\begin{figure}[!ht]
    \centering
    \includegraphics[width=\linewidth]{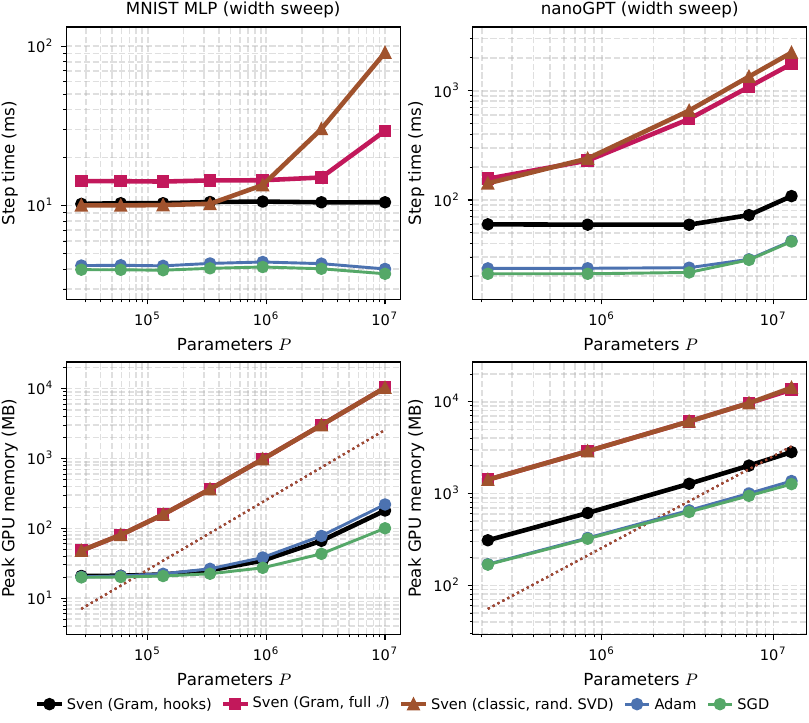}
    \caption{Step time and peak memory against parameter count over two width families: a perceptron sweep that runs past ResNet18's parameter count, and a nanoGPT sweep.}
    \label{fig:mnist_param_scaling}
\end{figure}

Figure~\ref{fig:mnist_profiling_memory_scans} sweeps the batch size, which is directly proportional to the Jacobian/Gram matrix dimensions; the dotted line on the memory plots shows the size of the dense Jacobian. Figure~\ref{fig:mnist_param_scaling} sweeps the parameter count over two model families: an MLP that runs past ResNet18's parameter count, and nanoGPT. The \texttt{hooks} step time is constant for the MLP while rising modestly for nanoGPT, whereas the \texttt{full} and \texttt{classic} backends steadily increase. The \texttt{hooks} memory usage closely tracks Adam for the MLP, and scales identically to Adam but at a slightly higher baseline for nanoGPT.

\end{document}